\documentclass{ifm-tr}

\usepackage{amsmath,amsfonts,bm}

\def\eqref#1{equation~\ref{#1}}

\def\1{\bm{1}}

\DeclareMathAlphabet{\mathsfit}{\encodingdefault}{\sfdefault}{m}{sl}
\SetMathAlphabet{\mathsfit}{bold}{\encodingdefault}{\sfdefault}{bx}{n}

\usepackage{amsmath,amsfonts,amssymb,mathtools,bm}  
\usepackage{thmtools}          
\usepackage{algorithm}         
\usepackage{algpseudocode}     
\usepackage{enumitem}          
\usepackage{siunitx}           
\usepackage{tikz}              

\usepackage{hyperref}       
\usepackage{url}            

\usepackage{xurl}          
\usepackage{booktabs}       
\usepackage{amsfonts}       
\usepackage{nicefrac}       
\usepackage{microtype}      
\usepackage{graphicx}
\usepackage{subcaption} 
\usepackage{xspace}
\usepackage[frozencache,cachedir=.]{minted}
\usepackage{longtable}
\usepackage{adjustbox}

\usepackage{nicematrix}
\usepackage{multirow}
\usepackage{makecell}
\usepackage{soul}  
\usepackage{wrapfig}
\usepackage{enumitem}

\usepackage{lipsum}
\usepackage{tcolorbox}
\usepackage{fancyvrb}
\tcbuselibrary{breakable}
\usepackage{fontawesome5} 
\usepackage{pdflscape} 

\usepackage{colortbl}
\usepackage{ulem}

\usepackage{pgfplots}
\usepackage[table]{xcolor}
\pgfplotsset{compat=1.17}

\definecolor{colbg}{HTML}{F2F2F2} 
\definecolor{dimrule}{HTML}{CCCCCC} 

\usepackage{microtype} 
\usepackage{listings}
\usepackage{placeins}
\usepackage{xspace}
\newcommand{\huggingface}{\raisebox{-1.5pt}{\includegraphics[height=1.05em]{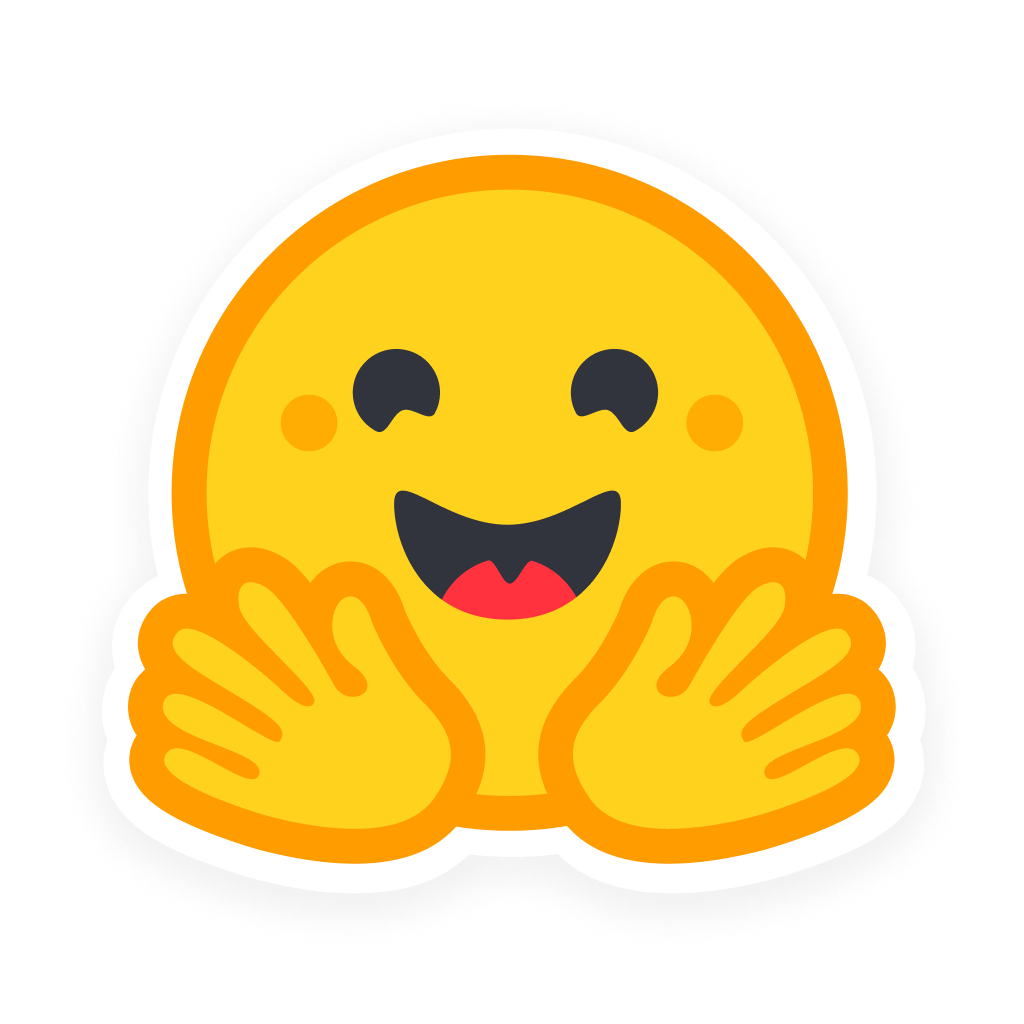}}\xspace}
\newcommand{\github}{\raisebox{-1.5pt}{\includegraphics[height=1.05em]{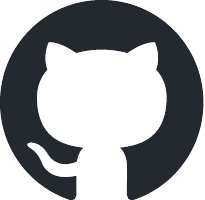}}\xspace}

\newcommand{\wandb}{\raisebox{-1.5pt}{\includegraphics[height=1.05em]{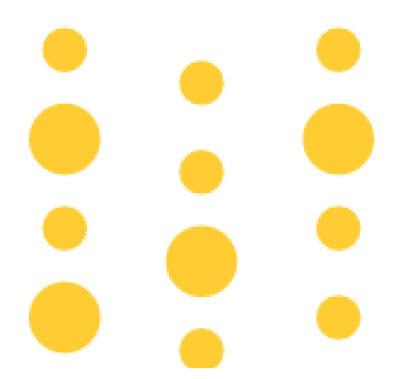}}\xspace}

\definecolor{cellHighlight}{HTML}{dbefff}
\definecolor{promptbg}{gray}{0.95}

\newcommand\rurl[1]{%
    \href{https://#1}{\nolinkurl{#1}}%
}

\newcommand\grurl[1]{%
    \href{https://github.com/#1}{\nolinkurl{#1}}%
}

\newcommand\hrurl[1]{%
    \href{https://huggingface.co/#1}{\nolinkurl{#1}}%
}

\newcommand\wrurl[1]{%
    \href{https://wandb.ai/#1}{\nolinkurl{#1}}%
}

\newcommand{\ktwo}{\texttt{K2}}

\newcommand{\role}[2]{%
  \noindent\textbf{#1}\\[1pt]%
  #2\par\vspace{3pt}%
}
\newtcolorbox{AIbox}[2][]{aibox,title=#2,#1}

\definecolor{blue-primary}{HTML}{0081FB}
\definecolor{blue-light}{HTML}{D4ECFF}
\newtcolorbox{keytakeaways}[1][]{
    enhanced,
    colback=blue-light, 
    colframe=blue-primary,
    colbacktitle=blue-primary,
    coltitle=white,
    fonttitle=\bfseries\large\color{white},
    title={\faLightbulb\, Key Takeaways},
    after=\par\vspace{0.5em},
    arc=2.5pt,
    boxrule=0.6pt,
    left=9pt,
    right=9pt,
    top=8pt,
    bottom=8pt,
    drop fuzzy shadow, 
    #1
}

\title{\bf K2-V2: A 360-Open, Reasoning-Enhanced LLM}

\author[*]{\textbf{K2 Team, Institute of Foundation Models}}

\affiliation[]{Mohamed bin Zayed University of Artificial Intelligence}

\reportnumber{2025-12-02}

\abstract{

We introduce K2-V2, a 360-open LLM built from scratch as a superior base for reasoning adaptation, in addition to functions such as conversation and knowledge retrieval from general LLMs. It stands as the strongest fully open model, rivals open-weight leaders in its size class, outperforms Qwen2.5-72B and approaches the performance of Qwen3-235B. We actively infuse domain knowledge, reasoning, long-context, and tool use throughout the training process. This explicitly prepares the model for complex reasoning tasks. We demonstrate this potential using simple supervised fine-tuning, establishing a strong baseline that indicates significant headroom for advanced alignment. By releasing the full training history and data composition, we maximize the effectiveness of continuous training, a key open source production scenario. We release the model weights and signature LLM360 artifacts, such as complete training data, to empower the community with a capable, reasoning-centric foundation.
}

\begin{document}
\maketitle

\renewcommand{\thefootnote}{\fnsymbol{footnote}}
\footnotetext[1]{Correspondence to: \texttt{\{hector.liu, eric.xing\}@mbzuai.ac.ae}}
\renewcommand{\thefootnote}{\arabic{footnote}}





\begin{minipage}{\textwidth}
\vspace{-18pt}

\centering
\scriptsize
\begin{tabular}{ l l l l l l }
\huggingface & Model Weights & \hrurl{collections/LLM360/k2-v2} & \huggingface & Vibe-Eval Gallery & \hrurl{spaces/LLM360/k2v2-vibe} \\
\wandb & Training Log & \wrurl{llm360/K2-V2} & \huggingface & Mid-training Data & \hrurl{datasets/LLM360/TxT360-Midas} \\
\github & Evaluation & \grurl{llm360/eval360} & \huggingface & Pre-training Data & \hrurl{datasets/LLM360/TxT360}\\
\github & Training & \grurl{llm360/k2v2_train}  & \huggingface & SFT Data & \hrurl{datasets/LLM360/TxT360-3efforts}\\
\end{tabular}

\vspace{-12pt}
\end{minipage}

\begin{figure}[htbp]
    \centering
    

    \includegraphics[width=0.85\linewidth]{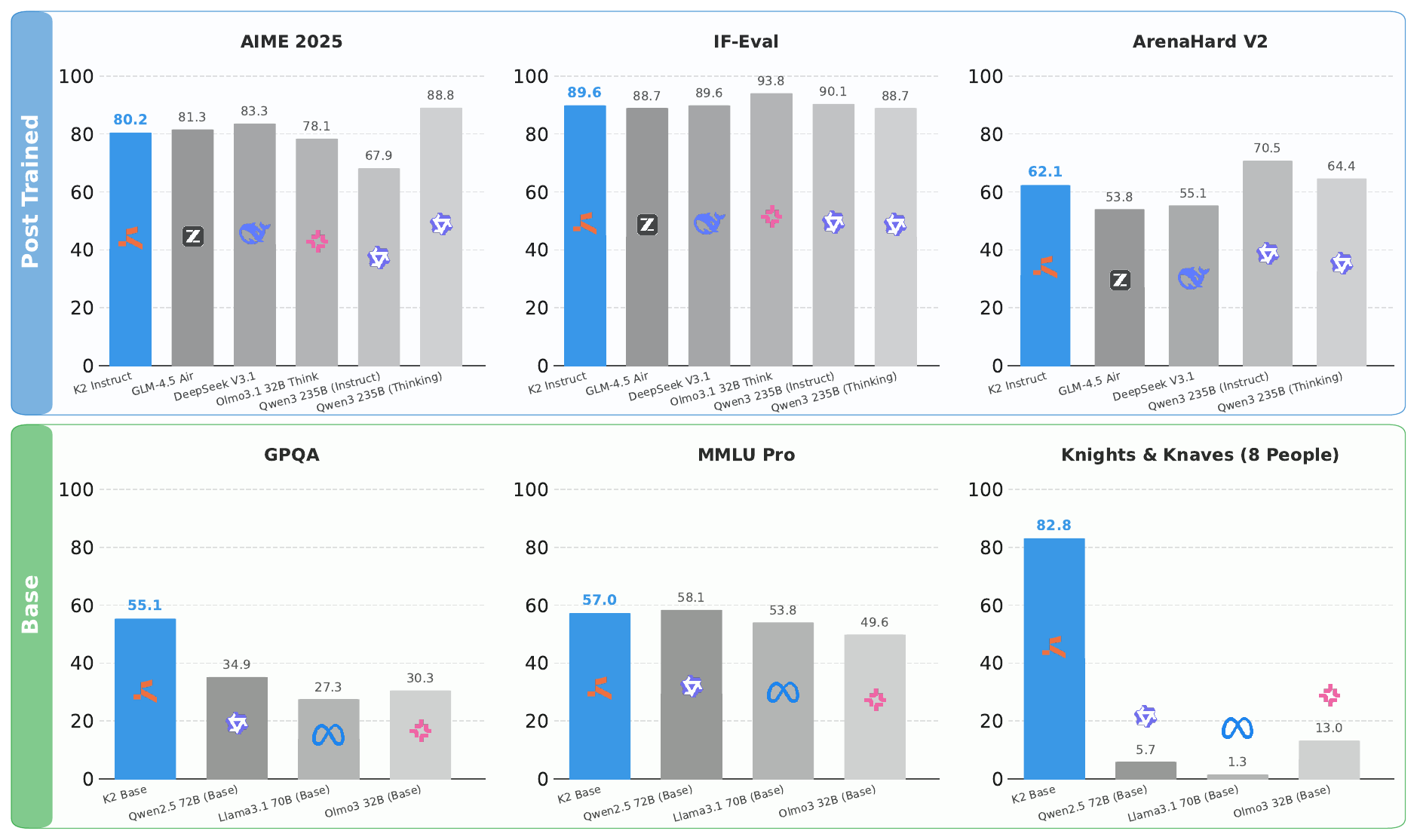}
    \vspace{-9pt}
    \caption{K2 delivers a strong base model for various reasoning scenarios (bottom), which can be elicited via a brief SFT stage: K2 Insturct is competitive with reasoning-tuned models of larger sizes (top). Olmo 3.1 32B is included as a recent fully open source model for reference.}
    \label{fig:top}
\end{figure}

\newpage
\setcounter{tocdepth}{2}
\tableofcontents
\newpage


\section{Introduction}\label{sec:intro} 

We introduce \ktwo{}, the best fully open-source pretrained large language model (LLM) to date, and ranks competitively against the best open-weight models of its class. As the latest base model in the LLM360 family~\citep{liu2023llm360, crystalcoder2024model, K2, k2_think}, \ktwo{} features a dense architecture with 70 billion parameters. Beyond standard competencies like knowledge and conversation, \ktwo{} provides advanced capabilities, including long context consistency, deep mathematical knowledge, and reasoning behaviors. These serve as foundational building blocks that enable sophisticated downstream use cases, such as solving complex math problems and executing agentic workflows. Our supervised fine-tuning experiments illustrate this potential, confirming that \ktwo{} is a robust base primed for further post training. Continuing the 360-open-source tradition of LLM360, we release \ktwo{} along with its complete development transparency -- a documentation of the entire development-cycle, and a full suite of artifacts along the way, providing the community with an end-to-end blueprint and toolkit for building and evolving a state-of-the-art AI model.

The urgency of providing a strong open base LLM  stems from a growing divergence in the industry: while model capabilities have surged, the transparency required to scientifically understand and continous adaptation has significantly diminished. Specifically, as frontier models become more powerful, their development details, specifically data compositions, infrastructure configurations, and training dynamics, have increasingly become ``black boxes''. Crucially, we acknowledge that in this landscape, merely being ``open'' is not enough; models that lack sufficient power offer limited utility for both research and production. For example, reasoning tuning on weak base models yields little scientific insight~\citep{wang2025octothinker, cheng2025revisiting}.

This reality underpins the design philosophy of \ktwo{}, we ensure the model is not only transparent but also practically usable and competitive with state-of-the-art models of similar scales. Our supervised fine-tuning experiments confirm its potential, illustrating that \ktwo{} is a robust base primed for further post-training. Consequently, we position the \ktwo{} project to contribute to the field from two distinct and critical angles: model utility and acceleration of open science.

\paragraph{K2 provides unique model utilities.} \ktwo{} provides practitioners with a uniquely adaptable base. \ktwo{} establishes itself as the best-in-class performer for its size (Fig.~\ref{fig:top}), validating the effectiveness of our open recipe. Crucially, the model delivers the advanced capabilities introduced above, such as long context consistency and reasoning. Further, its utility extends beyond raw performance, and lies in its adaptability. For open models, continuous training and domain adaptation are essential. By documenting our mid training stages and data compositions, we remove the guesswork often associated with continuous training. 
This full transparency allows practitioners to implement targeted strategies, such as data replay~\citep{ibrahim2024simplescalablestrategiescontinually}, to maintain the model's base capability. For production environments, this ensures the model can be adapted easily without the trial-and-error required by closed models.

\paragraph{K2 accelerates open science for AI.} \ktwo{} offers a highly capable testbed for academic research. Unlike weaker baselines, \ktwo{} meets the performance threshold required to investigate complex cognitive behaviors, such as mathematical reasoning and agentic workflows. By serving as a transparent alternative to the limited set of capable open models currently available, we enable researchers to rigorously dissect how these high-level abilities emerge.

Furthermore, we bridge the significant information gap between model builders and evaluators. Without access to training corpora, it is difficult to trace model strengths and weaknesses back to their origins or to isolate the variables that drive performance. This transparency is particularly critical for post-training research; the efficacy of techniques such as Reinforcement Learning often varies significantly in the literature, which is likely due to unobserved distributions in the base model's pre-training data. By fully releasing the training data, \ktwo{} allows the community to gain a deeper understanding of these dependencies, linking post-training outcomes directly to pre-training signals.

Finally, the \ktwo{} project continues a track record of promoting scientific progress through high-impact contributions. Building on our development of TxT360~\citep{txt360data2024}, a widely adopted dataset featuring innovative upsampling methodologies, we significantly reduce the engineering burden of repetition in the field. In this release, we expand this ecosystem with new foundational assets: \textit{TxT360-Midas}, an open mid-training dataset designed to shape reasoning behaviors and extend context lengths; and \textit{TxT360-3efforts}, a supervised fine-tuning dataset curated with mixed reasoning efforts. By democratizing these resources, we empower researchers to bypass the costly work of reproduction and focus their resources directly on innovation.

\section{Overview}\label{sec:overview}

\begin{figure}
    \centering
    \includegraphics[width=\linewidth]{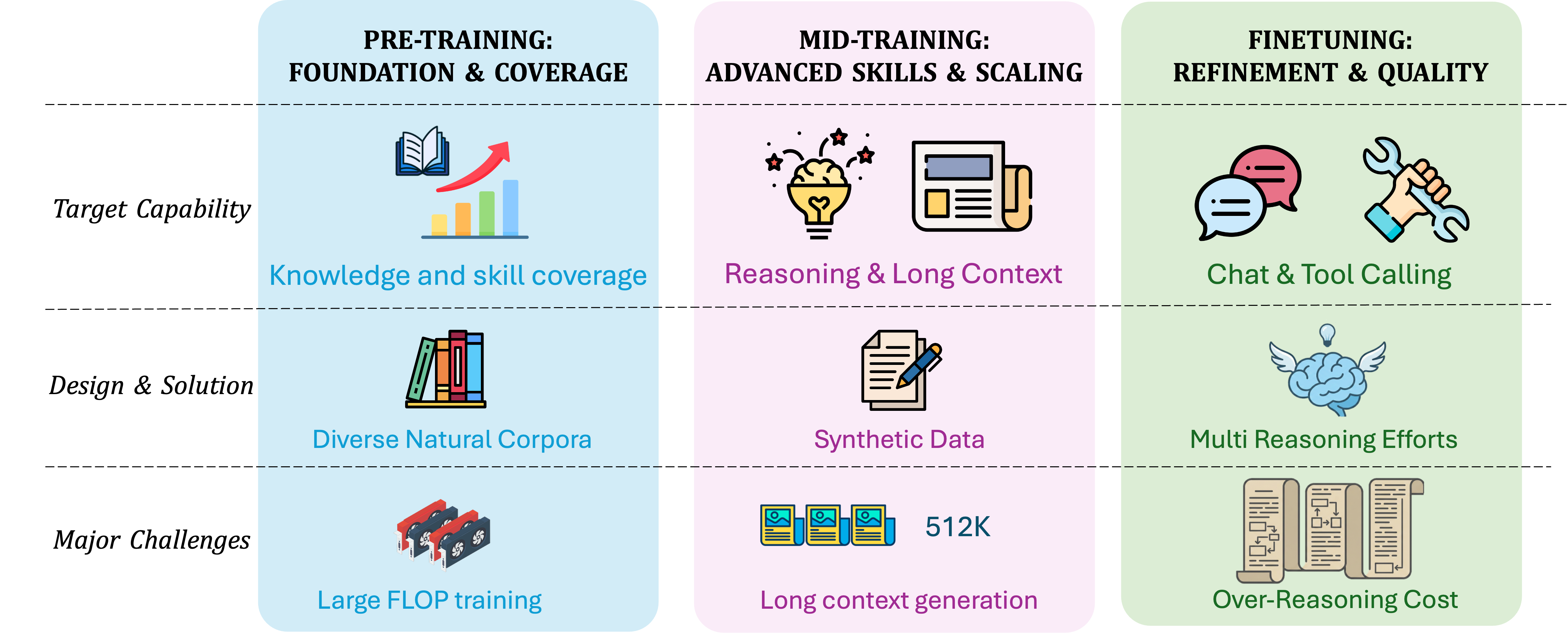}
    \caption{The training phases of K2 are designed to progressively enable specific capabilities. Each phase introduce different types of challenges that we have to address. \label{fig:k2-train-phases}}
\end{figure}

The training strategy of \ktwo{} is driven by a capability first approach. While modern Large Language Models are widely used as general purpose assistants, the landscape has shifted. OpenAI o1~\citep{jaech2024openai} demonstrated the potential of LLMs to solve complex tasks via long horizon reasoning. Subsequent frontier models have further pushed these boundaries by coupling reasoning capabilities with external tool use~\citep{openai2025gpt5systemcard, kimik2thinking, grok4modelcard}. Inspired by these developments, we designed the \ktwo{} base to not only serve as a strong general purpose model but also provide a robust foundation for a specialized reasoning engine. To achieve this, we prioritized five foundational pillars:

\begin{enumerate}
    \item \textbf{Broad Knowledge Coverage:} 
    This capability establishes a robust linguistic foundation. It ensures fluency and generalization, acting as the critical substrate upon which advanced reasoning capabilities are built.
    
    \item \textbf{Deep Domain Knowledge:} 
    To support expert level problem solving, we target technical depth beyond general breadth. This involves integrating concentrated data from specialized fields, including mathematics, programming, and scientific literature.
    
    \item \textbf{Consistent Long Context Capabilities:} 
    We design the model to maintain coherence over extended sequence lengths. While essential for document analysis, this capability plays a central role in maintaining logical consistency across long horizon reasoning chains.
    
    \item \textbf{Exposure to Reasoning Behaviors:} 
    Unlike traditional pipelines that reserve reasoning for fine tuning, we introduce reasoning behaviors early in the training phase. This strategy aims to establish reasoning behaviors such as planning, logical deduction, backtracking as core  primitives.
    
    \item \textbf{Robust and Native Tool Calling Capabilities:} 
    Tool calling serves a dual purpose: it empowers general assistant scenarios, such as web search, and supports complex reasoning by delegating precise tasks, like calculation or code execution, to external environments.
\end{enumerate}

To effectively synthesize these capabilities, we structure the training pipeline into three major phases, as illustrated in Fig.~\ref{fig:k2-train-phases}.

The pretraining phase establishes a robust knowledge base through the massive ingestion of diverse natural data. The primary objective here is effective scaling, which encompasses not only raw compute regarding token count and model size but also the maintenance of optimal training dynamics. We address these complexities through a tripartite strategy: rigorous training dynamic design and monitoring, robust infrastructure capable of supporting large scale execution, and the curation of massive token counts to drive data scaling.

In the mid training phase, the curriculum shifts toward specialization, long context extension, and exposure to reasoning behaviors. This introduces a distinct set of challenges, including computational bottlenecks, the risks of distribution drift, and the scarcity of high quality long context data. We navigate these by implementing efficient Context Parallel strategies to resolve computational constraints, engineering precise token mixtures to manage domain shift, and augmenting natural sources with synthetic generation to address data availability. Furthermore, we adopt specific learning dynamic adjustments, such as a low and constant learning rate, to ensure stability during this specialization process.

Finally, we conduct a simple supervised fine-tuning to demonstrate the potential of the model. Here, we inject tool use capabilities and finalize model behaviors by curating a focused, high value corpus. The results of this stage confirm the effectiveness of the previous phases, demonstrating the model's potential as a robust base for further post training.

Crucially, addressing the unique challenges of each phase requires synchronized design across three domains: data composition, training infrastructure, and machine learning dynamics. Consequently, the organization of this paper mirrors this lifecycle, dissecting the challenges of each phase through these lenses.

\begin{itemize}
    \item \textbf{Section \ref{sec:pre-training}} details the Pre-training phase. We first describe the model architecture choice in \S\ref{sec:architecture}. We discuss how we achieved effective scaling by optimizing training dynamics (\S\ref{sec:training-dynamics}), curating a data mix with most natural data (\S\ref{sec:pretrain_data}). \S\ref{sec:pretrain_infra} presents the parallelism strategies for high token throughput.
    \item \textbf{Section \ref{sec:mid-training}} describes the mid-training phase. Here, we break down our solutions regarding the challenges faced during long context and advanced model capabilities. We describe our in-house implementation for context parallelism for efficient long context scaling (\S\ref{sec:mid-train-infra}), synthesizing data to solve scarcity, and designing data curriculum to alleviate the effect of domain shift (\S\ref{sec:mid-train-data}). \S\ref{sec:mid-train-run} presents our observation during the run.
    \item \textbf{Section \ref{sec:base_eval}} presents the evaluation of the base checkpoints, presents the evaluation methodologies and challenges. With the LLM360 trademark metric curves, we show how a model develops its capability from each stage.
    \item \textbf{Section \ref{sec:sft}} presents a simple supervised fine-tuning recipe that demonstrate the potential of the base models.
    \item \textbf{Section \ref{sec:chat_eval}} concludes the benchmark results, analyzes the model performance across challenging tasks spanning general knowledge, STEM, logic, coding and tool-calling.
    \item \textbf{Section \ref{sec:capability_study}} provides a qualitative study of the development of the model capabilities over time.
\end{itemize}

\section{Pre-training}\label{sec:pre-training}

\subsection{Model Architecture}\label{sec:architecture}
Our model follows a decoder-only dense transformer architecture consisting of 80 layers. It has a hidden dimensionality of 8,192 and an intermediate size of 28,672 in the feed-forward network. The attention module includes 64 heads, corresponding to a per-head dimension of 128. To improve memory and compute efficiency during attention computation, we adopt grouped-query attention (GQA), using 8 key-value heads shared across all query heads. Rotary position embeddings (RoPE) are applied with $\theta = 500,000$. RMSNorm is used with an epsilon of $10^{-5}$.

\subsection{Pre-training Recipe}\label{sec:training-dynamics}
\begin{keytakeaways}
\begin{itemize}
    \item \textbf{Prioritize Sample Efficiency over Step Efficiency.} For data-limited runs (fixed token budget) optimizing model quality, do not target the Critical Batch Size (which balances step and token efficiency). Instead, operate in the \textit{sample-efficient regime} (often $B << B_{crit}$).
    \item \textbf{Scale hyperparameters to preserve optimizer dynamics.} 
    Avoid tuning learning rate ($\eta$) and weight decay ($\lambda$) in isolation. We recommend characterizing optimizer behavior via the \textit{averaging timescale} $\tau_{\text{epoch}}$ and scaling this target proportionally to $1/\sqrt{\text{TPP}}$ (Tokens Per Parameter) to ensure consistent training dynamics across model scales.
    \item \textbf{Hyperparameter choices are co-designed by learning dynamics, numerical implementations, hardware constraints and curriculum design.} There are challenges to match scaling law predicted optimal hyperparameters: batch size choice is constrained by the number of accelerators, while small learning rates may interact negatively with numerical precision and additional training curriculum considerations.
\end{itemize}
\end{keytakeaways}

We aim to minimize validation loss under a fixed token budget. To achieve this, we focus on jointly optimizing batch size ($B$), learning rate ($\eta$), and weight decay ($\lambda$).

\paragraph{Hyperparameter Selection.}
Following \citet{wang2025how}, we characterize optimizer dynamics via the effective averaging timescale $\tau_{\text{epoch}} = \frac{B}{\eta \lambda D}$, which approximates the effective number of epochs over which AdamW exponentially averages parameter updates. We derive our target $\tau_{\text{epoch}}$ by tuning $\lambda$, $\eta$, and $B$ at a smaller scale (20 tokens per parameter or TPP) and scaling the optimal timescale proportionally to $1/\sqrt{\text{TPP}}$ based on \citet{bergsma2025power}.

We determine the batch size using a scaling law similar to \citet{bergsma2025power}. Unlike the critical batch size of \citet{mccandlish2018empirical}, which balances sample and step efficiency, we target the optimal batch size that minimizes final validation loss given a fixed token budget. This theoretical value is then adjusted to align with MLSys constraints (accelerator counts). For the learning rate, we sweep $\eta$ at small scale and extrapolate using the $1/d_{\text{model}}$ scaling convention (standard practice from \citep{yang2021tuning}).

\paragraph{Scheduler Ablation.}
We first evaluate several decay schedules using a 257M parameter pilot model trained on $41.48 \times 10^9$ tokens. These utilize a peak learning rate of $1.88 \times 10^{-3}$, a batch size of $0.9 \times 10^6$ tokens, and a weight decay of $0.137$. Consistent with \citet{bergsma2025straight}, we find negligible differences between d2z schedules on the SlimPajama validation set \citep{cerebras2023slimpajama}: 2.815 for cosine decay-to-zero (d2z), 2.845 for cosine decay-to-10\%, and 2.817 for linear d2z. These pilot results suggests that the specific shape of the decay is less critical than the peak and final values.

\paragraph{Final Hyperparameter choices.}
Based on the methodologies above, we reach the \ktwo{} run's final hyperparameters. The learning rate is set to $\eta=1.5 \times 10^{-4}$, selected by referring to similar scale runs such as Llama3-70B~\citep{llama3} and validating against our small-scale extrapolations.

While the scaling law suggested a batch size of 1136 sequences (sequence length 8192), we adjust this to 1200 sequences to satisfy hardware constraints, resulting in a global token batch size of $B=9.8 \times 10^6$. As we collected $D=12.25 \times 10^{12}$ total tokens, we train for $T=1.25 \times 10^6$ steps. We set weight decay to $\lambda=0.05$ to achieve our target $\tau_{\text{epoch}}=0.1066$.

Finally, we determine the schedule. While our pilot ablations supported a simple cosine decay-to-zero, during the main run we observed that the training loss and parameter norm became stale when the learning rate dropped below $1\%$ of the peak. We therefore modified the schedule to decay to a lower bound of $1.5 \times 10^{-6}$ and held it constant for the remainder of pre-training. This floor prevented the stagnation observed in the "long tail" of the standard cosine schedule while simplifying the transition into mid-training phases.

\subsection{Pre-training Data: TxT360}\label{sec:pretrain_data}

We highlight our curation strategy for the pretraining mix. Unlike later mid-training stages where we introduce more synthetic data, we remain conservative during pretraining. We prioritize natural text to ensure the model is grounded in real-world distributions. Beyond this foundation, we focus on improving domain diversity and maintaining the structure of formats like tables. A key design element is our precise control over the duplication rate \citep{txt360data2024}. We apply several measures, such as URL filtering and deduplication, to ensure every data entry is unique. This guarantees that any repetition in the training process is intentional.

\begin{keytakeaways}
\begin{itemize}
    \item \textbf{Ground synthetic data in natural text.} We focus on natural data distributions to establish a strong foundation in pre-training. Synthetic data is effective when it is derived from existing documents. Around 12 trillion tokens are used in K2's pre-training run.
    \item \textbf{Deduplication enables precise upsampling based on quality.} Global deduplication is the necessary foundation for controlling data mix: it ensures repetition is a design choice, not an accident. Data points should be upsampled based on quality signals, the number of years a document persists is a simple but effective quality signal.
\end{itemize}
\end{keytakeaways}

We construct our pre-training corpus from a diverse set of data sources to comprehensively cover multiple domains, including web text, academic papers, multilingual corpora, code, mathematics, and more. Most datasets are extracted from out prior curation in \texttt{TxT360}~\citep{txt360data2024}. A detailed breakdown of the data composition is provided below:

\begin{itemize}
    \item \texttt{English Web} (\S\ref{sec:pretrain-data-web}):  Large-scale web corpus from \texttt{TxT360}~\citep{txt360data2024} web portion, comprising approximately 3.5 trillion unique words deduplicated from 99 CommonCrawl dumps. Inspired by DCLM~\citep{li2024datacomp} and Fineweb-Edu~\citep{lozhkov2024fineweb-edu}, A high-quality subset: \texttt{TxT360-BestOfWeb}, filtered by the ProX \citep{zhou2024programming} document filtering model, is also selected.
    
    \item \texttt{Synthetic QA} (\S\ref{sec:txt360-qa}): Synthetic question–answer pairs generated from \texttt{TxT360} documents using \texttt{Mistral-7B-Instruct-v0.3}~\citep{jiang2023mistral}, totaling roughly 950 billion words.
    
    \item \texttt{Papers} (\S\ref{sec:papers}): Academic articles from ArXiv, PubMed, S2ORC, and PhilPapers, each processed with dedicated dataset-specific preprocessing and filtering  pipelines, comprising approximately 107 billion words in total.
    
    \item \texttt{Math} (\S\ref{sec:megamath}): A collection of large-scale, math-focused corpora curated base on the \texttt{MegaMath} project \citep{zhou2025megamath}, \texttt{DM Math}~\citep{gao2020pile, saxton2019analysingmathematicalreasoningabilities}, \texttt{InfiMM-WebMath-40B}~\citep{han2024infimmwebmath40badvancingmultimodalpretraining}, \texttt{OpenWebMath}~\citep{paster2023openwebmath}, and \texttt{MathPile}~\citep{wang2024mathpilebilliontokenscalepretrainingcorpus}, consisting of approximately 344 billion tokens in total. 
    
    \item \texttt{Code} (\S\ref{sec:opencoder}): Programming code primarily sourced from \texttt{RefineCode}~\citep{huang2025opencoder}. The dataset is further augmented with two additional variants: one organized by repository and topology sorted~\citep{guo2024deepseekcoderlargelanguagemodel}; one with Fill-in-the-Middle (FIM) \citep{bavarian2022efficienttraininglanguagemodels} applied.
    
    \item \texttt{Multilingual} (\S\ref{sec:multilingual}): Our multilingual data is focused on Arabic collected mainly from CommonCrawl and web sources. In addition, we incorporate other languages from high-quality sources like Wikipedia and EuroParl.

    \item \texttt{Other High-Quality Data Sources} (\S\ref{sec:txt360-other}):  The remaining high-quality subset of \texttt{TxT360} that does not fall into the categories above. This includes a diverse collection of well-curated domains such as Wikipedia, EuroParl, FreeLaw court opinions, StackExchange forums, USPTO patents, PG-19, HackerNews, and Ubuntu IRC logs, each processed with dedicated dataset-specific preprocessing, filtering, and deduplication pipelines.

\end{itemize}
An overview of the complete pre-training data is presented in Table~\ref{tab:pretraining_data}. Note that we use word count instead of token count to eliminate the effect of tokenizer. Further details are discussed in the subsequent sections.

\begin{figure}[h!]
    \centering
    \includegraphics[width=0.9\textwidth]{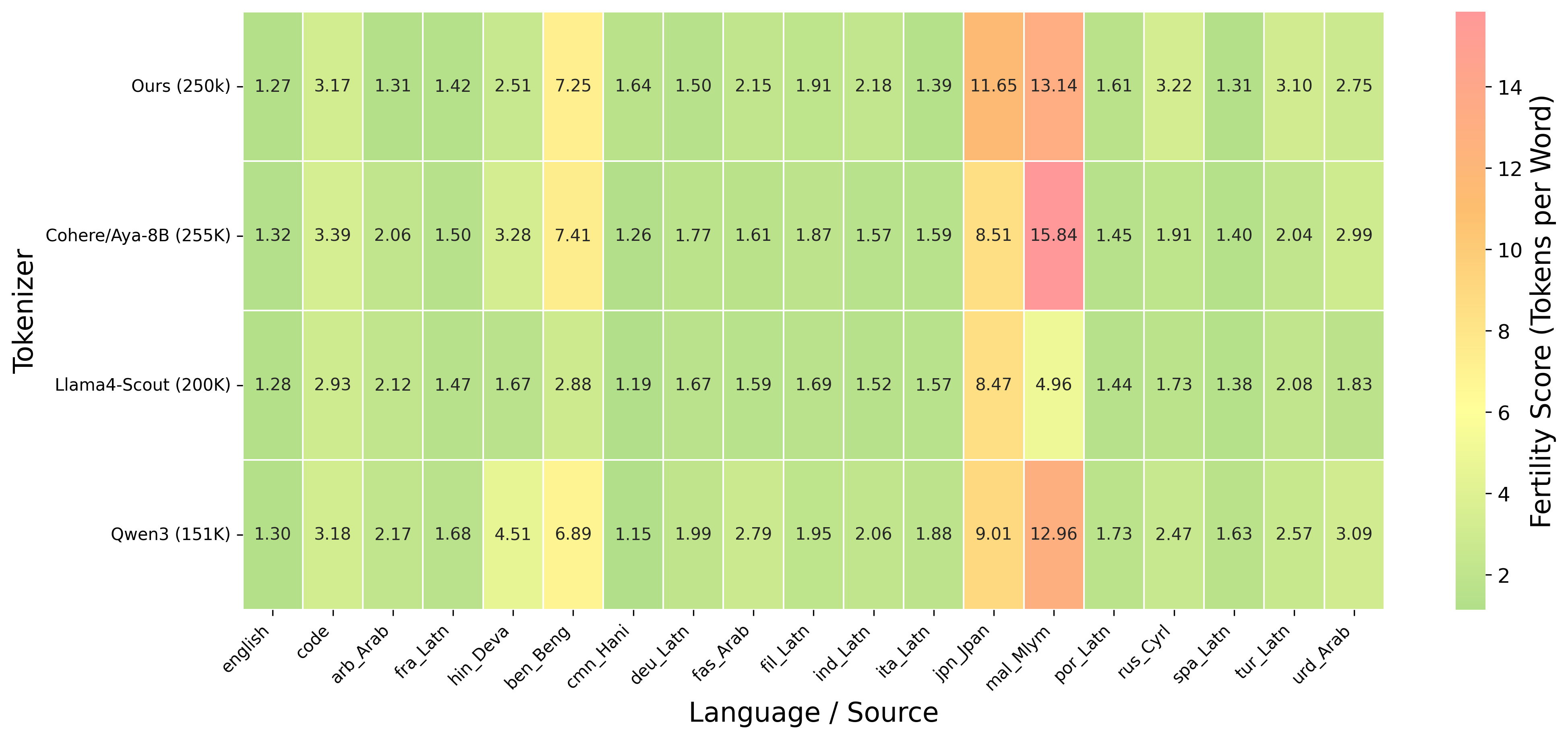}
    \caption{A comparison between the \ktwo{} tokenizer vs. a few multilingual LLM tokenizers. Our tokenizer is tailored towards English and Arabic (MSA), as shown by the lower fertility scores.\label{fig:tokenizer_fertility}}
\end{figure}

\paragraph{Tokenizer.} To translate this diverse data composition into an efficient input representation, we trained a Byte-Level Byte-Pair Encoding (BPE)~\citep{sennrich2016neuralmachinetranslationrare} tokenizer on a weighted mixture of multilingual data, specifically sampled to optimize fertility scores for English and Arabic (Fig.~\ref{fig:tokenizer_fertility}).
To ensure robust character coverage and prevent fragmentation, we initialized the vocabulary with comprehensive Unicode ranges spanning Arabic, Latin, Cyrillic, Devanagari, and CJK scripts. The pre-tokenization pipeline employs a custom regex pattern designed to handle code syntax, whitespace consolidation, and punctuation isolation efficiently. Finally, the vocabulary was augmented with special tokens to support fill-in-the-middle (FIM) objectives and repository-level coding contexts. In the SFT stage we introduce additional special tokens for reasoning and tool-use (see \S\ref{sec:sft-recipe}).

\begin{table}[h!]
\centering
\small
\resizebox{\textwidth}{!}
{%
\begin{tabular}{p{3cm} p{5cm}ccc}
\toprule
\textbf{Category} & \textbf{Data Source} & \textbf{Word Count} & \textbf{Avg. Words/Document} & \makecell{\textbf{Information}\\\textbf{Cut-Off Date}} \\
\midrule
\multirow{2}{*}{\texttt{English Web}} 
    & CommonCrawl & 3.5 T  & 544 & July 2024 \\
    & TxT360-BestOfWeb & 695 B & 1.2 K & Q3 2024 \\
\midrule
\multirow{1}{*}{\texttt{Synthetic QA}}
    & TxT360-QA & 950 B & 1.6 K & Q3 2024 \\
\midrule
\multirow{6}{*}{\texttt{Papers}}
    & Arxiv & 11.1 B & 6.3 K & Q4 2023 \\
    & Pubmed Central & 24 B & 5.2 K & Q4 2023 \\
    & Pubmed Abstract & 4.7 B & 205 & Q4 2023 \\
    & S2ORC Abtract & 11.3 B & 166 & Q4 2023 \\
    & S2ORC Full text & 55.4 B & 7.6 K & Q4 2023 \\
    & Phil Papers & 544 M & 11.8 K & Q4 2023 \\
\midrule
\multirow{4}{*}{\texttt{Math}} 
    & MegaMath Web & 137 B & 1.2 K & Nov 2024 \\
   & MegaMath Code & 5.7 B & 432 & Nov 2024 \\
   & DM Math & 2 B & 18 & -- \\
   & Other Math (InfiMM-WebMath-40B,
OpenWebMath, Textbooks-MathPile)  & 24.4 B & 1.2 K & Q4 2023 \\
\midrule
\multirow{3}{*}{\texttt{Code}} 
    & RefineCode & 265 B & 340 & - \\
   & RefineCode FIM & 238 B & 355 & - \\
    & RefineCode Topo-Sorted & 22 B & 2.8 K & - \\
\midrule
\multirow{2}{*}{\texttt{Multilingual}}
  & Jais Arabic Data & 203 B  & 407 & Q1 2024 \\
  & Europarl-Aligned & 666 M & 63 K & 2012 \\
\midrule
\multirow{9}{*}{\parbox{2cm}{\texttt{Other High-Quality Data Sources}}}
    & Wikipedia & 15 B & 302 & Q4 2023 \\
    & Wikipedia-Extended & 46 B & 1.4 K & Q4 2023 \\
    & Europarl & 670 M & 9.7 K & 2012 \\
    & FreeLaw & 12 B & 2.5 K & Q1 2024 \\
    & StackExchange & 10.7 B & 464 & Q4 2023 \\
    & USPTO & 3.8 B & 742 & Q3 2024 \\
    & PG-19 & 1.95 B & 69 K & 1919 \\
    & HackerNews & 658 M & 821 & Q4 2023 \\
    & Ubuntu IRC & 862 M & 28 K & Q3 2024 \\
\bottomrule
\end{tabular}
}
\caption{Comprehensive overview of pre-training data sources across major categories and subcategories. Word counts are computed using whitespace as the delimiter. }
\label{tab:pretraining_data}
\end{table}

\subsubsection{English Web}\label{sec:pretrain-data-web}
The largest component of our pre-training corpus is mainly composed of natural English text from the \texttt{TxT360}~\citep{txt360data2024} dataset, including its CommonCrawl subset (\texttt{TxT360-CC}) and the filtered high-quality variant, \texttt{TxT360-BestOfWeb}.

\begin{figure}[h!]
    \centering
    \includegraphics[width=0.98\textwidth]{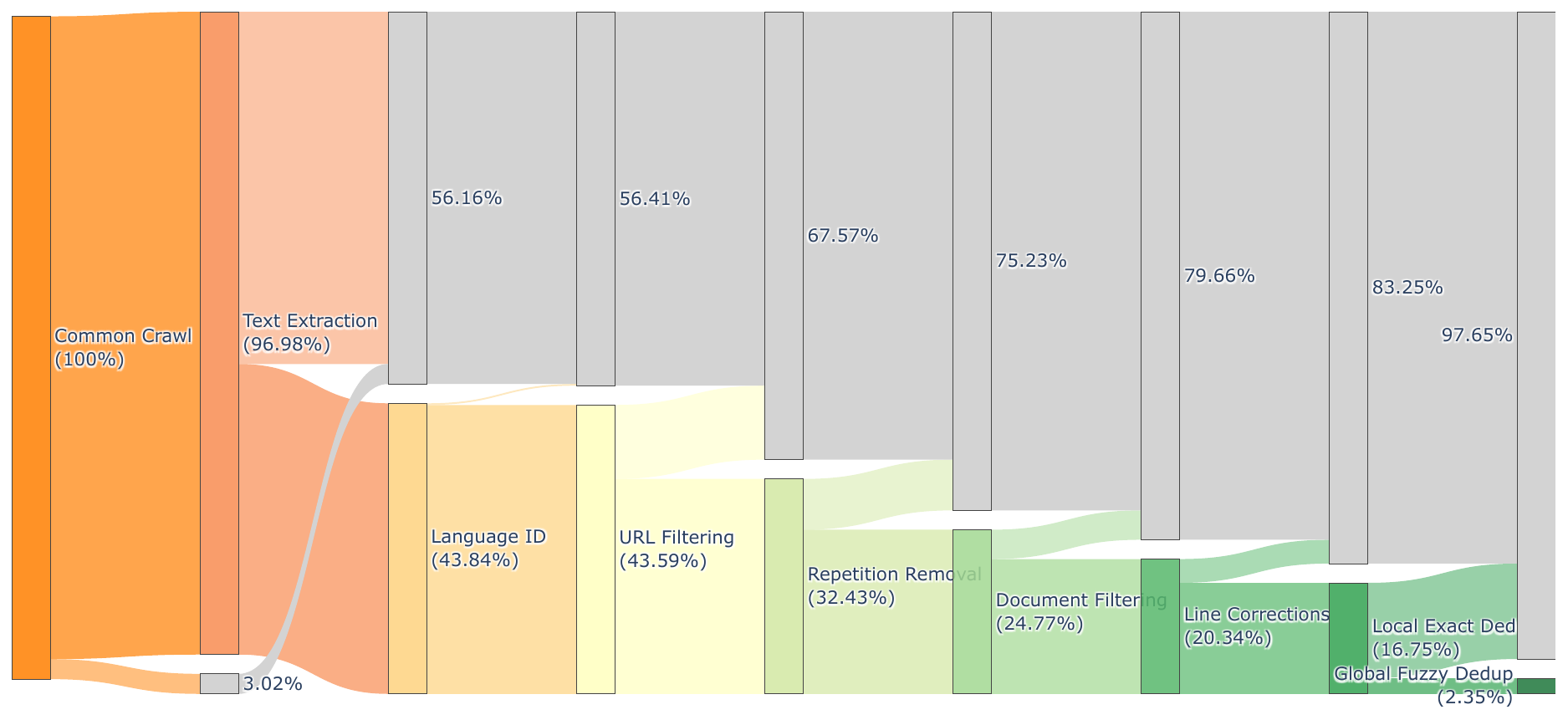}
    \caption{Overview of our CommonCrawl data-curation pipeline, showing filtering percentages (by document count) at each stage. Grey bars indicate filtered documents. \textit{Line-level cleaning} and \textit{PII removal} eliminate almost no documents and are omitted from the visualization. The document-level quality-filtering stage includes \textit{Repetition Removal}, \textit{Document Filtering}, and \textit{Line Correction}. A \textit{Local Exact Deduplication} step is applied prior to global fuzzy deduplication to reduce computational overhead.
    \label{fig:cc-filtering-pipeline}}
\end{figure}

\paragraph{TxT360-CC}
Starting from 99 raw WARC (Web ARChive) snapshots, we apply a comprehensive data-curation pipeline consisting of text extraction, language identification, URL filtering, line-level cleaning, document-level quality filtering, personally identifiable information (PII) removal, and global deduplication. An overview of this pipeline, along with filtering percentages at each stage, is shown in Figure~\ref{fig:cc-filtering-pipeline}.
The resulting \texttt{TxT360-CC} subset contains approximately 6.4 billion documents and more than 3.5 trillion words. Documents average 544 words in length (range: 40 to 100{,}000 words), for a total corpus size of roughly 9.4~TB. The document-length distribution has a median of 330 words, with the 25th and 75th percentiles at 156 and 645 words, respectively. Further details about this curation process are available in the \texttt{TxT360} blog~\citep{txt360data2024}.

\paragraph{TxT360-BestOfWeb.} Recent ``learned'' curation methods use LLM-driven scoring and refinement to improve pretraining corpora. ProX (``Programming Every Example'')~\citep{zhou2024programming} frames data refinement as a program executed on each document, implemented by a small language model.

TxT360 Best of Web is curated by applying a document level filter on the full \texttt{TxT360-CC} set.
This filter is a combination of the ProX document filter model, combined with a separate format score that captures document presentation features like headings, lists, tables, code blocks, link density, and boilerplate. This process selects high-quality, clean and well-structured pages. About 22\% of the documents are retained.

Since \texttt{TxT360-BestOfWeb} is a high-quality subset of \texttt{TxT360-CC}, distinguishing it allows us to flexibly upsample this data during pre-training. We discuss these mixing details in Section~\ref{sec:pretraining-mix}.

\subsubsection{Synthetic QA}\label{sec:txt360-qa}

In the pre-training phase, we introduce synthetic data conservatively. We rely on text synthesized from existing documents rather than pure generation. Document-grounded Question-Answering has a proven history. For example, \cite{alberti-etal-2019-synthetic} created a system to extract answers and generate corresponding questions. By filtering for consistency, they achieved significant gains on benchmarks like SQuAD 2.0~\citep{rajpurkar-etal-2018-know} and NQ~\citep{kwiatkowski-etal-2019-natural}.

\cite{puri-etal-2020-training} demonstrated that large generative models can synthesize high-quality (context, question, answer) triples at scale, sometimes training competitive Question Answering systems entirely from synthetic data. More recently, WRaP \citep{maini2024rephrasingwebrecipecompute} reframes arbitrary web text directly into multiple styles, including explicit question–answer format, and utilizes it in pre-training.

\begin{wrapfigure}[10]{r}{0.45\textwidth} 
  \centering 
  \includegraphics[width=\linewidth]{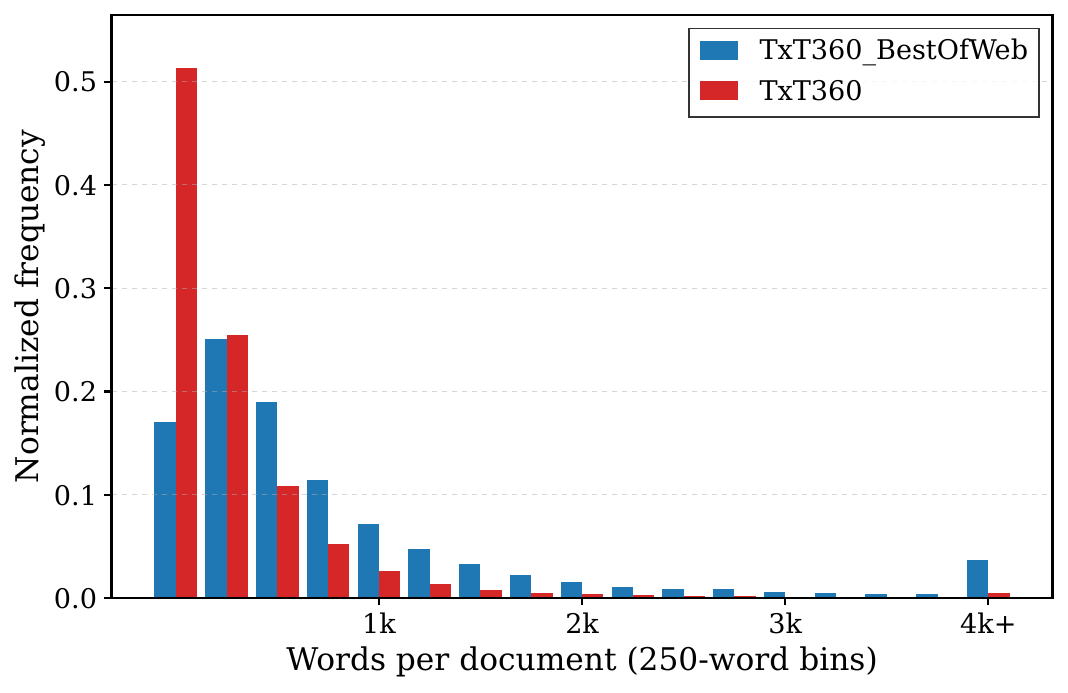} 
  \caption{Normalized document-length distributions for \texttt{TxT360-BestOfWeb} and \texttt{TxT360}. \texttt{TxT360-BoW} has a higher proportion of longer documents as a bias introduced by the quality classifier.}
  \label{fig:BestOfWeb}
\end{wrapfigure}

For our synthetic QA data (\texttt{TxT360-QA}), we augment each document from \texttt{TxT360-BestOfWeb} with multiple, synthetic document-grounded Q-A pairs generated by \texttt{Mistral-7B-Instruct-v0.3}\footnote{Due to a bug in the generation code, we failed to generate Q-A pairs for some of the longer documents.}. The pairs are appended to the end of the document text to enable simple, single-sequence training where a model reads the passage and immediately observes diverse Q–A supervision anchored in that same passage. The number of Q–A pairs per document varies, which naturally yields a mixture of easy and hard supervision signals.

\begin{lstlisting}
{
  "text": "{ORIGINAL_DOCUMENT_TEXT}\n\n
           Q: {QUESTION_1}\nA: {ANSWER_1}\n\n
           Q: {QUESTION_2}\nA: {ANSWER_2}
           ......{ANSWER_N}\n",
  "meta": { original TxT360 meta }
}
\end{lstlisting}

To validate the generated Q–A pairs, we used an LLM-as-a-judge protocol \citep{zheng2023judging}: for each document we asked a judge LLM to award 1 point per pair if both question and answer were fully supported by the passage, and we reported a document-level percentage score.
We apply this method on 8000 random selected documents, obtaining an average score over 95\%, confirming the quality of the Q-A pairs.

We further conducted a manual qualitative check. The low quality questions tend to be generic or template-like (e.g., “what is this used for”), weakly anchored to the page, and hallucinate answers beyond local facts; The high quality questions are attribute-centric and tightly grounded, referencing concrete fields or entities (e.g., dimensions, price, warranty) so that answers can be resolved from a single sentence or short span.

\subsubsection{Papers}
\label{sec:papers}
We incorporate a collection of academic paper in our pre-training data, including  ArXiv, PubMed, S2ORC, and PhilPapers. As part of our \texttt{TxT360} data project, data-specific processing and filtering were applied to each data source.

\paragraph{ArXiv.}  
ArXiv contains LaTeX-formatted preprint papers in mathematics, physics, computer science, and related areas. We download raw \texttt{.tex} files from the official S3 bucket and converted them to Markdown using Pandoc\footnote{https://github.com/jgm/pandoc}. We experiment with different Pandoc parsing options to preserve the structure of tables, math blocks, and section headers. Figures and reference lists were removed.

\paragraph{S2ORC Full Text / Abstract.} 
S2ORC provides scientific papers in JSON format. We extract the title, abstract, section text, and metadata. Papers are kept only if they had a title and abstract, were mostly English, and have enough text after filtering. For the abstracts, we require a valid English abstract and remove documents with fewer than 20 words. We also apply a unigram-probability-based filter.

\paragraph{PubMed Central / PubMed Abstract.}  
PubMed Central provides full-text biomedical papers in JATS XML format; PubMed Abstract contains abstracts in a simpler XML structure. We convert PMC using Pandoc and extracted PMA fields using BeautifulSoup\footnote{https://pypi.org/project/beautifulsoup4/}.

\paragraph{PhilPapers.} 

The PhilPapers dataset consists of philosophy papers in PDF format. We convert the files to text and apply a cleaning pipeline to remove hyphenation, strip headers and footers, normalize whitespace, and fix encoding issues. Finally, we filter the documents based on line length, letter content, and encoding artifacts to ensure document quality.
 
\subsubsection{Math}\label{sec:megamath}

To enhance mathematical and numerical reasoning during pre-training, we incorporate several large-scale math-focused corpora including \texttt{MegaMath} ~\citep{zhou2025megamath}, \texttt{DM-Math}, \texttt{InfiMM-WebMath-40B} ~\citep{han2024infimmwebmath40badvancingmultimodalpretraining}, \texttt{OpenWebMath} ~\citep{paster2023openwebmath} and \texttt{MathPile} ~\citep{wang2024mathpilebilliontokenscalepretrainingcorpus}, consisting of approximated 344 Billion tokens.

\paragraph{MegaMath}
\texttt{MegaMath}~\citep{zhou2025megamath} is a TxT360 project to extract a large-scale high-quality math data from the web, released as an open-source collection specifically designed to preserve mathematical structure across both natural language and code, as our main math corpus. We use the two subsets \texttt{MegaMath-Web} and \texttt{MegaMath-Code} for our pre-training. 

\texttt{MegaMath-Web} is constructed through a math-aware curation pipeline that begins with 99 CommonCrawl WARC snapshots (\texttt{2014-15} to \texttt{2024-46}) and applies a coarse-to-fine extraction and filtering process, including URL filtering, fastText-based language identification, math-specific HTML preprocessing, a two-stage text extraction pipeline, document-level quality filtering, and large-scale MinHash-based deduplication. This pipeline yields approximately 279\,B tokens of high-recall mathematical webpages with faithfully preserved equations and symbols. 

\texttt{MegaMath-Code} is produced by selecting code data from eleven programming languages within Stack~V2, a large public code corpus, and then applying a small language model trained to identify and retain high-quality code relevant to mathematical reasoning, logic puzzles, and scientific computation. The resulting \texttt{MegaMath-Code} split contains approximately 28.1\,B tokens. 

Together, these deduplicated and quality-filtered corpora are decontaminated to remove the overlap between \texttt{MegaMath} and 12 downstream benchmarks, removing 0.01\% of the documents from the dataset. The resulting corpus provides a substantial volume of mathematically rich content. It includes natural language mathematical documents with preserved equations and math-related source code.
More details are available in \cite{zhou2025megamath}.

\paragraph{Additional Math Corpora.}
In addition to \texttt{MegaMath}, we incorporate several widely used open math datasets to increase coverage, diversity, and domain robustness:

\begin{itemize}
    \item \textbf{\texttt{DM Math}} contains generated math question–answer pairs. We converted the dataset to JSONL and fixed a few formatting issues (e.g., byte-string artifacts and converting each row to have only one question and answer). No filtering was applied.

    \item \textbf{\texttt{InfiMM-WebMath-40B}}~\citep{han2024infimmwebmath40badvancingmultimodalpretraining}:
    A collection of math webpages sourced from CommonCrawl (2019–2023), filtered to focus on high-quality mathematical and scientific content in both English and Chinese.
    After deduplication, the corpus is reduced to 15{,}877k documents and 27B tokens.

    \item \textbf{\texttt{OpenWebMath}}~\citep{paster2023openwebmath}:  
     A math-focused dataset filtered and extracted from over 200B HTML files on Common Crawl down to a set of 6.3 million documents containing a total of 14.7B tokens. We obtain 4{,}316k documents and 10B tokens after deduplication.

    \item \textbf{\texttt{MathPile}}~\citep{wang2024mathpilebilliontokenscalepretrainingcorpus}: 
    \texttt{MathPile} is a diverse and high-quality math-centric corpus sourced from high-quality data sources including Textbooks (including lecture notes), arXiv, Wikipedia, ProofWiki, StackExchange, and Web Pages, comprising about 9.5 billion tokens in total.
    Several sources of \texttt{MathPile} are similar to \texttt{MegaMath}, to avoid the overlap, we only select the \textit{Textbooks} subset of \texttt{MathPile}, which includes approximately 4{,}000 documents totaling 0.042B tokens.
\end{itemize}

Finally, we conducted exact deduplication across the math datasets.

\subsubsection{Code}\label{sec:opencoder}

We used RefineCode~\citep{huang2025opencoder} as the base code dataset and produced two additional variants from it: a Fill-in-the-Middle (FIM) version and a topologically sorted version. All three splits originate from the same underlying collection of GitHub repositories.

\paragraph{RefineCode.}
This is the original cleaned code corpus. Each entry corresponds to a single source file extracted from a GitHub repository. No structural modifications were made to the code itself. For details about the data collection methodology, see~\cite{huang2025opencoder}.

\paragraph{RefineCode FIM.}
This split is created from RefineCode by converting each file into a Fill-in-the-Middle (FIM) format~\citep{bavarian2022efficienttraininglanguagemodels}. A span from the file is selected as the “middle” section, and the remaining text is split into a prefix and a suffix. These parts are then rearranged using special tokens such as \texttt{<|fim\_prefix|>}, \texttt{<|fim\_middle|>}, and \texttt{<|fim\_suffix|>} to mark their positions. The token order is chosen randomly among standard FIM patterns, so the same file may appear in different arrangements compared to RefineCode. Only the layout changes; the underlying code content remains the same.

A shortened example of an entry is:

\begin{center}
\small
\begin{tcolorbox}[width=0.85\textwidth, colback=white, colframe=black, arc=0pt, unbreakable]
\begin{Verbatim}[fontsize=\small]
<|fim_prefix|>{% extends 'base.html.twig' %}

{% block title %}Modifier un utilisateur{% endblock %}
{% block body %}
  <div class="container">
    {{ include('user/_form.html.twig') }}
  </<|fim_suffix|> "{{ path('user_index') }}">

<|fim_middle|>div>
    <a href=
\end{Verbatim}
\end{tcolorbox}
\end{center}

\paragraph{RefineCode Topo-Sorted.}

This variant groups files by GitHub repository and sorts them based on the topological structure induced by code imports. Our goal is to ensure that dependencies appear earlier in the sequence so that variables and classes are defined before they are used. We followed the procedure described by \citealp{guo2024deepseekcoderlargelanguagemodel}. First, we identified import statements using regular expressions and built a dependency graph for each multi-file repository. We then applied topological sorting to determine the file order. For isolated files or cases where the graph did not fully determine an order, we preserved the repository's original file listing. Finally, we concatenated the ordered files into a single document, inserting a comment with the filename before each code block.

These three splits represent the same underlying code data in different forms: the unmodified file-level version, an FIM-transformed version, and a topological-sorted version.

\subsubsection{Multilingual}\label{sec:multilingual}

We include a substantial amount of Arabic text in our pretraining corpus, primarily sourced from Common Crawl web scrapes extending the data used to train Jais model ~\citep{sengupta2023jais}.
In addition to web sources, we incorporate high-quality multilingual resources such as Wikipedia and Europarl, ensuring broader linguistic and domain coverage.

\paragraph{Arabic} Our Arabic dataset comprises approximately 498 million documents, totaling over 203 billion words. 
On average, each document contains about 407 words, with document lengths ranging from 2 to 27 million words.  
The document length distribution shows a median of 94 words, with the 25th and 75th percentiles at 38 and 445 words, respectively.

\paragraph{Europarl-Aligned.}
This dataset is a multilingual corpus derived from the Europarl v7 proceedings \citep{koehn-2005-europarl}, consisting of parallel texts from European Parliament debates. This version extends the original Europarl collection by aligning English source sentences with their corresponding parallel translations across multiple European languages (e.g., Finnish, German, Dutch, Greek, Italian, French, Portuguese, Swedish, Danish, Spanish). The alignment process preserves semantic equivalence across languages.

The original Europarl data\footnote{\url{http://www.statmt.org/europarl/v7/europarl.tgz}} is converted into \texttt{JSONL} format for efficient filtering and multilingual concatenation. Each record combines the same parliamentary segment across languages into a unified structure, and the order of the language is randomized (recorded by a metadata field). For example:

\begin{center}
  \begin{tcolorbox}[width=0.8\textwidth, colback=white, colframe=black, arc=0pt, unbreakable]
\begin{Verbatim}[fontsize=\small]
{
  "text": "# Finnish
[Finnish content]

# German
[German content]

# Dutch
[Dutch content]
...",
  "meta": {
    "language": "fi-de-nl-el-it-fr-en-pt-sv-da-es",
    "src_file": "ep-00-01-17"
  }
}
\end{Verbatim}
\end{tcolorbox}
\end{center}

\subsubsection{Other High-Quality Data Sources}\label{sec:txt360-other}

Besides the data set mentioned above, we adopt all the remaining high-quality data sources from \texttt{TxT360}, including FreeLaw, StackExchange, USPTO, PG-19, HackerNews, and Ubuntu IRC.
These datasets are different from general web text because they are published in structured formats such as XML, LaTeX, PDF, or CSV. These sources contain cleaner writing, clearer structure, and more consistent formatting than typical web pages, so they require different processing steps. 

Across all curated datasets, we apply a small number of shared filtering rules:  
(1) removing non-English documents when the dataset is expected to be English;  
(2) dropping very short documents or too high non alphanumeric characters;  
(3) removing documents with very low unigram log probability.  
For some datasets, only a subset of these filters was needed. Many filters and their thresholds have been set after detailed review by our team. Additional dataset-specific cleanup is added only when required by the source format. Details can be found in our TxT360 blog~\citep{txt360data2024}.

\begin{description}
\item[Wikipedia.] 
We use the official Wikimedia snapshot from HuggingFace and convert the parquet files to JSONL. This includes all 323 subsets representing different languages.

\item[Wikipedia Extended]

The Wikipedia Extended dataset is built from standard Wikipedia articles, with each article expanded by adding information from the pages it links to. For every source article, the abstract of each outgoing linked article is appended after the original text. Links and abstracts are obtained via DBpedia~\citep{dbpedia}. Each record contains the concatenated text, for example:

\begin{center}
\small
  \begin{tcolorbox}[width=0.8\textwidth, colback=white, colframe=black, arc=0pt, unbreakable]
\begin{Verbatim}[fontsize=\small]
{ORIGINAL ARTICLE}

{First Link Title}
"{First Link Abstract}"

{Second Link Title}
"{Second Link Abstract}"
...
\end{Verbatim}
\end{tcolorbox}
\end{center}

This produces a longer document that includes the main article followed by short summaries of related pages, providing additional data for training long-context capabilities.

\item[EuroParl.]
EuroParl contains parliamentary debate transcripts with XML-like tags. We remove HTML tags, filtered out very short lines, and keep only lines associated with speaker tags. Other structural tags such as paragraph and chapter markers are discarded.

\item[HackerNews.]  
HackerNews includes threaded discussions. We retrieve posts using the official API, construct threads from story IDs, remove system messages, and simplify comment hierarchies.

\item[USPTO.]  
USPTO patent documents appear in different formats depending on the publication year. We use format-specific extraction scripts and remove documents with fewer than 50 words.

\item[FreeLaw.]  
FreeLaw provides legal case text across multiple HTML and XML representations. We extract text from all relevant fields, normalize whitespace and line breaks, and remove artifacts such as form-feed characters. Local deduplication remove a large number of repeated entries across dumps.

\item[StackExchange.]  
We reconstruct full conversation threads from the StackExchange XML dataset. Using post IDs, we group questions with corresponding answers, comments, all code blocks.

\item[Ubuntu IRC.]  
Ubuntu IRC logs capture chat-style conversations. We remove system messages, anonymize usernames with generic placeholders, and clean message prefixes.

\item[PG-19.]  
PG-19 consists of public-domain books. We reduce long runs of whitespace, normalize line breaks, remove hyphenation across line breaks, and remove delimiter markers. 

\end{description}

\subsubsection{Pre-training Data Analysis}

\begin{figure}[t]
    \centering
    \includegraphics[width=0.7\linewidth]{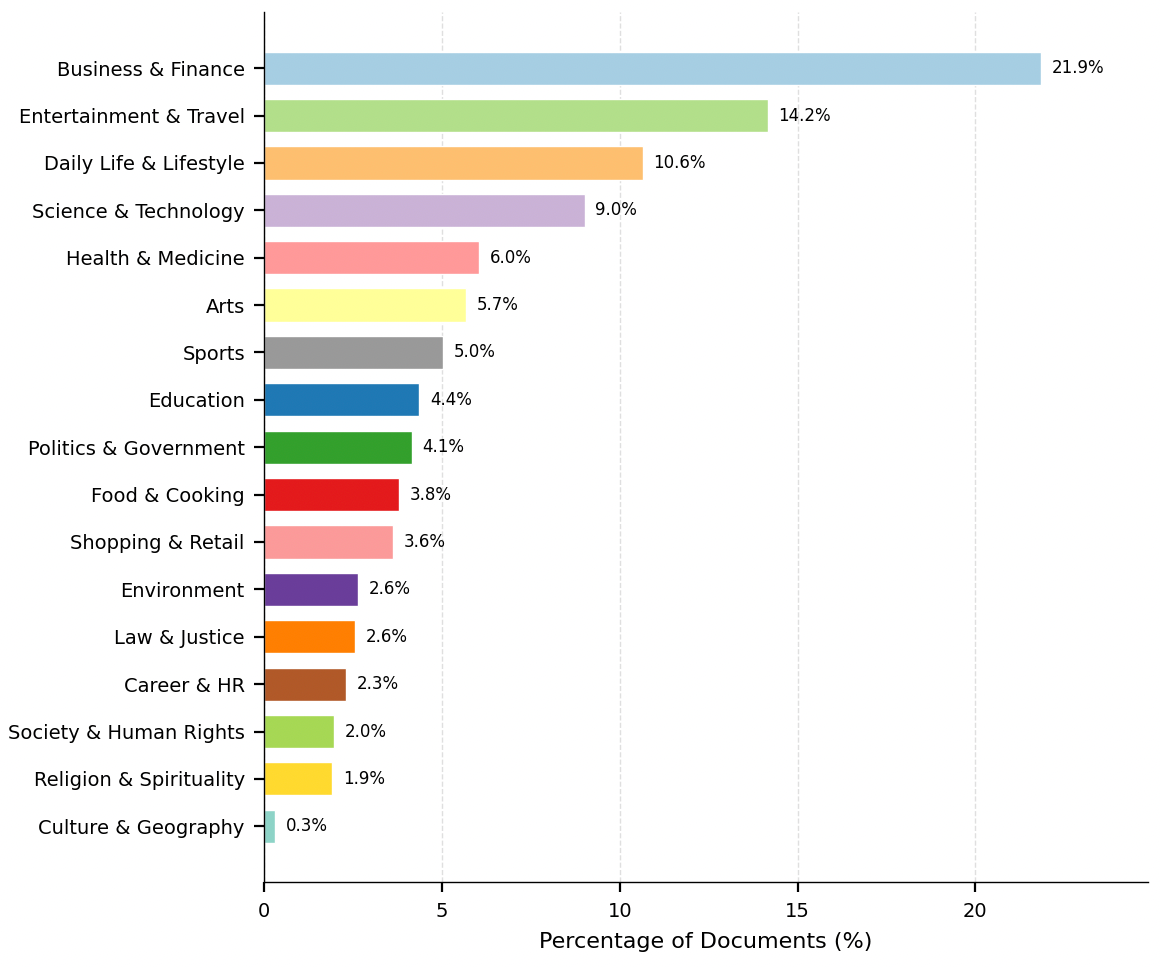}
    \caption{Topic distribution of documents from the Common Crawl portion of TxT360.}
    \label{fig:topic_analysis_overview}
\end{figure}

\paragraph{Topic analysis.}
To obtain a high-level view of the content distribution in our corpus, we perform a topic analysis on the Common Crawl portion of TxT360. We apply BERTopic~\citep{grootendorst2022bertopic} to cluster the documents into 17 broad topic groups. The clusters are produced by first generating 128 initial groups from a seed subset and then manually merging them into semantically coherent categories. For each topic, we aggregate several metadata-based metrics and inspect their averages to understand how different types of documents vary across simple structural and lexical attributes. A summary of the topic distribution is shown in Figure~\ref{fig:topic_analysis_overview}. Additional details about the clustering procedure and metric comparisons can be found in our blog post~\citep{txt360data2024}.

\paragraph{Global Near-Deduplication.}
Near-deduplication provides several benefits for language model (LM) pretraining, the most important being controllable upsampling. With unique data, we gain fine-grained control over the training distribution. Other advantages include preventing train–test overlap, which mitigates benchmark and development set contamination, and reducing the impact of the double descent phenomenon, where repeated data can cause test loss to increase midway through training~\citep{hernandez2022scaling}. Deduplication also lowers the risk of memorization~\citep{lee2022deduplicating}. By implementing deduplication and selective upsampling, we can directly control the pretraining data distribution rather than relying on the inherent distribution of the source datasets.

We start with 61.8~TB of filtered and compressed documents, totaling approximately 48.83~billion entries. First, we perform exact deduplication using a Bloom filter with a capacity of 1~billion and a false-positive rate of~0.001. This process reduce the corpus to 40.21~billion documents, eliminating about~17\% as exact duplicates. The Bloom filter’s constant memory usage makes this stage efficient and reduces the workload for subsequent near-deduplication.

For global near-deduplication, we adopt a methodology similar to prior work such as SlimPajama~\citep{cerebras2023slimpajama}, but scale it to encompass the entire dataset, including 99~Common Crawl snapshots and curated data. The process involves generating \texttt{MinHash} signatures for each document using the \texttt{datasketch} library with 128~permutations.

Before signature generation, the text of each document is cleaned:
\begin{itemize}[noitemsep,topsep=0pt]
    \item Stripping leading and trailing whitespace
    \item Converting all text to lowercase
    \item Removing punctuation, consecutive spaces, newlines, and tabs
\end{itemize}

From the cleaned text, we generate 13-grams as features to construct document signatures, each represented as a \texttt{MinHash} object. These signatures are then compared to identify near-duplicate documents, which are clustered to retain a single representative document while marking the others for deletion. When selecting representatives, we prefer curated documents over Common Crawl documents, and more recent documents over older ones.

Throughout all stages of deduplication, we use the \texttt{Dask} distributed processing library extensively for scalability and efficiency. In addition, we maintain detailed statistics for each matching cluster during the final deduplication stage.

\begin{table}[ht!]
\centering
\small
\renewcommand{\arraystretch}{0.9}
\begin{tabular}{p{3cm} p{6.5cm} c r}
\toprule
\textbf{Category} & \textbf{Data Source} & \textbf{Duplication Bucket} & \textbf{Proportion (\%)} \\
\midrule

\multirow{12}{*}{\texttt{English Web}} %
  & \multirow{6}{*}{\parbox{6cm}{TxT360-CC \\ (TxT360-BestOfWeb excluded)}}
      & 1-1 & 14.291033\\
  & & 2-5 & 9.280918\\
  & & 6-10 & 2.998148\\
 &  & 11--100 & 3.494916\\
 &  & 101--1000 & 0.099270\\
 &  & > 1000 & 0.003080\\
 \cmidrule(lr){2-4}
 
  & \multirow{6}{*}{TxT360-BestOfWeb}
    & 1--1 & 5.722522\\
 &  & 2--5 & 9.394343\\
 &  & 6--10 & 5.072045\\
 &  & 11--100 & 6.801969\\
 &  & 101--1000 & 0.216884\\
 &  & > 1000 & 0.004300\\

 \midrule
 \multirow{6}{*}{\texttt{Synthetic QA}}
  & \multirow{6}{*}{TxT360-QA}
    & 1--1 & 3.775652\\
 &  & 2--5 & 6.426003\\
 &  & 6--10 & 2.361838\\
 &  & 11--100 & 3.165823\\
 &  & 101--1000 & 0.098637\\
 &  & > 1000 & 0.001932\\

 \midrule
 \multirow{2}{*}{\texttt{Papers}}
   & \multirow{2}{*}{ArXiv, PubMed, S2ORC, and PhilPapers}
    & 1--1 & 3.835154\\
 &  & others & 0.575500\\

\midrule
\multirow{4}{*}{\texttt{Math}}
  & MegaMath Code & all & 0.174410\\
  & MegaMath Web & all & 1.734292\\
   \cmidrule(lr){2-4}
  & DM Math & all & 0.105578\\
   \cmidrule(lr){2-4}
  & \parbox{6cm}{Other math (InfiMM-WebMath-40B, \\ OpenWebMath, Textbooks-MathPile)} & all & 0.977156\\

\midrule
\multirow{3}{*}{\texttt{Code}}
  & RefineCode & all & 7.102195\\
  & RefineCode FIM & all & 6.535289\\
  & RefineCode Topo-Sorted & all & 0.581025\\

\midrule
\multirow{2}{*}{\texttt{Multilingual}}
  & JAIS Arabic Data & all & 2.137719\\
   \cmidrule(lr){2-4}
  & Europarl-Aligned & all & 0.010014\\
\midrule
 
\multirow{15}{*}{\texttt{Other}}
     & \multirow{2}{*}{Wikipedia}
      & 1-1 & 0.893301\\
  & & others & 0.005294\\
\cmidrule(lr){2-4}
  & Wikipedia-Extended & all & 0.876997\\
 \cmidrule(lr){2-4}
  & \multirow{2}{*}{Europarl}
    & 1--1 & 0.030053\\
 &  & others & 0.000439\\
 \cmidrule(lr){2-4}
 
  & \multirow{2}{*}{FreeLaw}
      & 1--1 & 0.297616\\
  & & others & 0.168791\\
\cmidrule(lr){2-4}
 
  & \multirow{2}{*}{Hackernnews}
    & 1--1 & 0.023127\\
 &  & others & 0.000079\\
\cmidrule(lr){2-4}
 
  & \multirow{2}{*}{PG-19}
      & 1--1 & 0.033099\\
  & & others & 0.045906\\
\cmidrule(lr){2-4}
 
  & \multirow{2}{*}{StackExchange}
      & 1--1 & 0.475941\\
  & & others & 0.010132\\
\cmidrule(lr){2-4}
 
  & Ubuntu IRC & 1-1 & 0.049009\\
\cmidrule(lr){2-4}
 
  & \multirow{2}{*}{USPTO}
      & 1-1 & 0.111890\\
  & & others & 0.000680\\
\bottomrule
\end{tabular}
\caption{Pre-training data mix details. Documents are organized in ``Duplication Bucket'' based on number of duplicates they have. The bucket is used as one of the features to determine the mixing weights. \label{tab:pretrain-data-mix-detail}}
\end{table}

\paragraph{Analysis of Near-Duplicate Clusters.}

Analysis of the near-duplicate clusters reveals distinct patterns between the component sizes. Smaller clusters tend to exhibit greater overlap in their \texttt{MinHash} bands. The smallest components are typically near-exact duplicates—documents that differ only slightly due to minor formatting or textual variations—and were therefore not captured during the earlier stage of local exact deduplication.

We observe distinct patterns based on cluster size. Smaller clusters typically reflect the maintenance of a living document where content, such as a staff list, is modified incrementally over the years. However, the pattern shifts toward mass replication for clusters with 1,000 or more documents. These large clusters usually contain generic templated text, including disclaimers or terms and conditions, which commercial entities use to standardize legal and operational messaging across platforms.

\subsubsection{Data mix}
\label{sec:pretraining-mix}
\begin{figure}
    \centering
    \includegraphics[width=0.45\linewidth]{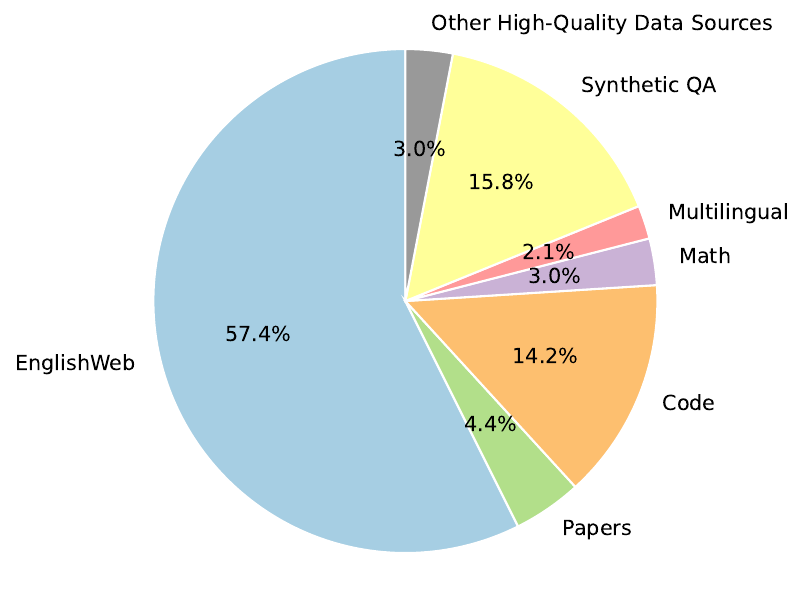}
    \caption{Overview of the domain distribution of the pre-training data mix. Majority comes from English Web.}
    \label{fig:pretrain-data-mix}
\end{figure}

Following ~\citet{txt360data2024}, we upsample documents based on their natural distribution of duplicates counts. However, since duplication is only an indirect indicator of quality, we upsample documents to a few predefined levels rather than using their exact count. Specifically, we set the upsampling weight to 3 for documents with 2 to 5 duplicates, 5 for those with 5 to 100 duplicates, 8 for 101 to 1000 duplicates, and 10 for documents with over 1000 duplicates. These values are selected heuristically and informed by preliminary small-scale experiments. For non-CommonCrawl data sources, we assign a weight of 2 if the document appears more than once. A high level domain distribution is depicted in Figure~\ref{fig:pretrain-data-mix}. The detailed data distribution is shown in Table~\ref{tab:pretrain-data-mix-detail}.

\subsection{Pre-training Infrastructure}\label{sec:pretrain_infra}
\begin{keytakeaways}
    \begin{itemize}
        \item \textbf{Parallelism Tradoff.} With sufficient GPU memory and a small global batch size, one should consider disabling pipeline parallelism to achieve higher training throughput.
        \item \textbf{Fault Tolerance System.} Our fault tolerance and auto-mitigation systems allow our model to quickly recover from training incidents.
    \end{itemize}
\end{keytakeaways}

To scale training for our largest models, we use hybrid parallelism, which combines four parallelism methods to shard the model. This approach efficiently distributes computation across many GPUs and ensures that each GPU can hold the model parameters, optimizer states, gradients, and activations in its HBM. In \ktwo{}, we use a combination of tensor parallelism (TP), sequence parallelism (SP), and data parallelism (DP). Our pretraining framework is based on Megatron Core~\citep{shoeybi2019megatron} and uses Transformer Engine as its backend~\citep{NVIDIATransformerEngine}.

\subsubsection{Hardware and Cluster Setup}
The \ktwo{} model is trained on an H200 GPU cluster, with each server equipped with eight NVIDIA H200 GPUs operating at 700 W TDP and 141 GB of memory. Each GPU node provides eight InfiniBand HCAs running at 400 Gb/s in a fully non-blocking fabric. To improve fault-tolerance and reliability of the training job, we develop an in-house tool, \texttt{submitter}, which continuously detects hardware and system faults and enables automatic fault-tolerant job submission, termination, and recovery over Slurm.

\subsubsection{Parallelism and Training Infrastructure}\label{sec:parallel}

\paragraph{Trade Offs Among Parallelism Strategies.}
Similar to what we discussed in~\citep{K2}, DP partitions the input data batch evenly among workers. At each iteration, each worker computes gradients over its assigned batch, and these gradients are synchronized among workers before the next iteration. DP requires each worker to hold a complete model replica, so it cannot be used to train models with very large parameter counts. Fully Sharded Data Parallelism (FSDP) can replace the DP dimension to further save GPU memory footprint at the cost of marginal communication overhead, but when we set up the training system, Megatron Core did not have full support for FSDP~\citep{zhao2023pytorch}. In TP, the weights of two consecutive layers are partitioned first along the row dimension (input dimension) and then along the column dimension (output dimension)~\citep{shoeybi2019megatron}. TP removes the need for synchronizing the intermediate output of the first layer but introduces heavy cross device communication afterward. SP can be applied along the TP dimension to further reduce GPU memory consumption, although it increases communication~\citep{korthikanti2023reducing}. In PP, layers are placed across GPUs, and the mini batch is split into microbatches. These microbatches are then pipelined in the forward and backward passes. PP requires less communication than DP and TP but suffers from idle GPU time, for example pipeline bubbles~\citep{huang2019gpipe,Narayanan2019PipeDreamGP}.

\paragraph{Heuristics for Tuning Hybrid Parallelism Strategy.}
Similar to what we discussed in~\cite{K2}, a good parallelism strategy follows the pattern in which TP is mapped to intra-node GPUs, PP and its associated microbatch sizes are then tuned to satisfy GPU memory constraints, and all remaining GPU degrees are assigned to DP (or FSDP). In \ktwo{}, we start from the K2-V1 parallelism strategy. Compared with prior work, we now use H200 which has larger GPU memory (141GB). \textit{Given larger GPU memory capacity, which parallelism dimension should be removed first?} Intuitively, TP should be reduced before PP, since its communication cost is much higher compared to PP. However, in practice, because TP can be used together with SP, its memory savings are much more significant than those from PP. This is especially true for \ktwo{}, where we pre-trained using an 8K context length, making SP essential. Another reason to disable PP rather than TP is that we use a relatively small global batch size, 
which results in a per-GPU batch size of $3$. Such a small per-GPU batch size almost always causes a large fraction of pipeline bubbles, which hurts utilization. Thus, for \ktwo{} pretraining, we disable PP and use only TP, SP, and DP. To mitigate GPU memory spikes, we also enable activation recomputation~\citep{korthikanti2023reducing}.

\paragraph{The Optimized Hybrid Parallelism Strategy.}
We use a carefully tuned strategy that combines DP, TP, and SP for \ktwo{} pretraining. More specifically, we use 8-way TP (with SP enabled along the TP dimension). The tuning procedure follows the heuristics aforementioned, as discussed above. In stage 1, the global batch size is 1200, and we use a mini batch size of 3 per DP group (i.e., no microbatching is used). We apply activation recomputation to reduce memory consumption, using full-layer activation checkpointing uniformly across layers and treating each transformer layer as the recompute unit (using \texttt{--recompute-granularity full}, \texttt{--recompute-method uniform}, and \texttt{--recompute-num-layers 1} in Megatron-LM). BF16 precision and FlashAttention 2 are enabled to increase training speed~\citep{dao2023flashattention}.

\subsection{The Pre-training Run}\label{sec:run}
\begin{keytakeaways}
\begin{itemize}
    \item \textbf{Intervene based on spike width, not just height.}
    Restart-based interventions incur computational costs. We therefore distinguish between \textit{narrow spikes} (transient instabilities that often self-correct) and \textit{wide, malignant spikes} (which indicate diverging dynamics). We implement automated rollbacks only for the latter to avoid unnecessary restart overhead.
    \item \textbf{Expect and tune for early-phase instability.}
    Instability is structurally concentrated in the first 30\% of training. Tuning the optimizer timescale ($\tau_{\text{epoch}}$) specifically for this phase can be helpful.
    \item \textbf{Extremely low learning rates stall progress.} While Decay2Zero is effective in some settings, we find that learning rates below $1.5 \times 10^{-6}$ may have hit numerical precision limits. We observe the parameter norm ceases changing and may have stopped learning.
\end{itemize}
\end{keytakeaways}

\subsubsection{The Effect of Decay2Zero}\label{sec:d2z_run}
\begin{figure}[h!]
    \centering
    \includegraphics[width=0.5\linewidth]{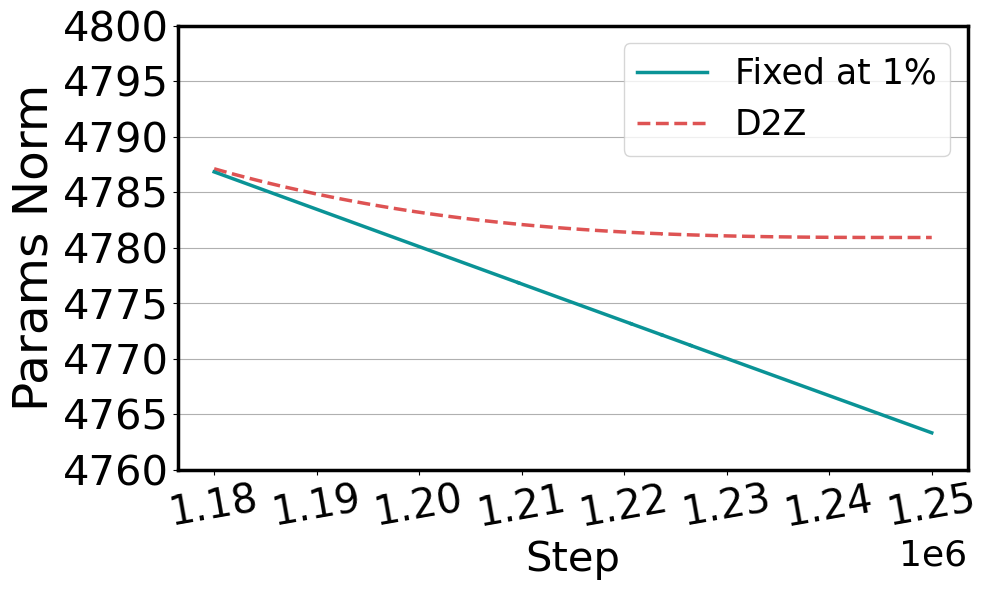}
    \caption{Params norm comparison: D2Z vs constant 1\% for the tail. We see that the params norm for D2Z flattens, while that fixed at 1\% in the tail continues to drift downward.}
    \label{fig:params_norm_comparison}
\end{figure}

Decay2Zero learning rate schedules are strong both for the full schedule \citep{bergsma2025straight} and in a short additional annealing phase at the end \citep{groeneveld2024olmo}. However, few results have been published at \ktwo{}'s scale (70B and 10x Chinchilla over-trained regime). In our actual run, we observe that when the learning rate drops below $1.5 \times 10^{-6}$, the parameter norm ceases to change—a symptom we hypothesize to indicate training stagnation. Figure~\ref{fig:params_norm_comparison} shows this in detail. From steps 1.23e6 to 1.25e6, the parameter norm change of the D2Z approach slows down significantly, while that where it is fixed at 1\% continues to decrease. Since we plan to conduct subsequent mid-training stages, we decide to use a non-zero learning rate to conclude pre-training.

\subsubsection{Training monitoring}\label{section:monitoring}

We develop an automated monitoring system to ensure stability and reliability during training. The system continuously tracks key metrics and:
\begin{itemize}
\item Detects abnormal events, such as significant spikes in the training loss curve.
\item Automatically rolls back to the most recent stable checkpoint, relaunches the job using the \texttt{submitter} tool discussed previously, and backs up relevant checkpoints.
\item Sends notifications for both detected anomalies and recovery actions taken.

\end{itemize}

\paragraph{Abnormal events detection.}
To ensure stable long-horizon training, we implement an abnormal events detection system designed to identify and react to severe loss spikes as early as possible. We adopt a deterministic, dual-threshold, sliding-window algorithm that distinguishes persistent divergence from transient noise.

\begin{itemize}
\item \( w \): width of the historical window (persistence width).
\item \( v_{\min} \): minimum value of the monitored metric within the window.
\item \( v_{\max} \): maximum value of the monitored metric within the window.
\end{itemize}

At every training iteration \( t \), the system examines the most recent \( w \) iterations and evaluates the loss trajectory against two scalar thresholds: a \textbf{sustained floor} \( T_{\min} \) and a \textbf{severity peak} \( T_{\max} \). An abnormal event, as known as a loss spike, is triggered if and only if the following conditions hold simultaneously:

\begin{itemize}
\item \textbf{Persistence condition:} \( v_{\min}(t) > T_{\min} \).  
This ensures the loss has not re-entered a safe range for the entire duration of the window, filtering out single-step fluctuations.

\item \textbf{Severity condition:} \( v_{\max}(t) > T_{\max} \).  
This ensures the loss has reached a critical magnitude, preventing false positives caused by high-but-stable plateaus.
\end{itemize}

We operate the detector in two sensitivity tiers. The \textbf{alert tier} uses a narrower window and lower thresholds to identify early warning signs and triggers a level-one Slack notification. The \textbf{restart tier} uses a wider window and stricter thresholds to confirm sustained divergence; exceeding these thresholds triggers a level-two Slack notification with automatic restart of the training job.

\begin{figure}
    \centering
    \includegraphics[width=0.45\linewidth]{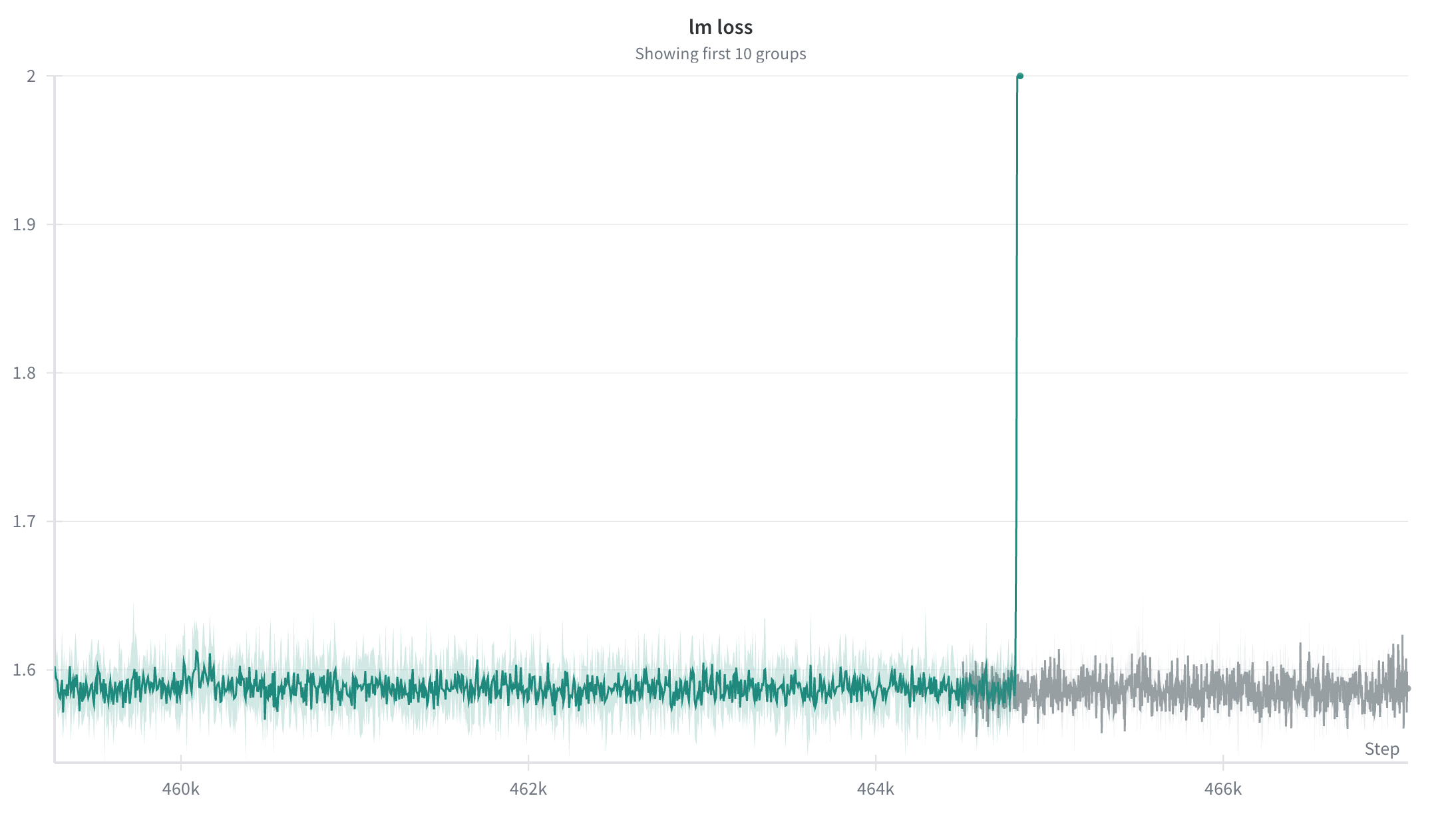}\includegraphics[width=0.5\linewidth]{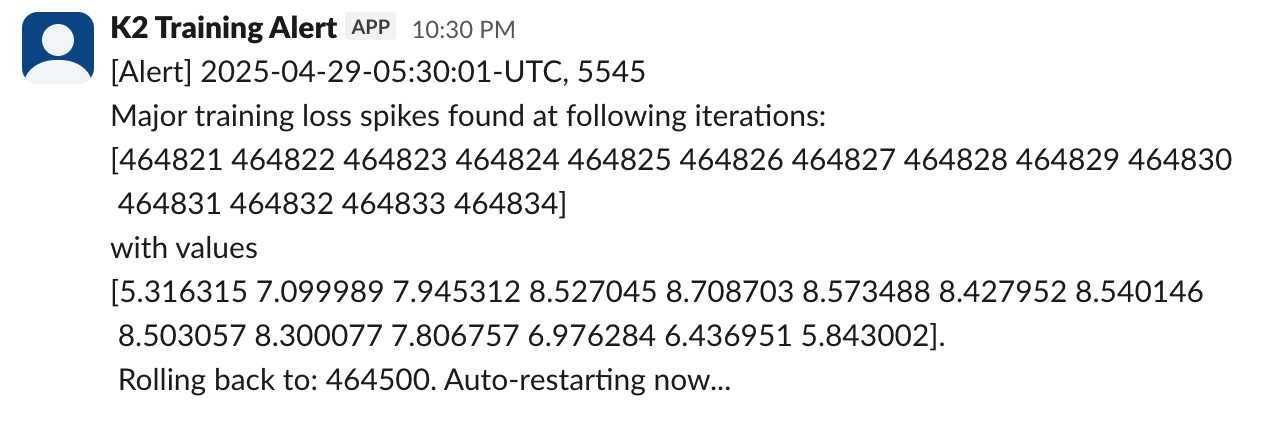}
    \caption{a) A severe loss spike appeared at around 464,000 step. b) The spike was immediately detected. The job was auto-restarted and a notification was sent through Slack}
    \label{fig:bot-notification}
\end{figure}

When a restart-tier event is detected at iteration \( t_{\text{spike}} \), the system rolls back to the most recent fully committed checkpoint based on a fixed checkpoint interval \( I \). The rollback iteration is computed as:
\[
t_{\text{rollback}} = \left\lfloor \frac{t_{\text{spike}}}{I} \right\rfloor \times I,
\]
ensuring recovery always begins from a known-good checkpoint on disk. Figure~\ref{fig:bot-notification} shows an example of the notification mechanism in action.

\subsubsection{Training Dynamics \& Stability}\label{sec:dynamics}

In this section, we present training dynamics observations. We focus on three key signals: loss spikes, gradient norms, and parameter norms. Monitoring these metrics allows us to detect potential instabilities or irregularities during the run. The following analysis details how these values evolved from the start of pre-training to the final steps.

\paragraph{Loss Spikes.}
Loss spikes \citep{molybog2023theory,takase2025spike} are sharp jumps in training loss that interrupt an otherwise stable loss curve. In Large Language Models, they often stem from a mix of aggressive learning-rate schedules and first or second-moment estimator lag in Adam-style optimizers.

To help visualize the dynamics, we quantify spikes using a robust local z-score that measures how abnormal the current loss is relative to its recent history. For loss values $y_t$ over training steps $t$, we compute a rolling median $m_t$ over a trailing window $W(t)$ (set to approximately 1\% of the full training duration). To robustly estimate local variability while accounting for the non-stationary trend of the loss curve, we compute the median of the absolute deviations from the rolling median within the window:
\begin{equation}
    \mathrm{MAD}_t = \operatorname{median}_{s \in W(t)} \bigl|\, y_s - m_s \,\bigr|,
\end{equation}
where $m_s$ is the rolling median at step $s$. We then define the local z-score $z_t$ as:
\begin{equation}
    z_t = \frac{y_t - m_t}{1.4826\,\mathrm{MAD}_t}.
\end{equation}
Points with $z_t > 5$ are flagged as loss spikes. Intuitively, this score measures how many typical fluctuations above the local baseline a specific step represents, capturing abrupt deviations while adapting to the non-stationarity of the loss curve. This formulation is the loss-domain, locally adaptive analogue of the Gradient Spike Score (GSS) introduced in \citet{huang2025spam}, which applies a similar standardized score to gradient norms.

To comprehensively evaluate training stability, we analyze both the final successful model trajectory (the ``clean'' run) and the divergent branches that were abandoned during training. We reconstruct the clean run by stitching together the chain of checkpoints that led to the final model. The discarded segments are then treated as candidate failure events. We filter these abandoned runs, retaining only those that exhibit a significant deviation from the clean run's baseline prior to termination. In our analysis, both transient spikes within the clean run and fatal spikes in the abandoned runs are evaluated against the robust statistics ($m_t$ and $\mathrm{MAD}_t$) derived from the successful trajectory, ensuring a consistent definition of anomaly across all data.

\begin{figure}
    \centering
    \includegraphics[width=0.45\linewidth]{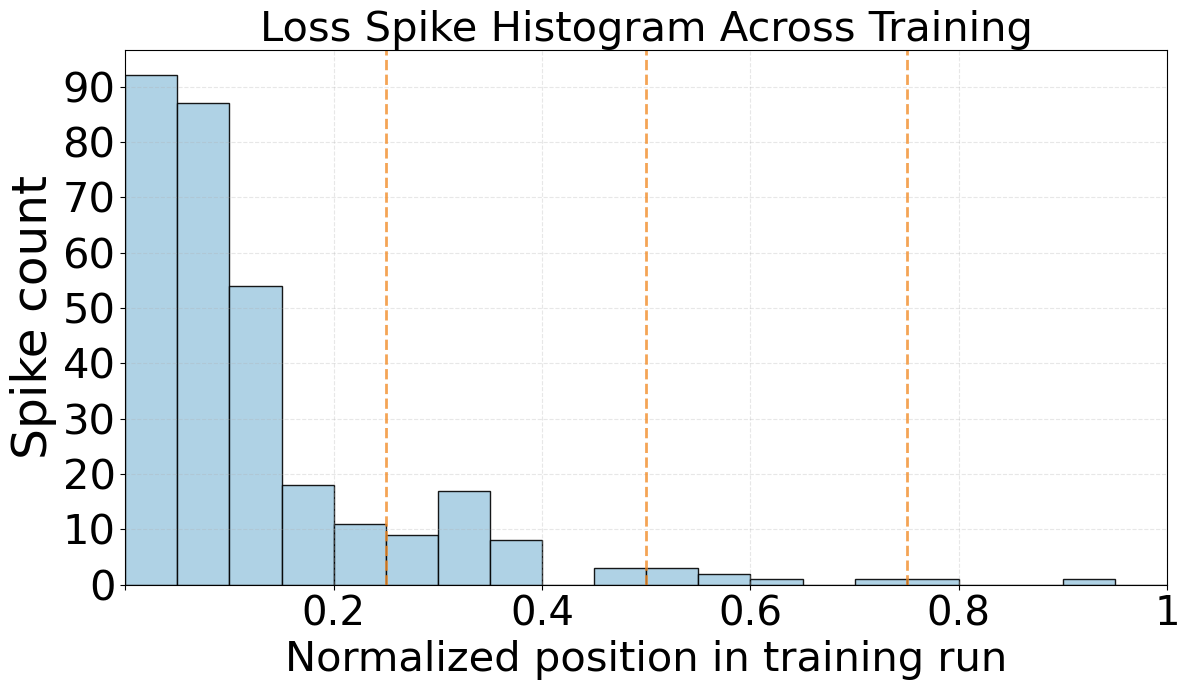}\includegraphics[width=0.45\linewidth]{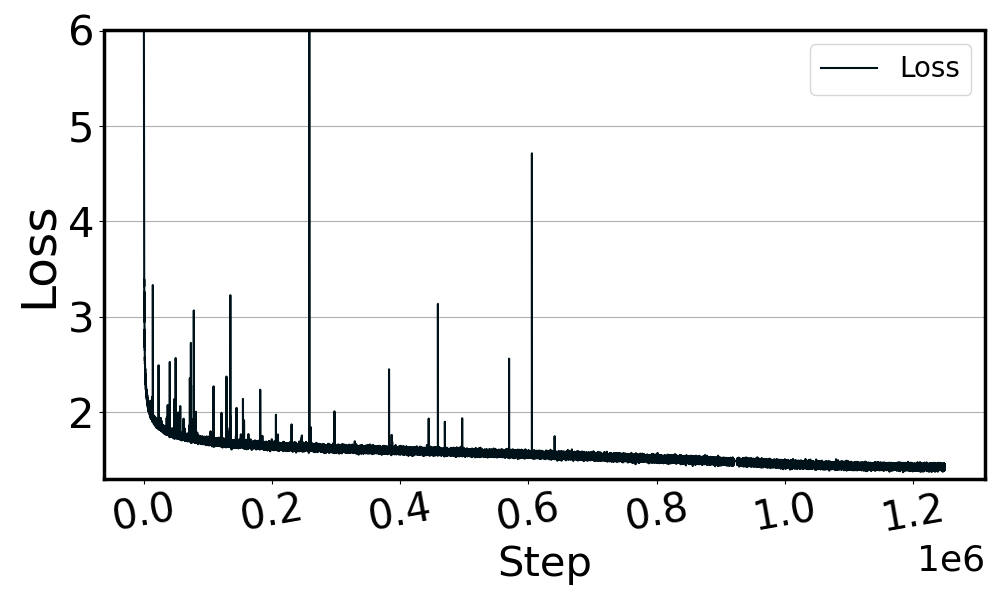}
    \caption{a) Histogram of Loss Spikes (including those erased via restart) across pre-training. We see that the spikes are highly concentrated in the first 40\% of training. See Section \ref{sec:dynamics} for our definition of loss spikes. b) Pre-training loss curve (excluding regions erased via restart). Again, we see that transient spikes are highly concentrated in the first 40\% of training.}
    \label{fig:loss-spike-histogram}
\end{figure}

Figure \ref{fig:loss-spike-histogram} illustrates that the vast majority of loss spikes occur during the first 40\% of training, with a particularly high frequency in the first 20\%. We address these instabilities via two mechanisms. First, we observe that poorly tuned $\tau_{\text{epoch}}$ values \citep{wang2025how} exacerbate spike frequency at small scale; we therefore explicitly tuned this hyperparameter to 0.1066 (see Section \ref{sec:training-dynamics}). Second, as described in Section \ref{section:monitoring}, we intervene by rolling back the model state only when observing wide, malignant spikes \citep{liu2025llm360}. Figure \ref{fig:narrow-wide-spikes}a illustrates a narrow spike where training was allowed to continue, whereas Figure \ref{fig:narrow-wide-spikes}b shows a cluster of wide spikes that triggered a rollback.

\begin{figure}
    \centering
    \includegraphics[width=0.45\linewidth]{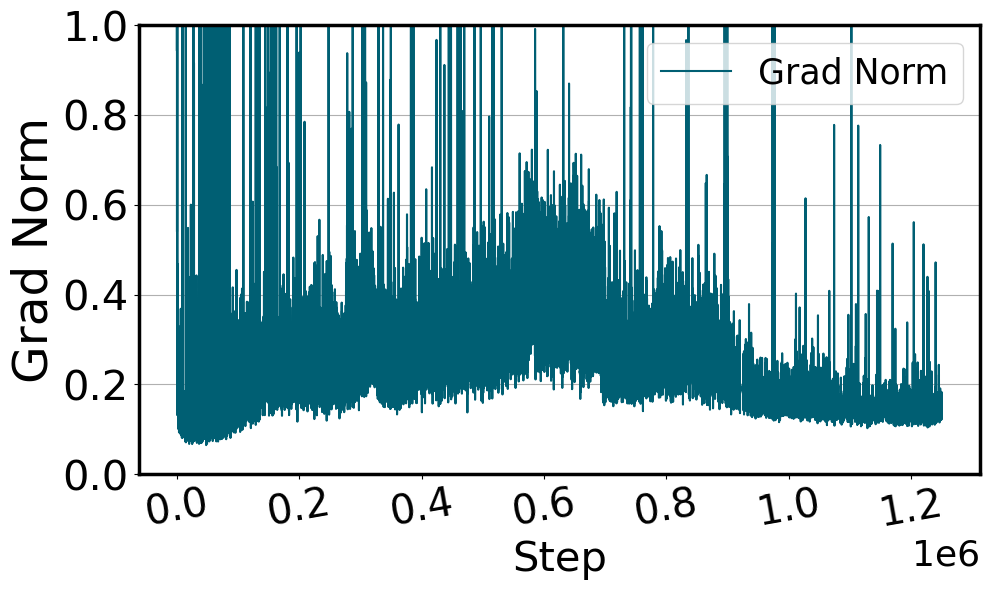}\includegraphics[width=0.45\linewidth]{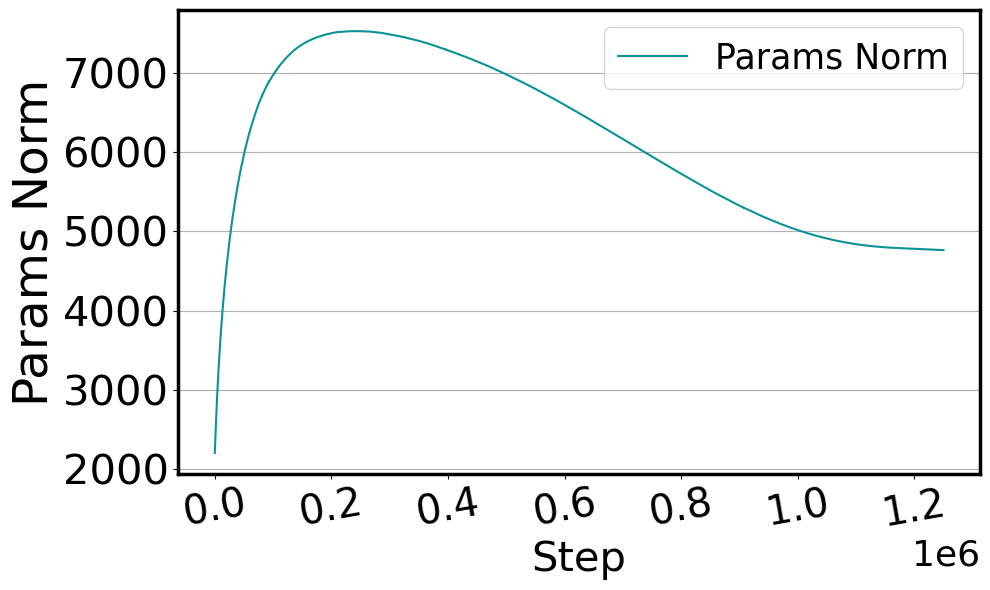}
    \caption{a) Grad norm throughout pre-training (y-axis on log scale). We see that initially the norms are low, then they drift up, peak during the middle of training and then drift back down. b) Params norm throughout pre-training. They rise initially before settling. This matches the behavior of AdamW in \cite{wen2025fantastic} Figure 6 (middle left).
    }
    \label{fig:grad-norm}
\end{figure}

\paragraph{Gradient and Parameter Norms.}
As shown in Figure \ref{fig:grad-norm}a, gradient norm spikes follow a similar temporal distribution to loss spikes, appearing more frequently in the first half of pre-training. Ignoring transient spikes, the gradient norm tends to drift upward, peaking halfway through training before decaying, though it remains elevated compared to initialization. Interestingly, the decay we observe is \textit{not} standard behavior, which involves gradient norms increasing towards the end of training. This is discussed in detail including theoretical analysis in \cite{defazio2025gradients}, and Figure 6 of \cite{wen2025fantastic} show a similar phenomena. Future investigation of our grad norm behavior is thus warranted. Figure \ref{fig:grad-norm}b illustrates the parameter norm throughout training. After peaking near step 200k, the norm drifts downward. This decline slows in the final stages, aligning with the learning rate decay. This pattern matches the behavior of over-trained (as per Chichilla scaling law~\citep{hoffmann2022trainingcomputeoptimallargelanguage}) AdamW observed by \cite{wen2025fantastic}.

\begin{figure}
    \centering
    \includegraphics[width=0.45\linewidth]{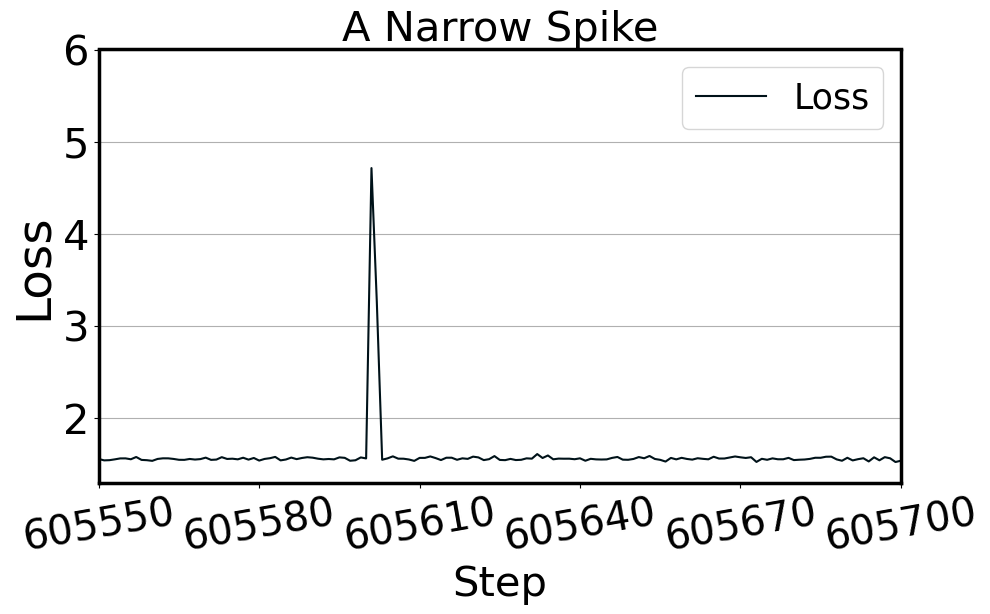}\includegraphics[width=0.45\linewidth]{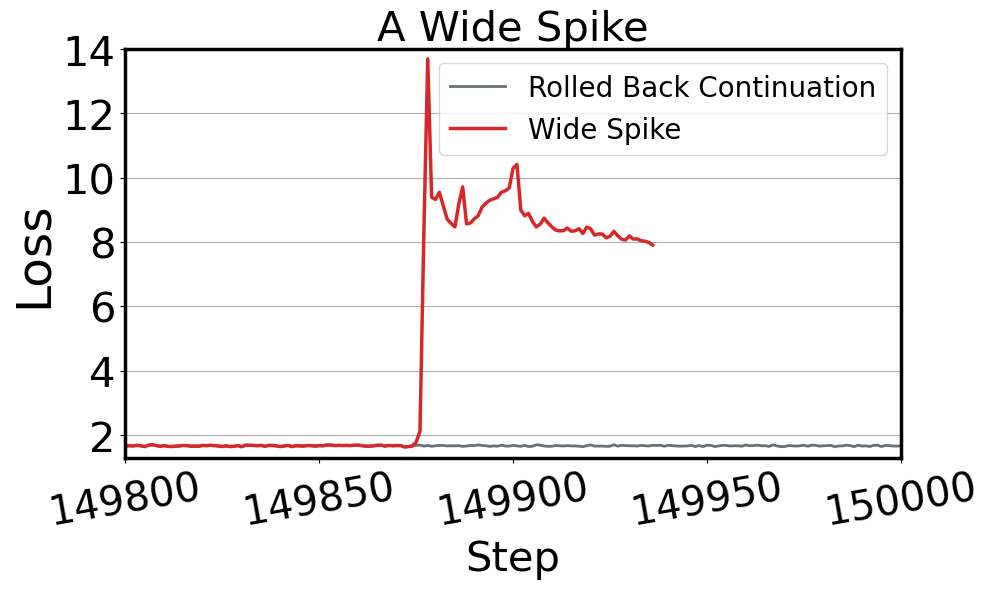}
    \caption{a) A narrow spike. We do not roll back if it does not meet the pre-defined threshold. b) Wide spikes that meet the threshold are rolled back, as we have observed that they sometimes hurt downstream task performance.}
    \label{fig:narrow-wide-spikes}
\end{figure}

\section{Mid-training}\label{sec:mid-training}

\begin{keytakeaways}
    \begin{itemize}
    \item \textbf{Training framework efficiency.} We implement features like on-the-fly tokenization and best-fit packing in our in-house training system to improve flexibility and throughput.

    \item \textbf{Selecting appropriate RoPE configuration is critical.} The choice of base frequency for Rotary Positional Embeddings (RoPE) is the determining factor for successful long-context extension.

    \item \textbf{Balancing reasoning with general capabilities.} Mid-training on synthetic thinking traces drastically alters model behavior. We find that a careful ratio of natural web data to reasoning data is required to trigger these new capabilities without degrading performance on classical pre-training benchmarks.

    \item \textbf{Mid-training enables test-time scaling.} Incorporating reasoning data allows the base model to demonstrate strong test-time scaling even before post-training. With a simple reasoning-inducing prompt, we achieve nearly 50\% accuracy on AIME 2025 and over 90\% on MATH-500, significantly outperforming similar open-source base models.
    \end{itemize}
\end{keytakeaways}

\subsection{Mid-training Recipe}\label{sec:mid-train-recipe}

\ktwo{} uses a progressive context-length expansion strategy to extend the training context up to 512K tokens. Specifically, we employ a four-stage training procedure to systematically up-sample long-context data across diverse length ranges, while preserving performance on short contexts. 
The precision  of BFloat16 is used throughout the mid-training process.
The details of training length, training data size and RoPE base frequency and learning rate are shown in \autoref{tab:mid-training}. 

\begin{itemize}
    \item \texttt{mid-1}: In the first stage, we aim to shift from pre-training toward higher-quality data while beginning to introduce reasoning behaviors and thinking traces. We maintain the same context length of 8192 tokens and a RoPE base frequency of 0.5M as in the pre-training stage. During this stage, given the relatively large-scale training corpus of 1.7T tokens, we adopt a cosine learning rate schedule, peaking at \num{1.5e-5} and gradually decaying to \num{6e-6}. The model is trained on NVIDIA H200 GPUs, with a global batch size $13{,}107{,}200$ tokens.
    
    \item \texttt{mid-2, mid-3, mid-4}: In the subsequent three stages, the context lengths are expanded to 64K, 128K, and 512K tokens. To ensure stable optimization when extending to long context, we keep a constant learning rate of \num{6e-6}, with the RoPE base frequencies set to 1M, 10M, and 10M, respectively. Changing the RoPE frequency base to a larger value significantly improves long-context performance~\citep{prolong, xiong-longcontext}. We also observe that scaling up the RoPE base to 10M is particularly useful when context lengths exceeds 128K tokens. To avoid short-context performance degradation, we consistently mix at least 30\% short-context data at each stage. Detailed context mix is presented in \autoref{fig:mid-training-context-pies}. We use the same global batch size ($13,107,200$ tokens) as the first stage while halving the node set. In \texttt{mid-3, mid-4}, we apply context parallelism technique to compute full attention upon long sequences. 
\end{itemize}

\begin{table}[t]
\centering
\resizebox{.7\columnwidth}{!}{
\begin{tabular}{@{}lcccccccc@{}}
\toprule
\textbf{Stage} & \textbf{Context} & \textbf{\# Tokens} &
\textbf{RoPE} & \textbf{Max LR} & \textbf{Min LR} & \textbf{LR Decay} & \textbf{CP} \\
\midrule
mid-1 & $8{,}192$   & 1769B & 0.5M & \num{1.5e-5} & \num{6e-6} & Cosine   & 1 \\
mid-2 & $65{,}536$  & 590B  & 1M   & \num{6e-6}   & \num{6e-6} & Constant & 1 \\
mid-3 & $131{,}072$ & 229B  & 10M  & \num{6e-6}   & \num{6e-6} & Constant & 2 \\
mid-4 & $524{,}288$ & 131B  & 10M  & \num{6e-6}   & \num{6e-6} & Constant & 8 \\
\bottomrule
\end{tabular}%
}
\caption{Hyper-parameters in different mid-training stages. CP is the size of context parallelism.}
\label{tab:mid-training}
\end{table}

\subsection{Mid-training Data: TxT360-Midas}\label{sec:mid-train-data}

In the mid-training phase, we increase our reliance on synthetic data to address specific limitations in natural text. While natural data provides a strong foundation, it lacks both the density required for complex reasoning and the length needed for long-context training. Synthetic generation allows us to construct these missing pieces, producing extended, consistent and logical rich documents that are rarely found in the wild. We curate the \texttt{TxT360-Midas} dataset for this purpose, with the following design goals:

\begin{itemize}
    \item Extend model's context length to 512k. We perform four stages of training with progressively increasing context length on high-quality mixes of synthetic and natural text data organized according to their token length.
    \item Prepare model for post-training by introducing synthetic reasoning data to facilitate improved performance via increased test-time computation. In addition to collecting LLM thinking traces across domains, we develop new synthetic data pipelines transforming natural user queries into over 100 reasoning behavior demonstrations such as System 1 and System 2 thinking, Red Team vs Blue Team thought experiment, detective-style abductive reasoning, and many others. 
    \item Compensate web data limitations by introducing higher amount of synthetic data, especially data targeting mathematical ability. We collect over 250 million mathematical problems and generate solutions with thinking from open-source models.
\end{itemize}

\subsubsection{LLM Thinking Traces}

Enabling LLMs to utilize test-time computation by means of generating ``thinking'' tokens before providing the final answer has become a key factor in rapid performance improvement of recent open- and closed-source models \citep{jaech2024openai,guo2025deepseek,yang2025qwen3}. \citet{wang2025octothinker} demonstrated importance of including such thinking traces in mid-training to incentivize reinforcement learning.

To create sufficiently large corpora of thinking traces we collect over 250 million unique (based on exact deduplication) mathematical problems from various sources. These sources are drawn from the mathematical datasets with permissive licenses studied in the OpenThoughts project \citep{guha2025openthoughtsdatarecipesreasoning}\footnote{\url{https://www.open-thoughts.ai/blog/thinkagain}}.
To avoid contamination we only use train subsets where relevant. Problems include templated arithmetic and numeric tasks (for example, checking if a large number is prime), verification of mathematical statements, and other math problems of varying difficulty. For each problem we generate solutions using Qwen3-32B with thinking \citep{yang2025qwen3} and GPT-OSS-120B with high reasoning effort \citep{agarwal2025gpt}. We augment the data with Nemotron-Postraining-Dataset-v1 \citep{bercovich2025llamanemotronefficientreasoningmodels,NemotronPostTrainingDatasetV1} which, in addition to math, covers coding and STEM problem-solving and utilizes DeepSeek-R1 \citep{guo2025deepseek} to generate solutions with thinking traces. We refer to this data source as \texttt{thinking\_traces}.

\subsubsection{Synthetic Reasoning Behaviors}
Our pipeline for generating synthetic data that demonstrates diverse \texttt{reasoning\_behaviors} consists of two key components: natural user queries \citep{zheng2023lmsys,zhao2024wildchat,don2025sharelm} and prompt templates for over 100 reasoning behaviors. 

Our goal in exposing the model to a large number of reasoning behaviors across many contexts (including ones that may seem only loosely related to reasoning) is to encourage behavioral diversity and to elicit reasoning in domains where it is difficult to obtain verifiable problems for reinforcement learning. Our approach is motivated by \citet{gandhi2025cognitive}, who demonstrated the importance of four reasoning types in mid-training in the context of RL training on logic puzzles such as Countdown. By mid-training on many reasoning behaviors across domains, we better prepare the model for subsequent reasoning improvements, such as RLVR style training.

We generate additional data with this pipeline using behaviors outside of reasoning. For example data-science related discussion, various forms of planning, and other categories like user manuals. As with reasoning behaviors, we utilize natural user queries as ``inspiration'' to achieve diversity at scale and vary the behavior prompt. We refer to this data as \texttt{other\_behaviors}. Both reasoning and other behaviors were generated with Qwen2.5 32B Instruct \citep{qwen2.5} and Qwen3 32B with non-thinking mode \citep{yang2025qwen3}.

An example of the \texttt{dual\_process\_reasoning\_system} behavior is shown below:

\begin{figure}[H]
\begin{tcolorbox}[
    colback=white,
    colframe=black,
    boxrule=2pt,
    arc=3mm,
    fonttitle=\bfseries\small,
    title=dual\_process\_reasoning\_system,
    coltitle=white,
    colbacktitle=black
]
{\bf \small Prompt:} 
\begin{lstlisting}[breaklines=true,breakautoindent=false,
  breakindent=0pt,basicstyle=\footnotesize\ttfamily]
Generate reasoning about the following problem:

{query}

Begin by restating the problem. First provide an intuitive (System 1) assessment, then transition to a deliberate (System 2) analysis. Show how these two reasoning modes lead to different conclusions and how they can be reconciled.
\end{lstlisting}
\end{tcolorbox}
\end{figure}

Corresponding synthetic snippet for ``How do I have an epic water fight?'' query:
\begin{figure}[H]
\begin{tcolorbox}[
    colback=white,
    colframe=black,
    boxrule=2pt,
    arc=3mm,
    fonttitle=\bfseries\small,
    title=How do I have an epic water fight?,
    coltitle=white,
    colbacktitle=black
]
{\bf \small Response:} 
\begin{lstlisting}[breaklines=true,breakautoindent=false,
  breakindent=0pt,basicstyle=\footnotesize\ttfamily]
The problem at hand is about how to have an epic water fight. Let's break this down by first using intuitive reasoning, and then transitioning to a more deliberate analysis.

Intuitively (System 1), an epic water fight might seem like a simple idea to execute. This could involve grabbing as many water guns as possible, filling them up at a nearby hose, and then running around with friends or family, spraying each other from various angles [...]

However, adopting a more deliberate (System 2) approach to the problem reveals additional considerations that could elevate a water fight to being truly epic. This involves more structured planning and execution:

1. **Preparation:** Consider the type and quantity of water guns, ensuring there's a mix of short-range and long-range weapons. Additionally, [...]

2. **Location and Setup:** The setting can make a significant difference. A spacious backyard with varied terrain (trees, a pool, or a trampoline) can add to the excitement [...]

3. **Rules and Teams:** [...]

4. **Safety and Supervision:** [...]

5. **Aftermath:** [...]

Reconciling the intuitive and deliberate approaches, the key is to balance spontaneity with planning. While the immediate joy and excitement of a water fight are important, thoughtful preparation can enhance the experience [...]
\end{lstlisting}
\end{tcolorbox}
\end{figure}












\subsubsection{Data mixes}

In our data mixes we combine the aforementioned synthetic datasets with natural text:
\begin{itemize}
    \item \texttt{Web}: select subsets of our pre-training data (Table \ref{tab:pretraining_data}): \texttt{CC-edu} (high-quality subset of CommonCrawl chosen using FineWeb-edu classifier \citep{lozhkov2024fineweb-edu}); \texttt{Wiki+} (combined Wikipedia, HackerNews, and StackExchange sources); \texttt{Legal} (combined Europarl, USPTO, FreeLaw sources); \texttt{Papers+} (combined Papers, PG-19, Ubuntu IRC sources); \texttt{TxT360-QA} (Section \ref{sec:txt360-qa}); \texttt{Arabic} (Section \ref{sec:multilingual}); \texttt{MegaMath} \citep{zhou2025megamath}. We additionally augment it with a few sources from Common Pile that were not present in our pre-training data \citep{kandpal2025common} (e.g., openly licensed transcribed YouTube videos).
    \item \texttt{Code}: \texttt{RefineCode} \citep{huang2025opencoder} data from pre-training augmented with \texttt{Stack-Edu} \citep{allal2025smollm2smolgoesbig} Python subset.
    \item \texttt{Institutional Books} \citep{cargnelutti2025institutionalbooks10242b}, large collection of public domain books providing bulk of long-context data.
\end{itemize}

\begin{figure}[t]
    \centering
    \includegraphics[width=\linewidth]{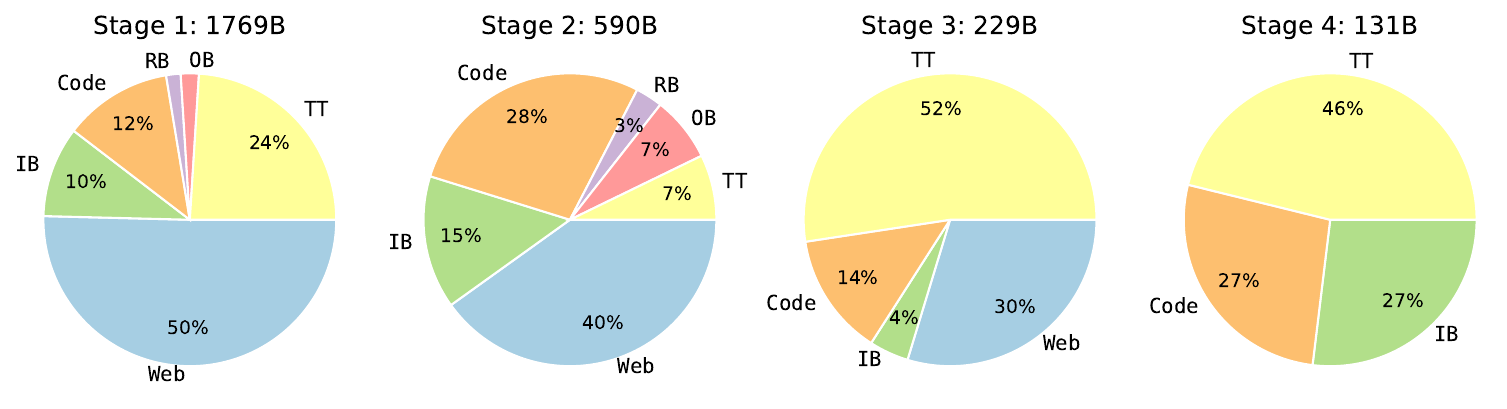}
    \caption{Data mixes for the four mid-training stages. Each stage is annotated with size in billions of tokens. \texttt{TT} refers to \texttt{thinking\_traces}; \texttt{RB} to \texttt{reasoning\_behaviors}; \texttt{OB} to \texttt{other\_behaviors}; \texttt{IB} to \texttt{Institutional Books}.}
    \label{fig:mid-training-mix-pies}
\end{figure}

\begin{figure}[t]
    \centering
    \includegraphics[width=\linewidth]{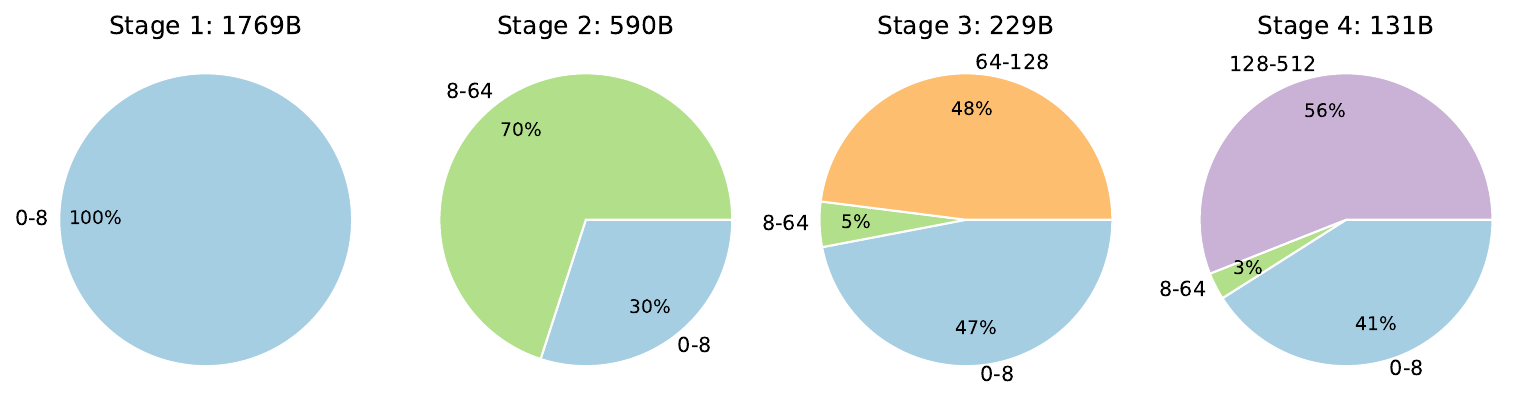}
    \caption{Context mixes for the four mid-training stages. Pie chart sectors correspond to percentage of \emph{tokens} from documents in each of the four context length buckets: 0k-8k; 8k-64k; 64k-128k; 128k-512k.}
    \label{fig:mid-training-context-pies}
\end{figure}

Natural text data is diverse in length and we organize it in 0k-8k, 8k-64k, 64k-128k, and 128k-512k length buckets for better control of context length during the four mid-training stages. We summarize data mixes for each stage in Figure \ref{fig:mid-training-mix-pies} and context mixes in Figure \ref{fig:mid-training-context-pies}.

\begin{figure}
  \begin{center}
     \includegraphics[width=0.6\linewidth]{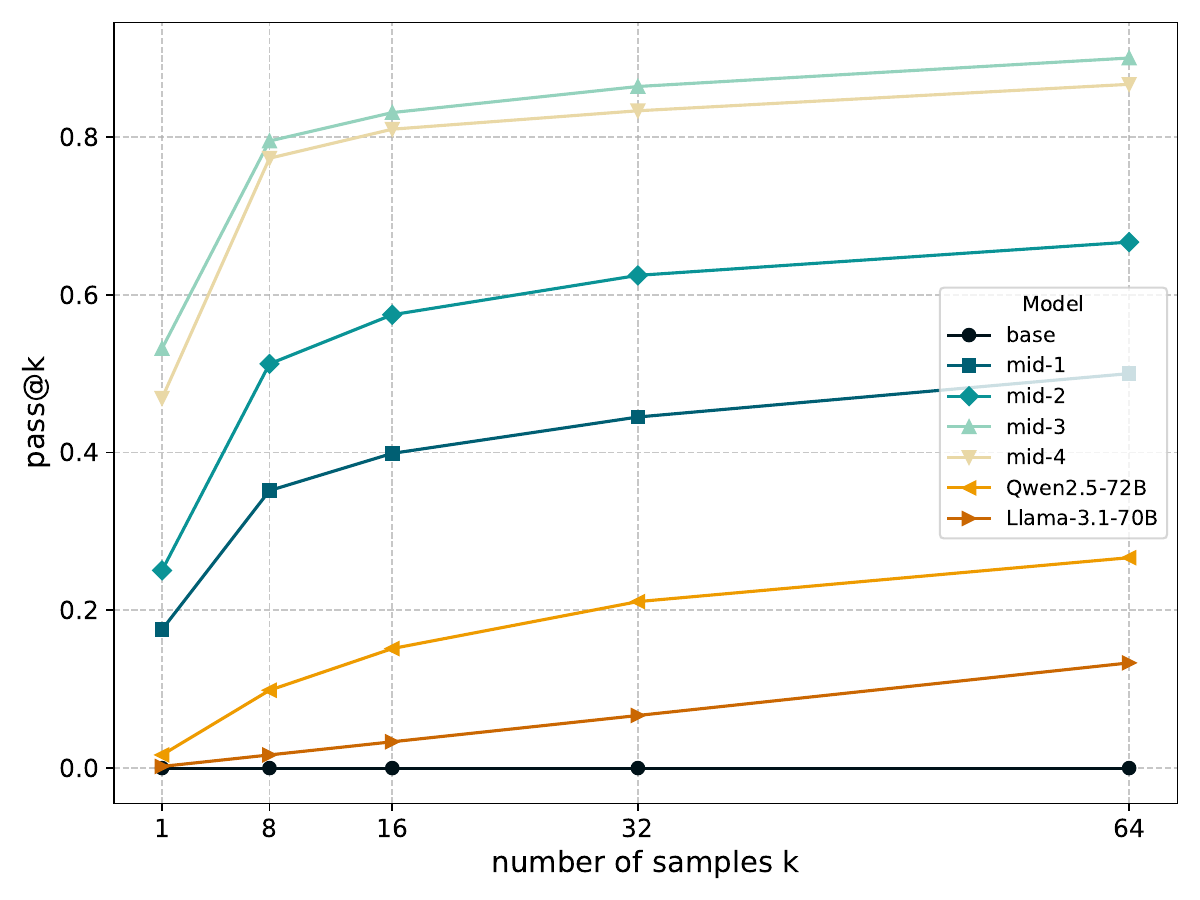} 
  \end{center}
\caption{Coverage (pass$@k$) on AIME-2025 throughout mid-training stages compared to models of similar size trained without reasoning data.}
\label{fig:aime-midtrain}
\end{figure}

To demonstrate effect of reasoning in mid-training we measure pass$@k$ on AIME-2025 for varying $k$ in Figure \ref{fig:aime-midtrain}. AIME consists of challenging mathematical questions that LLMs only recently began to solve well. Typically high performance on AIME is only achieved by post-trained models after SFT and RL. As we show in the figure base models like Llama 3.1 70B and Qwen2.5 72B (which were mid-trained without reasoning data to the best of our knowledge) achieve low pass rate even at $k=64$ attempts despite their strong performance on simpler mathematical benchmarks like GSM8K (see Table \ref{tab:comparison_k2_base} for details). Similarly, our \ktwo{} is pre-trained on predominantly natural web data and has 0 pass rate prior to mid-training. However its pass rate improves rapidly throughout reasoning-oriented mid-training, eventually reaching over 80\% pass$@64$ and non-trivial pass$@1$ (i.e., accuracy) performance. We observe mild degradation in stage 4 where we focused on long-context data above 128k as it is hard to obtain high-quality reasoning data of such length and we instead primarily relied on code and books. Our pass$@k$ analysis is motivated by recent works demonstrating correlation between base model's pass$@k$ performance on complex benchmarks and its subsequent performance post RL \citep{yue2025does,yurochkin2025limitationsofrl}.

\paragraph{Memorization analysis} 

\begin{wraptable}[7]{r}{0.45\textwidth} 
    \vspace{-11pt} 
    \centering
    \small
    \setlength{\tabcolsep}{6pt}
    \begin{tabular}{lcc}
    \toprule
     & AIME 2024 & AIME 2025 \\
    \midrule
    \ktwo{}           & 23.32 & 14.43 \\
    Qwen 2.5 72B  & 41.96 & 23.37 \\
    Llama 3.1 70B & 9.84  & 9.85  \\
    Olmo 3 32B    & 37.82 & 14.47 \\
    \bottomrule
    \end{tabular}
    \vspace{-4pt}
    \caption{Sentence-level memorization rates (\%).}
    \label{tab:memorization_aime}
\end{wraptable}

Typical decontamination approaches rely on n-gram matching, however prior works reported that effect of such decontamination on benchmark performance is unclear \citep{li2024datacomp}. In practice there could be many factors affecting whether an LLM memorizes a benchmark question. Motivated by recent work on data extraction directly from trained models \citep{cooper2025extracting}, we conduct memorization analysis on AIME 2024 and AIME 2025 questions. We split each question into sentences and generate with beam search (width 10, up to 20 tokens at a time) conditioned on each subsequence of sentences until a period. We measure how often generated sentence \emph{exactly} matches the corresponding sentence in the question. Results are reported in Table \ref{tab:memorization_aime}. We see that the memorization rate is overall lower for AIME 2025 than for AIME 2024, as expected since this benchmark is more recent. For 2025, the majority of memorization comes from the last sentence, which is typically formulated to ensure that the final answer is a single number---a pattern present in most AIME iterations. We also notice that \ktwo{}’s memorization for both years is low relative to other models, suggesting that its strong AIME performance is driven by reasoning ability rather than memorization alone.

\subsection{Mid-training Infrastructure}\label{sec:mid-train-infra}

We have developed our own training framework instead of directly using Megatron-LM for mid-training, as Megatron-LM's support for context parallelism and FSDP is not yet mature. Our in-house framework supports TP, CP, and FSDP, with a finer-grained automatic activation recomputation mechanism to better balance memory footprint and training speed. We also provide improved asynchronous communication among operations in LLMs, as well as customized fused kernels.

The in-house training infrastructure features on-the-fly tokenization and online best-fit packing. On-the-fly tokenization will avoid the hassle of pre-tokenization when adjusting the dataset mix, making it much more flexible for experiments involving changes in dataset sources or dataset weights.

We modify the Best-Fit packing algorithm \citep{ding2024fewertruncationsimprovelanguage} and implement an online version in our framework. We enable this feature starting from stage 2. It significantly reduces unnecessary document truncations at the cost of only a few additional padding tokens. Our ablation experiments show that this feature improves loss given the same model and data.

\paragraph{Parallelism Strategy}
The tensor parallelism size (TP) is fixed to 8 throughout the four mid-training stages. 
For context parallelism size (CP), we set it to 1 for stage 1 and stage 2, and gradually increasing it to 2 for stage 3 and to 8 for stage 4 to keep memory usage under control. In our implementation of context parallelism, we use standard all-gather communication of keys and values between ranks in a context-parallel group, with asynchronous communication to hide overheads behind the computation of queries. We find this simple implementation to be more efficient than the RingAttention algorithm~\citep{liu2024ring}.

\subsection{The Mid-training Run}\label{sec:mid-train-run}

\begin{figure}
    \centering
    \includegraphics[width=0.95\linewidth]{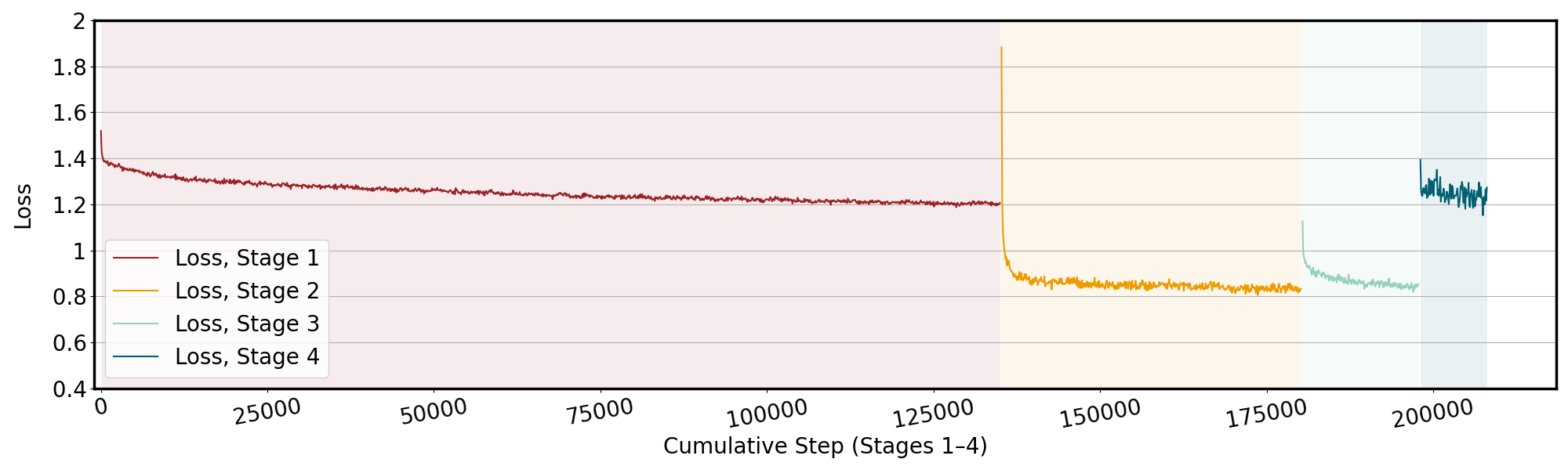} \includegraphics[width=0.95\linewidth]{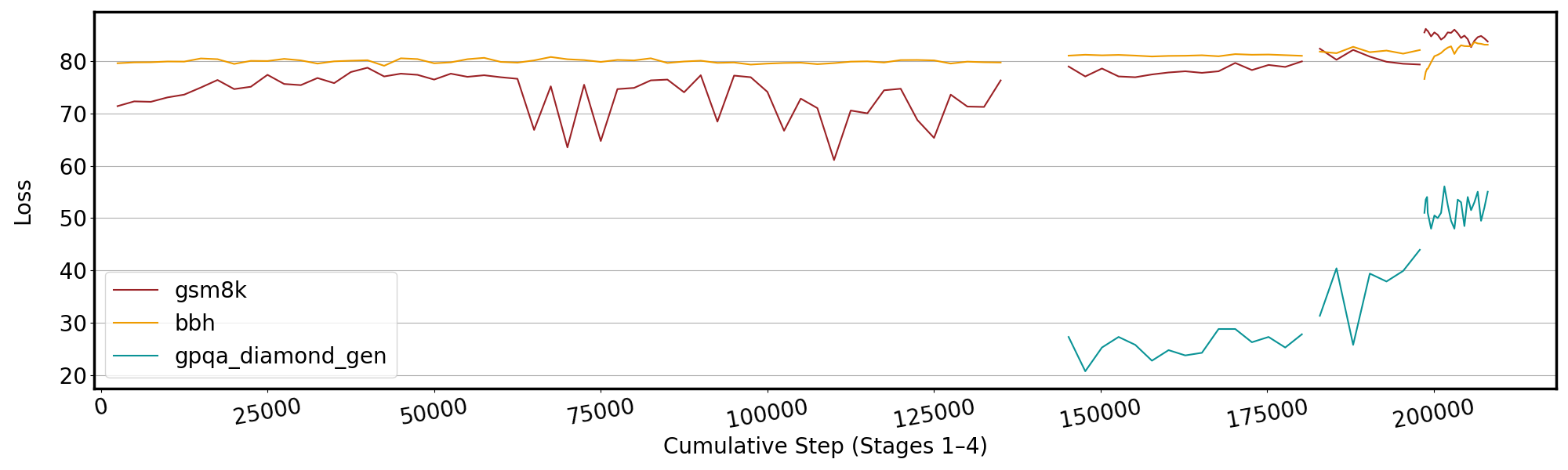}
    \caption{a) Loss curve of four mid-training stages. The losses may jump up or down at the beginning of next stage due to different data distributions. b) Trend of key metrics across four mid-training stages. Despite the fluctuations, the general upward trend could be observed across four stages. (We missed four evaluation data points at the beginning of stage 2)
    }
    \label{fig:k2-mid-loss}
\end{figure}

In this section, we discuss observations from the mid-training stages. Figure~\ref{fig:k2-mid-loss}a illustrates the loss curves. Although we included a replay of web data in each stage, the distinct loss ranges indicate significant domain differences across stages. Since we are simultaneously extending context length and adding new capabilities, it is challenging to reduce the distribution differences across the stages. Despite these shifts, we have observed the key metrics improve or remain competitive. The results will be presented in detail in \S\ref{sec:base_eval}.

Figure~\ref{fig:k2-mid-loss}b tracks key metrics throughout this process. For traditional benchmarks, especially those evaluated via perplexity, performance generally remains stable or improves only slightly. In contrast, complex reasoning benchmarks demonstrate consistent gains. For example, GPQA-Diamond shows steady improvement across the final three stages designed for it. However, GSM8K performance has been unstable. We attribute this to two factors: changes in the model's output format (which caused parsing errors) and the mathematical data distribution used in each stage. Additionally, we observe a distinct pattern in which certain scores increase significantly at the very beginning of a stage but remain flat thereafter. We interpret these rapid initial gains as a shift in model behavior or output style rather than the acquisition of new knowledge.

\begin{figure}
    \centering
    \includegraphics[width=0.45\linewidth]{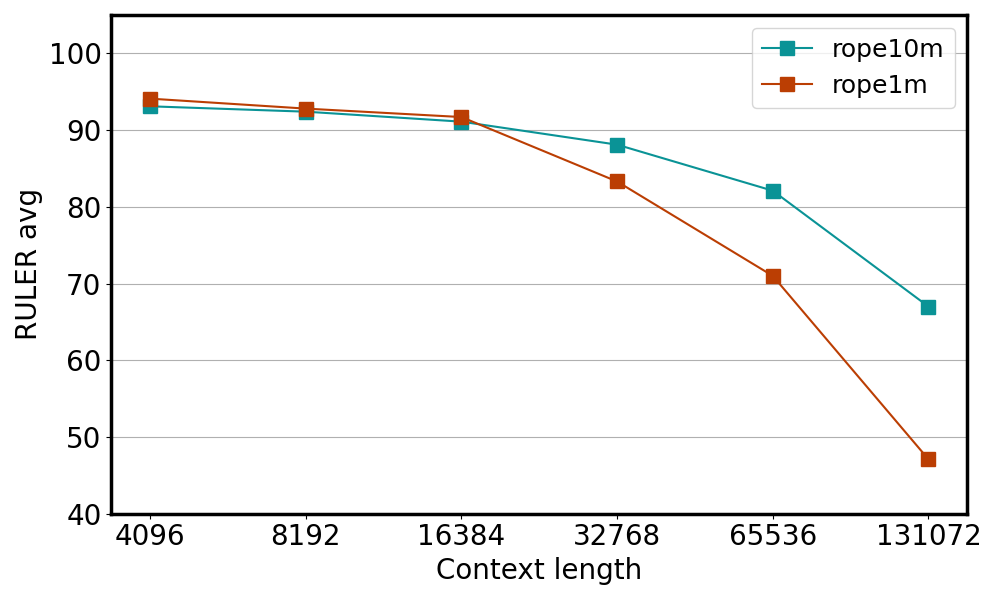} \includegraphics[width=0.45\linewidth]{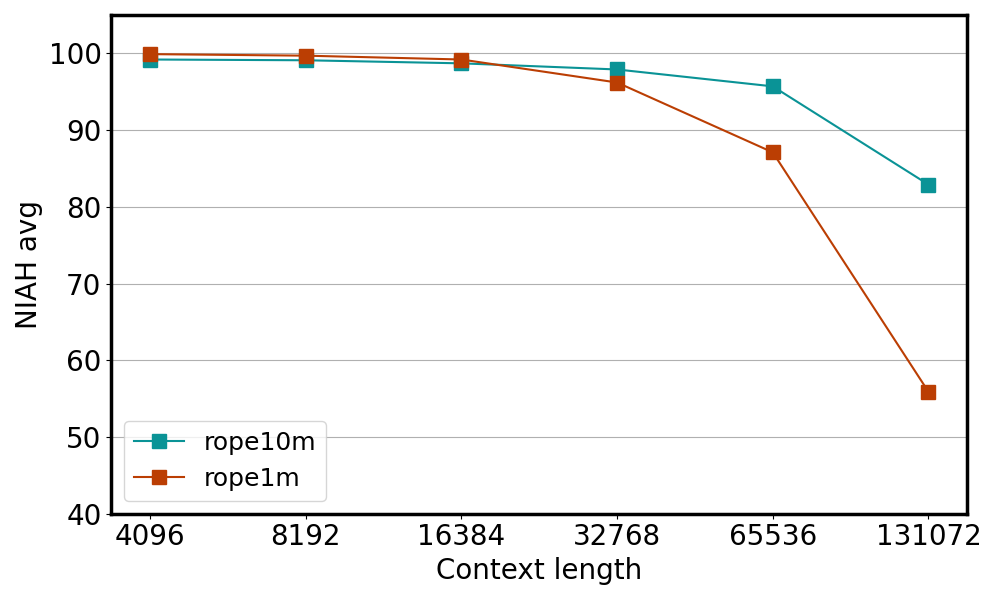}
    \caption{Impact of RoPE base frequency configuration on long-context performance at the $17{,}500$-step checkpoint of stage 3 mid-training. An insufficient RoPE base frequency leads to degraded performance, especially for longer evaluation context length cases. a) RULER average for runs using base frequencies 1M and 10M. b) NIAH average for the same comparison.}
    \label{fig:k2-mid-rope-comp}
\end{figure}

Finally, we report a key finding regarding the Rotary Positional Embedding (RoPE) configuration. In an early experiment targeting a 128K context length, we set the RoPE base frequency to 1M. We find this value insufficient to support extended context capabilities, which resulted in degraded performance. We subsequently adjusted the base frequency to 10M, after which performance recovered to expected values. See Figure \ref{fig:k2-mid-rope-comp} for the comparison.

\section{Base Model Evaluation}\label{sec:base_eval}

\begin{keytakeaways}
    \begin{itemize}

    \item \textbf{Reasoning-focused mid-training unlocks superior capabilities.} Throughout our mid-training stages, \ktwo{} dramatically improved in performance across mathematics, STEM, and logic domains. By the end of mid-training, our model significantly outperforms other base LLMs and rivals fully-trained models on challenging math and logic benchmarks.
    
    \item \textbf{Reasoning in mid-training boosts model capabilities but presents evaluation challenges.} To comprehensively quantify model improvements due to reasoning we expanded classical pre-training evaluations and considered complex generative benchmarks typically only used for fully-trained LLMs. We also needed to depart from the traditional few-shot prompting approaches and develop new prompts, robust parsers, and extend generation length when evaluating.

    \item \textbf{Breadth of evaluation is crucial to measure progress.} In our experiments several classical pre-training benchmarks slightly degraded throughout mid-training, which was discouraging initially. Extending our evaluation suite allowed us to better quantify model's abilities and gain confidence in the mid-training recipe. 
    \end{itemize}
\end{keytakeaways}

In this section, we present a comprehensive evaluation of the \ktwo{} series across its fundamental training stages: pre-training and mid-training. Our evaluations include traditional completion-likelihood-based benchmarks as well as a variety of generative tasks covering math, STEM, coding, instruction following, and logic puzzles.
The results of these evaluations are summarized in Table~\ref{tab:comparison_k2_base}.

\begin{table}[ht]
\centering
\scriptsize
\begin{tabular}{lcccccccccc}
\toprule
 & \textbf{base} & \textbf{mid-1} & \textbf{mid-2} & \textbf{mid-3} & \textbf{mid-4} & \textbf{Qwen2.5-72b} & \textbf{Llama3.0-70b} & \textbf{Llama3.1-70b} & \textbf{Olmo3-32b}\\
 \midrule
\# Total Params & 70B & 70B & 70B & 70B & 70B & 72B & 70B & 70B & 32B \\
\midrule
\multicolumn{10}{c}{\textit{\textbf{General Tasks}}} \\
MMLU & 74.3 & 74.4 & 73.5 & 75.0 & 75.2 & \textbf{86.1} & \uline{79.5} & 79.3 & 75.2 \\
MMLU-Pro & 43.7 & 46.8 & 48.1 & \textbf{59.8} & 57.0 & \uline{58.1} & 52.8 & 53.8 & 49.6 \\
BBH & 68.4 & 79.8 & 81.1 & 82.2 & \uline{83.2} & \textbf{86.3} & 82.2 & 82.1 & 77.6 \\
HELLASWAG & \uline{87.8} & 86.9 & 86.6 & 86.6 & 86.0 & 87.6 & \textbf{88.0} & 85.0 & 84.8 \\
WINOGRANDE & 82.6 & 83.7 & 83.7 & 83.7 & 83.0 & 83.9 & \uline{85.3} & 79.8 & \textbf{90.3} \\
PIQA & 84.2 & 84.0 & 83.3 & 82.9 & 83.1 & 83.5 & \uline{84.6} & 84.3 & \textbf{85.6} \\
TRUTHFULQA & 54.0 & 54.9 & 55.1 & \uline{55.8} & 53.9 & \textbf{60.5} & 45.6 & 49.7 & 54.9 \\
\midrule
\multicolumn{10}{c}{\textit{\textbf{Math \& STEM Tasks}}} \\
GPQA-DIAMOND & 26.3 & 31.3 & 27.8 & \uline{43.9} & \textbf{55.1} & 34.9 & 21.2 & 27.3 & 30.3 \\
GSM8K & 68.0 & 76.4 & 82.1 & \textbf{93.6} & \uline{92.5} & 91.2 & 83.2 & 81.1 & 80.5 \\
MATH & 27.8 & 38.2 & 41.1 & \textbf{94.7} & \uline{91.4} & 58.5 & 41.9 & 41.6 & 43.4 \\
AIME 2025 & 0.0 & 17.6 & 25.1 & \textbf{53.2} & \uline{46.9} & 1.7 & 0.1 & 0.2 & 14.7 \\
ARC-CHALLENGE & 64.9 & 66.4 & 66.4 & 66.0 & 66.3 & \textbf{72.4} & \uline{69.2} & 64.9 & 65.4 \\
\midrule
\multicolumn{10}{c}{\textit{\textbf{Coding Tasks}}} \\
MBPP & 57.6 & 57.8 & 58.2 & 59.8 & 61.8 & \textbf{75.4} & \uline{69.2} & 64.4 & 60.2 \\
HUMANEVAL & 50.0 & 51.2 & \uline{53.7} & \textbf{54.3} & \textbf{54.3} & \textbf{54.3} & 42.1 & 50.6 & 36.0  \\
\midrule
\multicolumn{10}{c}{\textit{\textbf{Logic Puzzles}}} \\
COUNTDOWN & 1.3 & \uline{53.3} & 53.1 & 35.9 & \textbf{75.6} & 6.0 & 1.0 & 0.5 & 23.2 \\
KK-4 PEOPLE & 4.8 & 44.9 & \uline{68.0} & 64.5 & \textbf{92.9} & 26.1 & 4.2 & 7.6 & 42.4 \\
KK-8 PEOPLE & 0.5 & 23.2 & 41.3 & \uline{51.6} & \textbf{82.8} & 5.7 & 1.1 & 1.3 & 13.0 \\
ORDER-15 ITEMS & 4.7  & 30.7 & 47.2 & \uline{55.8} & \textbf{87.6} & 37.0 & 3.5 & 4.5 & 25.0 \\
ORDER-30 ITEMS & 0.0  & 0.3  & 3.0        & \uline{34.1} & \textbf{40.3} & 0.7  & 0.2 & 0.1 & 0.6  \\
\midrule
\multicolumn{10}{c}{\textit{\textbf{Instruction Following}}} \\
IFEVAL & 17.4 & 26.2 & 28.5 & \uline{34.5} & 26.7 & \textbf{40.3} & 15.1 & 17.4 & 13.2 \\
\midrule
\multicolumn{10}{c}{\textit{\textbf{Arabic}}} \\
MMLU-Arabic & 65.4 & 66.1 & 64.5 & 66.6 & 65.5 & \textbf{74.1} & 65.0 & \uline{66.8} & 47.8 \\

\bottomrule
\end{tabular}
\caption{Comparison among \ktwo{} pre-training, mid-training stages and other strong open-source benchmarks in similar parameter scale. The highest, the second-best scores are shown in \textbf{bold} and \uline{underlined}, respectively.\label{tab:comparison_k2_base}}
\end{table}

\subsection{Evaluation Methodology}
The evaluation of base models mainly focuses on their performance in general knowledge, reasoning, mathematics,
scientific knowledge, coding, and multilingual (arabic) capabilities. The evaluation datasets for base
model checkpoints include 19 benchmarks:

\begin{itemize}
\item General Tasks: MMLU \citep{hendryckstest2021} (5-shot), MMLU-Pro \citep{wang2024mmlu} (5-shot), BBH \citep{suzgun2022challenging} (3-shot, CoT), HELLASWAG \citep{zellers2019hellaswag} (10-shot), WINOGRANDE \citep{ai2:winogrande} (5-shot), PIQA \citep{Bisk2020} (0-shot).
\item Math \& STEM Tasks: GPQA-DIAMOND \citep{rein2024gpqa} (0-shot, CoT), GSM8K \citep{cobbe2021gsm8k} (5-shot, CoT and 0-shot reasoning prompt),
MATH \citep{hendrycksmath2021} (4-shot, CoT and 0-shot reasoning prompt), AIME 2025 (0-shot reasoning prompt), ARC-CHALLENGE \citep{Clark2018ThinkYH} (25-shot).
\item Coding Tasks: MBPP \citep{austin2021program} (3-shot), HUMANEVAL \citep{chen2021evaluating} (0-shot).
\item Logic Puzzles: COUNTDOWN \citep{gandhi2024stream} (0-shot reasoning prompt),
Knights and Knaves (KK) \citep{xie2024memorization,johnson1990meta,smullyan1978what} (0-shot, reasoning prompt), ORDER (0-shot reasoning prompt) where an LLM needs to infer the overall order in which a set of items was sold based on pairwise relations.
For KK and ORDER we consider varying levels of difficulty and report results for medium and hardest levels (4 and 8 people for KK; 15 and 30 items for ORDER).
\item Instruction Following: IFEVAL \citep{zhou2023instructionfollowingevaluationlargelanguage} (0-shot).
\item Adversarial: TRUTHFULQA \citep{lin2021truthfulqa} (0-shot)
\item Arabic: MMLU-ARABIC \citep{koto2024arabicmmlu} (5-shot).
\item Long Context Tasks: RULER (0-shot)\citep{hsieh2024rulerwhatsrealcontext}, NIAH (0-shot)\citep{nelson2024needlehaystackmemorybased}.
\end{itemize}
Table~\ref{tab:benchmarks_types} summarizes the metric types reported for each benchmark, including perplexity-based multiple choice (MC), generation-based exact match (EM), and accuracy-based (Pass@k). Long context evaluation will be discussed in \S\ref{sec:long-context-eval}.

\begin{table}[h]
    \centering
    \small
    \setlength{\tabcolsep}{5pt}
    
    \begin{tabular}{
        >{\columncolor{colbg}}l >{\columncolor{white}}c   
        @{\hspace{1.5em}}                                 
        >{\columncolor{colbg}}l >{\columncolor{white}}c   
        @{\hspace{1.5em}}                                 
        >{\columncolor{colbg}}l >{\columncolor{white}}c   
        @{\hspace{1.5em}}
        >{\columncolor{colbg}}l >{\columncolor{white}}c
    }
        \toprule
        \rowcolor{white} 
        \textbf{Benchmark} & \textbf{Metric} & 
        \textbf{Benchmark} & \textbf{Metric} & 
        \textbf{Benchmark} & \textbf{Metric} &
        \textbf{Benchmark} & \textbf{Metric} \\
        
        \arrayrulecolor{dimrule}
        \midrule
        \arrayrulecolor{black}        
        
        MMLU & MC & PIQA & MC & MBPP & Pass@1 & AIME 2025 & EM\\
        MMLU-Pro & EM & GPQA-Diamond & EM & HumanEval & Pass@1 & CountDown & Pass@1\\
        BBH & EM & GSM8K & EM & IFEval & MC & KK & Pass@1 \\
        HellaSwag & MC & MATH & EM & TruthfulQA & MC & Order & Pass@1 \\
        Winogrande & MC & ARC-Challenge & MC & MMLU-Arabic & MC \\
        
        \bottomrule
    \end{tabular}
    \caption{Summary of evaluation benchmarks and corresponding metrics (MC: Multiple Choice, EM: Exact Match).}
    \label{tab:benchmarks_types}
\end{table}

As baselines we select leading open-source base models of comparable scale, 
Llama 3 and 3.1 70B \citep{grattafiori2024llama3herdmodels}, Qwen2.5 72B \citep{qwen2.5}, and Olmo-3 32B \citep{Olmo3TechnicalReport2025}. All models are evaluated using the same evaluation pipeline
to ensure fair comparison. On GSM8K and MATH we report best performance across few-shot CoT and reasoning prompt setups (see Section \ref{sec:eval:details} for additional details). For non-\ktwo{} models, we report the best score across results that are publicly reported or that we reproduced ourselves. 


As discussed in subsequent sections, the details of the implementation can substantially affect the final reported numbers. To mitigate this, we also open source our evaluation code in Eval360\footnote{https://github.com/LLM360/Eval360}, to make the whole process transparent and trustworthy. The core evaluation logic is adapted from lm-evaluation-harness \citep{eval-harness} and we build support for more custom datasets, metrics, filters, and parsers. The numbers in this section are obtained by setting the maximum generation length to 32,768 tokens for all checkpoints, and temperature to 1.0 for all of the non-greedy decoding benchmarks. For AIME and all logic puzzles we report results averaged over 64 evaluation runs.

\subsection{Evaluation Results}
Our \ktwo{} models show consistent improvements during the pretraining and four mid-training stages, producing strong performance on a broad range of benchmarks. The gains are most pronounced on Mathematics, STEM reasoning, and logic-intensive tasks, where \ktwo{} mid-stage models substantially outperform all open-source baselines of comparable scale.

For general-domain tasks, while \ktwo{} does not match the overall MMLU performance of Qwen2.5-72B (86.1), our models achieve competitive or state-of-the-art scores on several sub-benchmarks. Notably, \texttt{mid-3} obtains the highest score among open models on MMLU-Pro (59.8), and achieves strong results on TruthfulQA (55.8) and BBH (83.2), outperforming LLaMA~3/3.1 by a large margin. These improvements suggest that the staged mid-training pipeline enhances robustness and general reasoning capabilities even when not optimized for raw MMLU.

In Mathematics and STEM, \ktwo{} models demonstrate clear state-of-the-art performance. The \texttt{mid-4} model achieves the highest scores among all compared models on GPQA-Diamond (55.1), GSM8K (93.6), and MATH (94.7). These substantial gains highlight the effectiveness of our progressive mid-training strategy for developing structured reasoning and multi-step problem-solving ability.

For coding tasks, \ktwo{} shows steady improvements, though Qwen2.5-72B remains the strongest on MBPP (75.4). The \texttt{mid-4} model reaches 61.8 on MBPP and matches Qwen2.5-72B on HumanEval pass@1 (54.3), outperforming LLaMA 3.0/3.1 and most other baselines. Despite these gains, MBPP remains a relative weakness and warrants further discussion.

On logic puzzles, \ktwo{} achieves dramatic and consistent state-of-the-art performance. The \texttt{mid-4} model attains 75.6 on \textsc{Countdown}, 92.9 on \textsc{KK-4 People}, 82.8 on \textsc{KK-8 People}, and up to 87.6 on \textsc{Order-15 Items}, often surpassing other 70B-scale models by large margins. Further, our performance on \textsc{KK-8 People} (hardest difficulty level) matches leading fully-trained models such as DeepSeek-R1 (83\%) and o3-mini-high (83\%) based on the results reported by \citet{xie2025logic}. These results demonstrate robust generalized reasoning capabilities beyond what traditional NLP benchmarks capture.

Overall, the \ktwo{} pretraining and staged mid-training pipeline delivers steady, meaningful, and often state-of-the-art improvements across diverse domains. The strongest gains appear in Math, STEM, and logic reasoning, establishing \ktwo{} as a highly capable 70B-scale open model.

\begin{figure}[h!]
    \centering
    \includegraphics[width=0.45\textwidth]{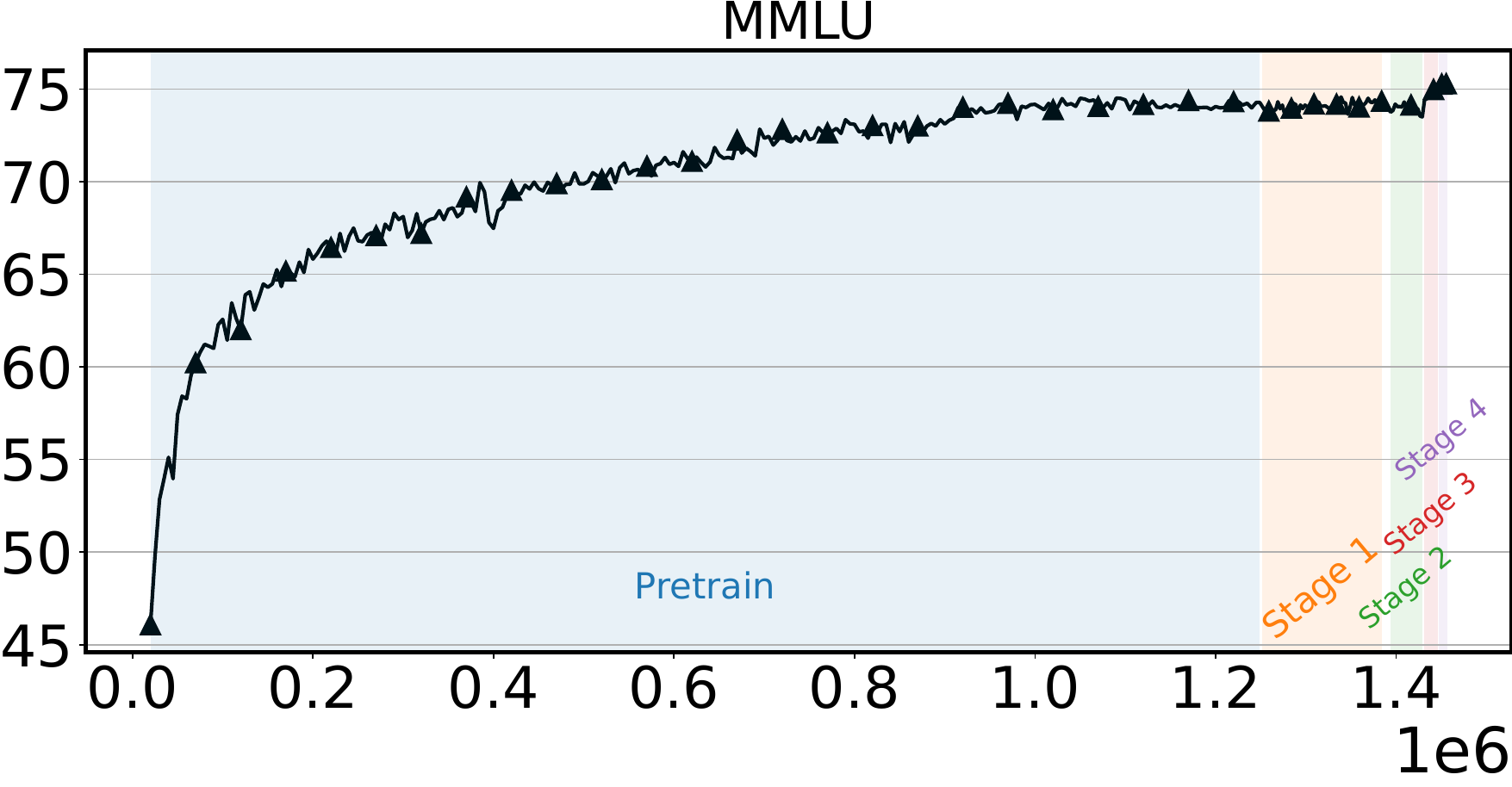}
    \includegraphics[width=0.45\textwidth]{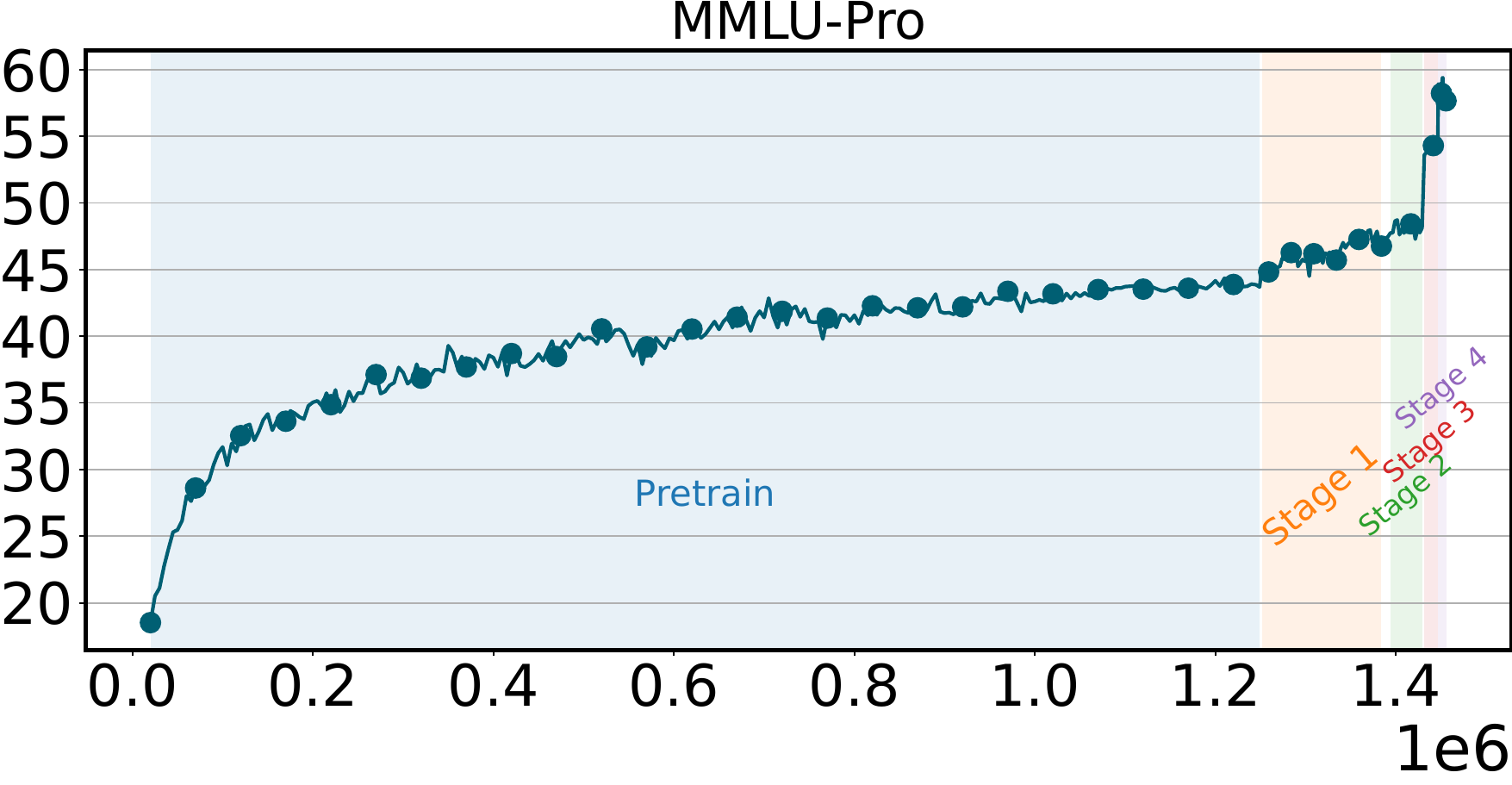}\\
    \includegraphics[width=0.45\textwidth]{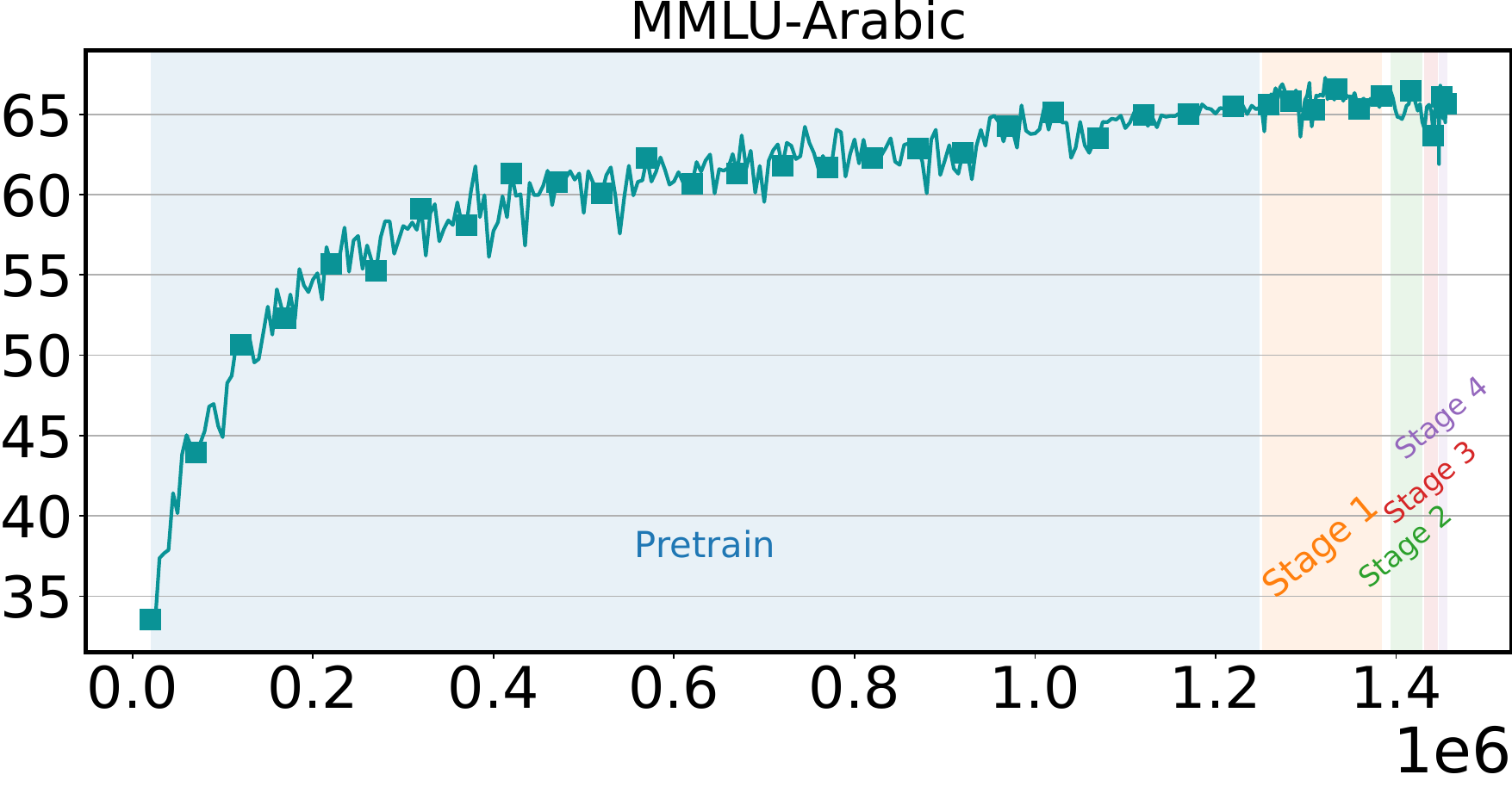}
    \includegraphics[width=0.45\textwidth]{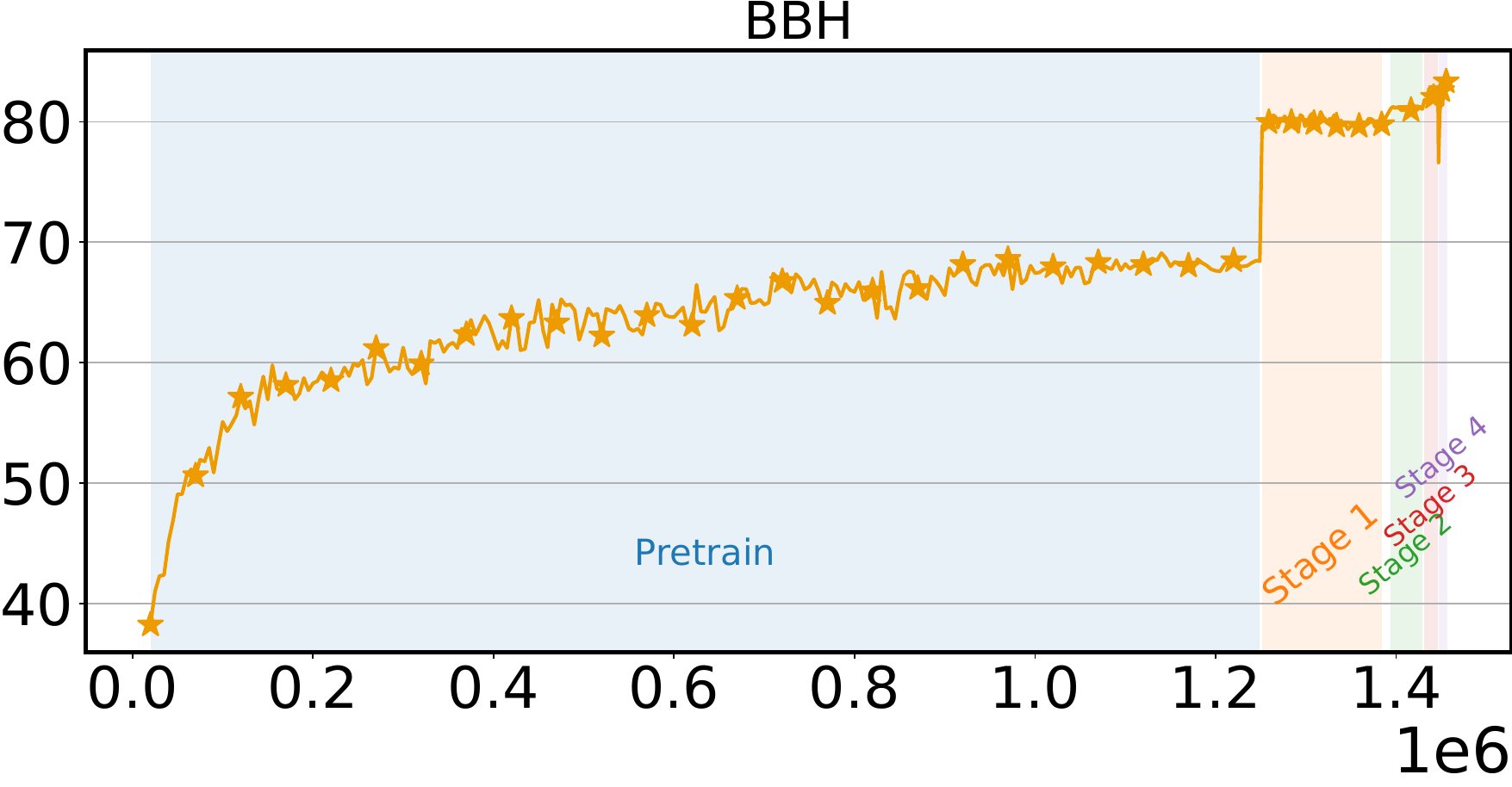}\\
    \includegraphics[width=0.45\textwidth]{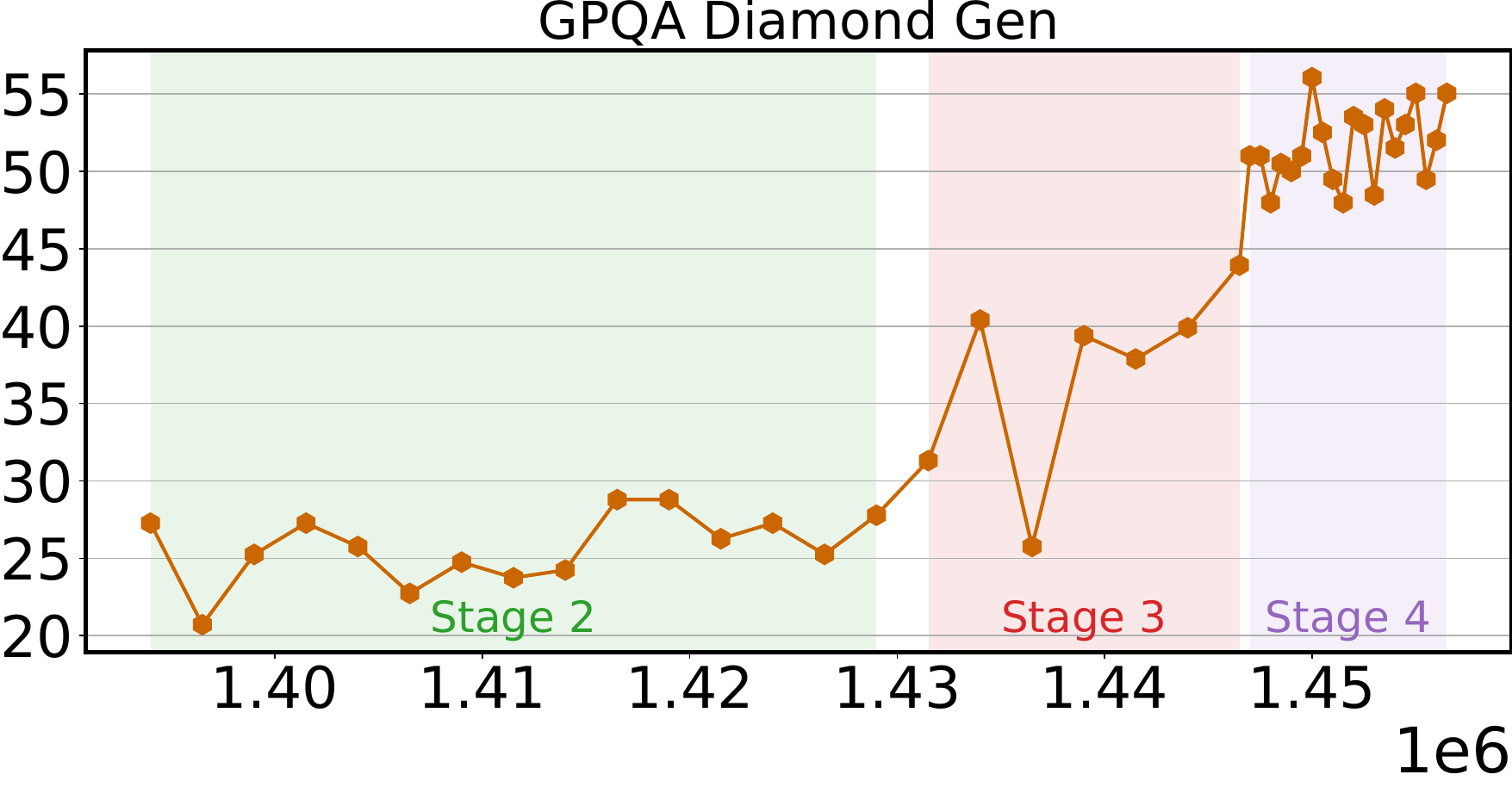}
    \includegraphics[width=0.45\textwidth]{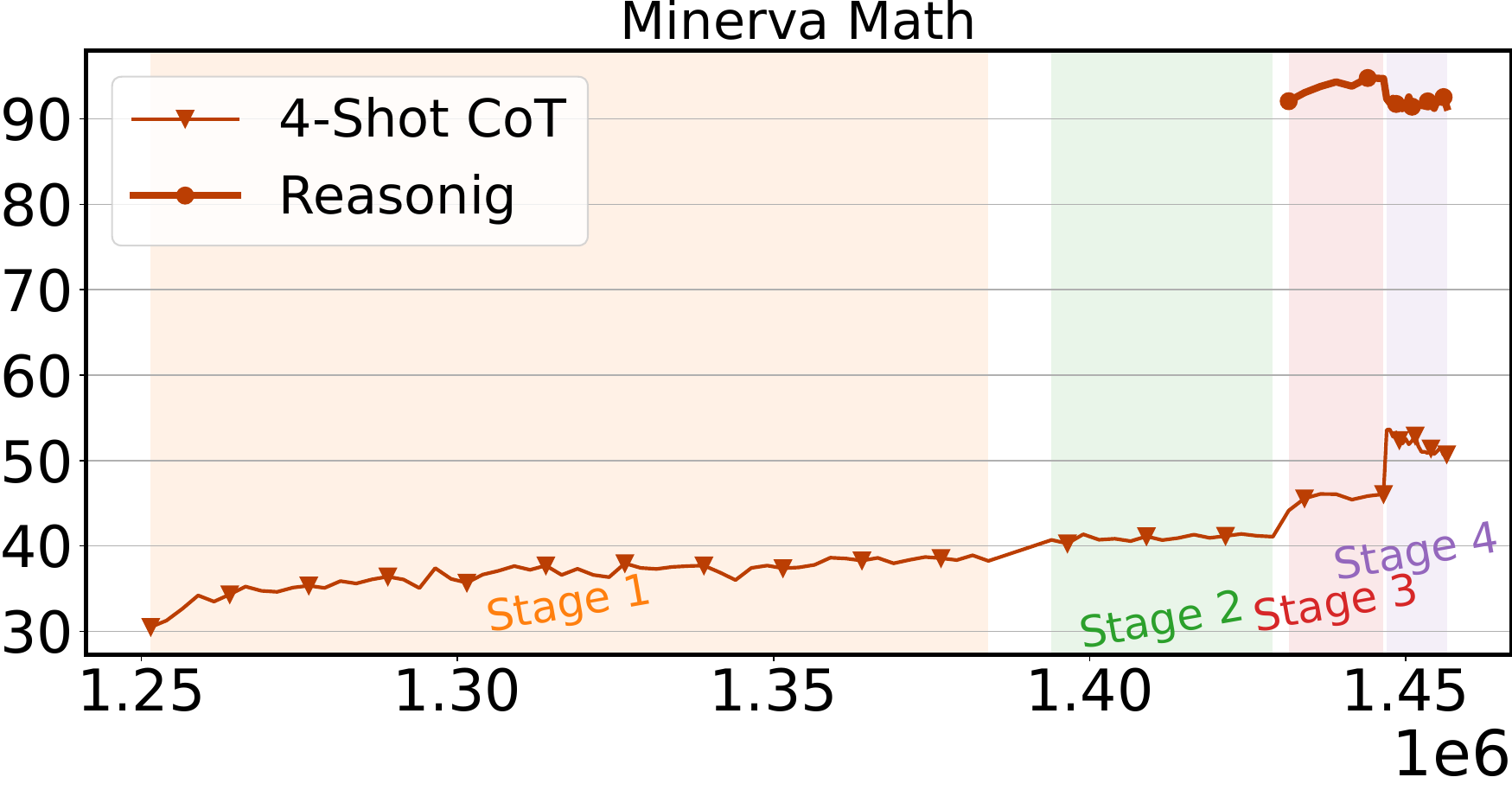}\\
    \includegraphics[width=0.45\textwidth]{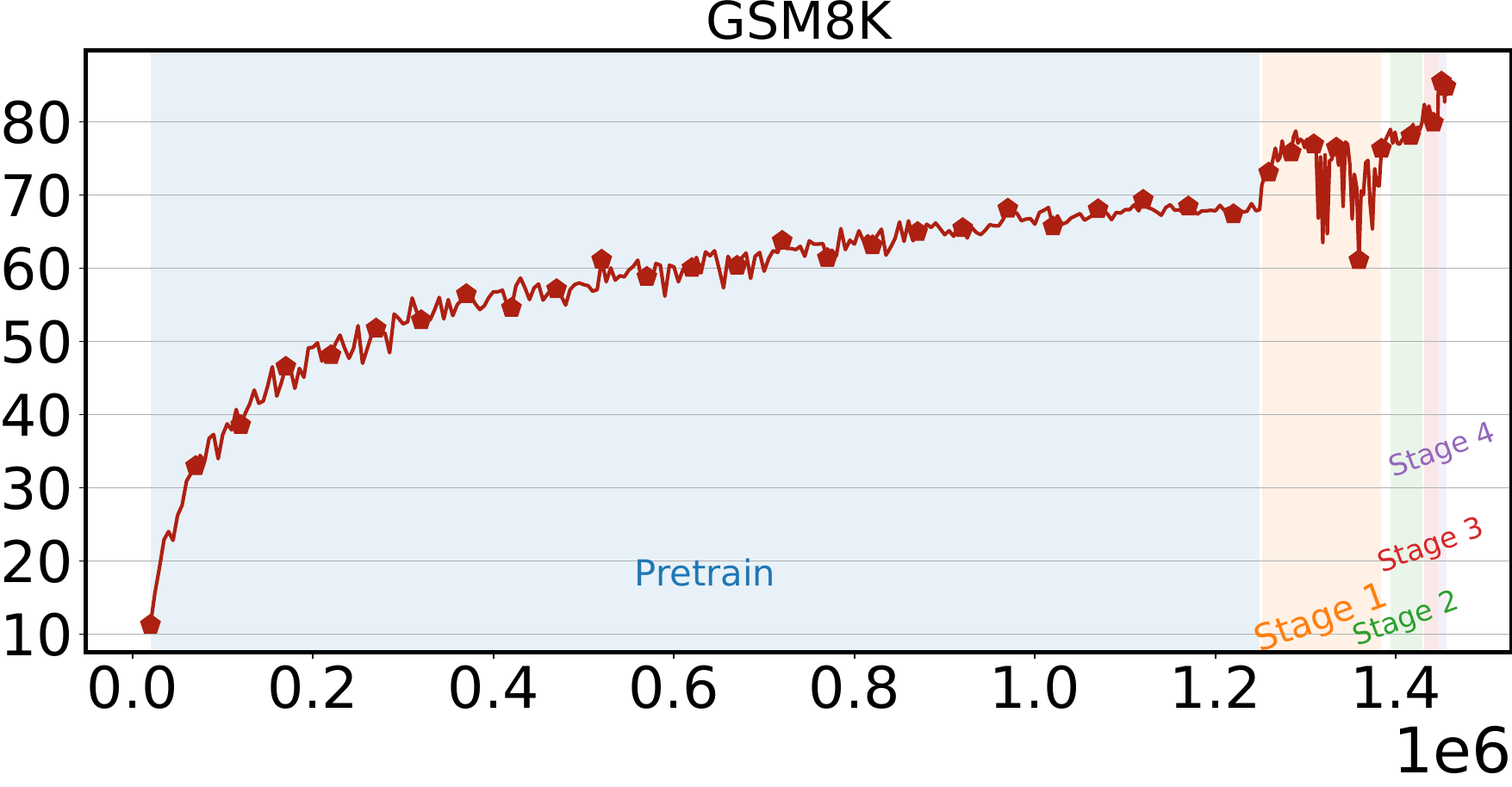}
    \includegraphics[width=0.45\textwidth]{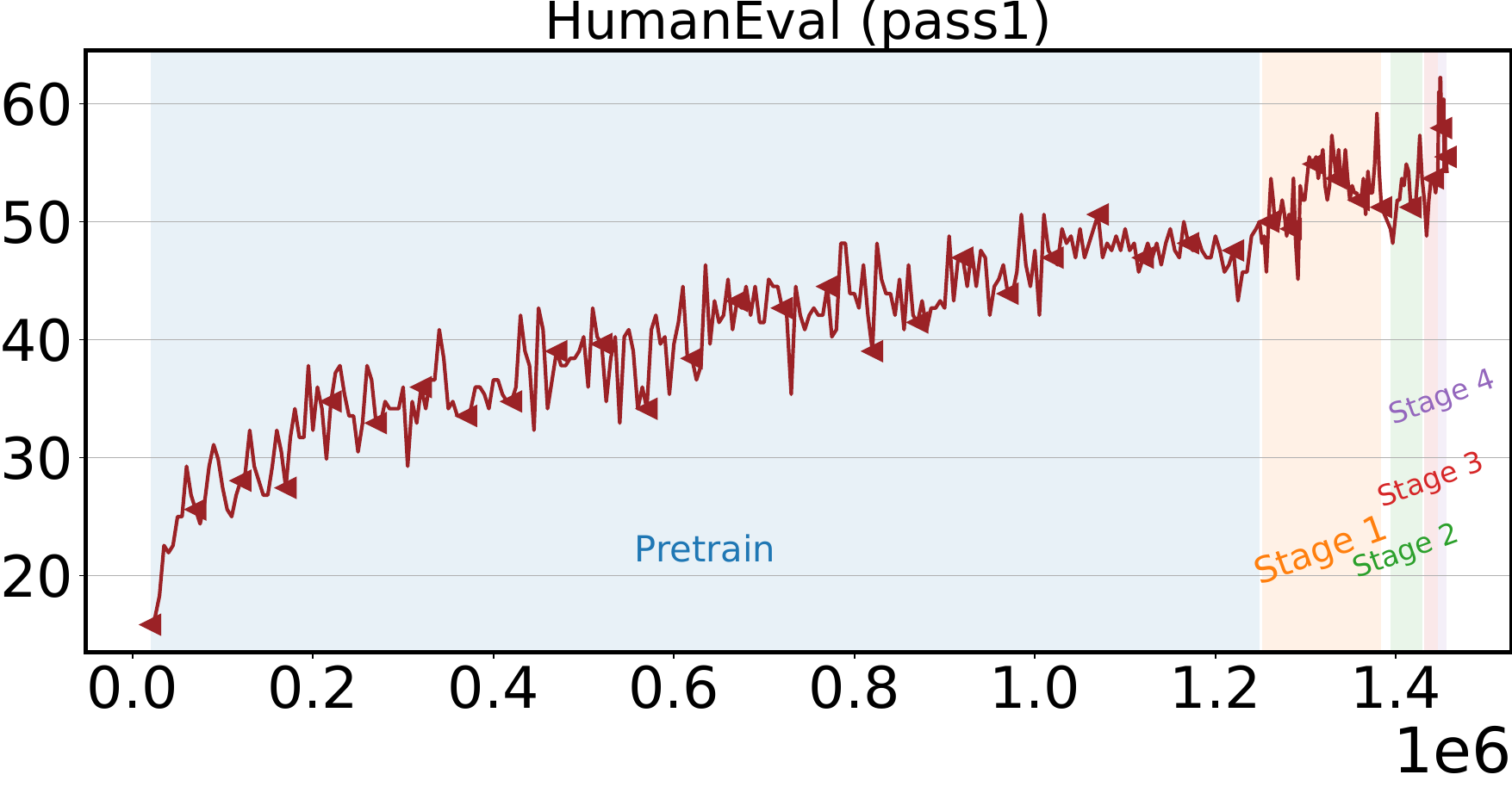}\\
    \caption{Trajectories of evaluation metrics for \ktwo{} during the pretraining and mid-training stages (stages 1--4). For the GPQA Diamond Gen benchmark, we did not evaluate checkpoints during pretraining. For the Minerva Math benchmark, we evaluated checkpoints from stages 1--4 using the non-reasoning prompt and checkpoints from stages 3--4 using a 4-shot CoT prompt.} \label{fig:all_curves}
\end{figure}

\subsubsection{Evaluation with Reasoning}
\label{sec:eval:details}

In order to quantify effect of reasoning data introduced in mid-training we expanded our evaluation set of benchmarks and experimented with alternative prompting strategies. Specifically we evaluated our models on complex mathematical problems from AIME 2025 and a variety of logic puzzles---benchmarks typically only reported for fully trained models---with a prompt designed to incentivize structured reasoning in base models. We also used this prompt for GSM8K and MATH benchmarks in addition to standard few-shot CoT approaches.

\paragraph{Structured Reasoning Prompt} We use prompt similar to the one DeepSeek-R1-Zero \citep{guo2025deepseek} used for reinforcement learning from their base model. We adjust it slightly based on the task at hand. As an example we provide reasoning prompt for AIME 2025 below.

\begin{figure}[H]
\begin{tcolorbox}[
    colback=white,
    colframe=black,
    boxrule=2pt,
    arc=3mm,
    fonttitle=\bfseries\small,
    title=AIME 2025 Reasoning Prompt,
    coltitle=white,
    colbacktitle=black
]
{\bf \small Prompt:} 
\begin{lstlisting}[breaklines=true,breakautoindent=false,
  breakindent=0pt,basicstyle=\footnotesize\ttfamily]
You are a helpful assistant. To answer the user's question, you first think about the reasoning process and then provide the user with the answer. The reasoning process and answer are enclosed within <think> </think> and <answer> </answer> tags, respectively, i.e., <think> reasoning process here </think><answer> answer here </answer>. Provide a single number as the answer, for example, <answer> 47 </answer>. Now the user asks you to solve a math problem.

User: {problem}
Assistant:
<think>
\end{lstlisting}
\end{tcolorbox}
\end{figure}

\paragraph{Parsing and Token Budgets} We observe that performance on tasks that require logical generation, such as GSM8K, is sensitive to the details such as prompt format, response length, and parsing logic. Accurate scoring depends not only on the model's reasoning but also on successfully extracting the final answer. Variations in output format, such as placing answers in \texttt{\textbackslash boxed} tags versus using headers like \texttt{\#\# Answer}, can significantly alter benchmark results if the parser is not aligned. Furthermore, evaluating these reasoning behaviors introduces a computational trade-off. Unlike standard generation, reasoning protocols require a substantially larger token budget (up to 32k tokens) to accommodate extended thinking traces. This increases the inference cost and must be factored into evaluation design.

\begin{wraptable}[7]{r}{0.45\textwidth}
    \centering
    \small
    \setlength{\tabcolsep}{2pt} 
    \begin{tabular}{lccccc}
    \toprule
     & \textbf{base} & \textbf{mid-1} & \textbf{mid-2} & \textbf{mid-3} & \textbf{mid-4} \\
     \midrule
    GSM8K & 68.0 & 76.4 & 80.0 & 79.4 & 83.8  \\
    GSM8K (CoT) & 63.2 & 77.3 & 82.1 & 87.0 & 90.0 \\
    GSM8K (Reas.) & 78.1 & 60.7 & 67.9 & 93.6 & 92.5 \\
    \bottomrule
    \end{tabular}
    \vspace{-3pt}
    \caption{Comparison of standard QA, CoT, and Structured Reasoning prompts for GSM8K.}
    \label{tab:comparison_gsm8k}
    \vspace{-10pt} 
\end{wraptable}

Besides these details, more explicit prompts yield consistent gains. In the standard QA format, accuracy increases steadily from the \texttt{base} model (68.0) to the \texttt{mid-4} checkpoint (83.8). Chain-of-thought (CoT) prompting surpasses these results at every stage, reaching 90.0. The structured reasoning prompt, which leverages the expanded token budget, yields the strongest results: performance rises from 78.1 at \texttt{base} to 93.6 at \texttt{mid-3} (see Table \ref{tab:comparison_gsm8k}). This confirms that structured prompts with precise parsing are important to evaluating the model correctly.

\subsubsection{Long Context}\label{sec:long-context-eval}

Table~\ref{tab:long-context-base} compares the long-context evaluation of the \ktwo{} checkpoint against several baselines. \texttt{mid-4} stands out for its robust stability. While its short-context RULER scores trail top baselines like Llama 3.1 slightly, its degradation from 4K to 128K is significantly smoother. Unlike earlier stages which experience sharp performance cliffs at extended lengths (>128K), \texttt{mid-4} remains consistent. Similarly, it delivers high NIAH accuracy, remaining above 95\% even at 128K. We also observe that Qwen2.5-14B-1M outperforms the larger Qwen2.5-72B on these metrics. This suggests that specialized long-context training is often more critical for extended input tasks than raw model scale.

We note one challenge we encountered when extending the context length of our model. Despite training with a context window of up to 512k, performance at the extreme limits is constrained by data scarcity. Naturally occurring long documents are rare compared to shorter text. We anticipate that supplementing the training mix with synthetic data designed for these extreme lengths will further bridge this gap.

\begin{table}
\small
\resizebox{\textwidth}{!}{
\begin{tabular}{p{3cm}llllll llllll}
\toprule
 & \multicolumn{6}{c}{RULER Avg.} & \multicolumn{6}{c}{NIAH Avg.} \\
 \cmidrule(lr){2-7} \cmidrule(lr){8-13}
Model & 4k & 8k & 16k & 32k & 64k & 128k & 4k & 8k & 16k & 32k & 64k & 128k \\
\midrule
Llama3.1-8b-instruct 
    & 96.2 & 94.8 & 94.4 & 92.5 & 86.4 & 80.7
    & 99.9 & 99.6 & 99.5 & 99.6 & 99.4 & 96.9 \\

Qwen2.5-14B-Instruct-1M
    & 96.6 & 95.5 & 94.3 & 94.0 & 92.7 & 77.3
    & 100.0 & 99.9 & 99.8 & 99.6 & 99.6 & 98.3 \\

Qwen2.5-72B+yarn
    & 94.6 & 90.7 & 89.1 & 87.1 & 81.7 & 75.0
    & 97.4 & 94.3 & 92.8 & 90.4 & 85.6 & 80.1 \\

Qwen2.5-72B
    & 96.3 & 94.7 & 93.2 & 90.1 & 74.7 & 53.9
    & 99.9 & 99.6 & 99.1 & 93.9 & 76.4 & 46.3 \\

\midrule

mid-2
    & 94.3 & 92.2 & 89.2 & 73.4 & 51.6 & 12.0
    & 99.9 & 99.7 & 98.2 & 91.0 & 69.0 & 7.57 \\

mid-3
    & 93.1 & 92.4 & 91.1 & 88.1 & 82.1 & 67.0
    & 99.2 & 99.1 & 98.7 & 97.9 & 95.7 & 82.9 \\

mid-4
    & 93.9 & 93.2 & 92.9 & 89.6 & 85.3 & 74.6
    & 99.6 & 99.4 & 99.6 & 98.8 & 98.8 & 95.2 \\

\bottomrule
\end{tabular}}
\caption{Evaluation results of long context benchmarks up to 128K. A context length enhanced model (Qwen2-14B-1M) can outperform stronger model. K2 demonstrate stable context performance up to 128K. \label{tab:long-context-base}}
\end{table}

\section{Supervised Fine-tuning}\label{sec:sft}

\begin{keytakeaways}
\begin{itemize}
    \item \textbf{Supervised fine-tuning with mixed reasoning efforts.} We present a novel SFT approach enabling our model to learn three reasoning efforts activated via chat template. Our findings demonstrate that a single LLM can balance efficiency and test-time compute gains, contrary to the recent practice of training separate ``instruct'' and ``thinking'' models.
    \item \textbf{The model learns to adapt to problem difficulty.} We observe gradual improvement of capabilities and increased generation length as we increase the reasoning effort in evaluations. For the high reasoning effort our model demonstrates potential to dynamically adapt its generation length to the problem at hand.
    \item \textbf{Self-identity system prompt needs to be injected at random.} We found that training with a default self-identity system prompt did not yield the desired results. Instead we include self-identity system prompt \emph{at random} for a subset of SFT data.
\end{itemize}
\end{keytakeaways}

During our light SFT phase, our goal is to capitalize on the reasoning capabilities obtained during mid-training while allowing users to experience the model without having to wait for lengthy reasoning to complete. Balancing these objectives is challenging as test-time computation via reasoning is crucial for good performance on many tasks. Recent model releases \citep{yang2025qwen3,Olmo3TechnicalReport2025} choose to train separate model variants for quick high quality responses (``instruct'') and extensive reasoning (``think''), however this necessitates serving both variants to allow users to flexibly adjust the setting based on their task. Instead we implement the Three Reasoning Efforts approach where a single LLM can be used with varying test-time compute budgets via a simple chat template setting leveraging special tokens for each mode. This behavior is similar to GPT-OSS \citep{agarwal2025gpt}, but differs in chat template and is the first open recipe to achieve it.

\subsection{SFT Recipe}\label{sec:sft-recipe}
\paragraph{Three Training Efforts.}
In the supervised fine-tuning we introduce three levels of reasoning effort: \texttt{low}, \texttt{medium}, and \texttt{high}. Reasoning efforts are similar to the GPT-OSS low, medium, and high reasoning effort settings, but differ in how they are induced in a chat template. Correspondingly we utilize GPT-OSS-120B when synthesizing our SFT data for a curated set of prompts detailed in Section \ref{sec:sft_data}.

\begin{figure}[t]
    \centering
    \includegraphics[width=\linewidth]{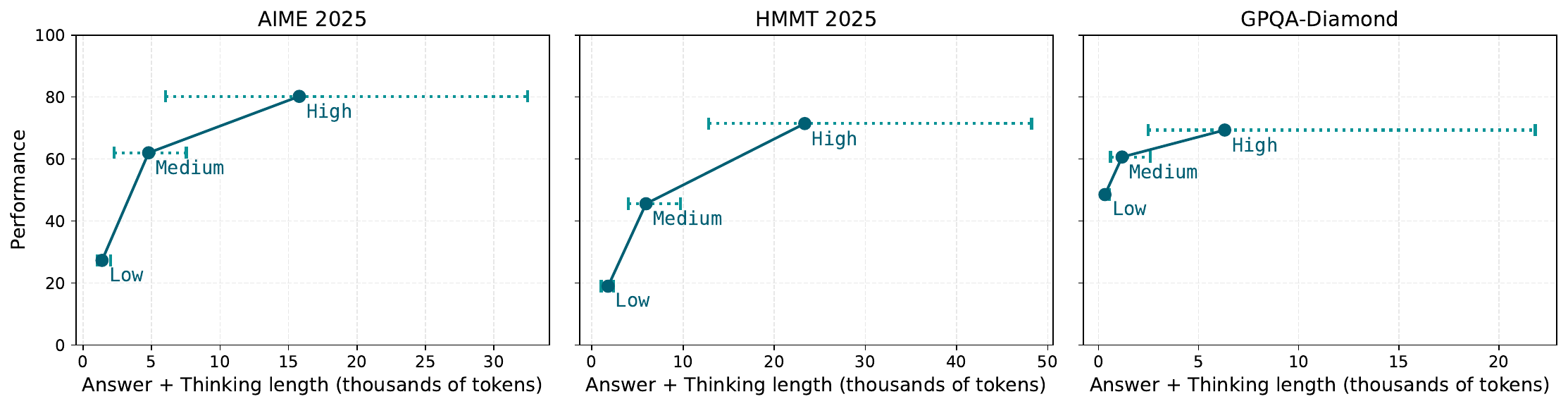}
    \vspace{-0.7cm}
    \caption{Performance of \ktwo{}'s three reasoning efforts on challenging math (AIME 2025 and HMMT 2025) and science (GPQA-Diamond) benchmarks plotted against corresponding generation length quartiles.}
    \label{fig:sft_3hats}
\end{figure}

In Figure \ref{fig:sft_3hats} we demonstrate the effect of the three reasoning efforts. We plot performance of each reasoning effort on challenging math and science benchmarks against the corresponding generation length (answer and reasoning) quartiles, i.e. 25th percentile, median, and 75th percentile. We observe gradual performance increase with the reasoning effort, corresponding to longer generations. We note that \texttt{low} and \texttt{medium} efforts have fairly consistent generation length, while \texttt{high} has a large spread suggesting that model has potential to adjust to the problem difficulty on \texttt{high} reasoning instead of consistently ``overthinking''.

\paragraph{Chat Template Design.}
We use a unified chat template that supports the three reasoning efforts and tool calling. The \texttt{high}, \texttt{medium}, and \texttt{low} reasoning efforts are controlled via special tokens \texttt{<think>}, \texttt{<think\_fast>}, and \texttt{<think\_faster>} that delimit an internal reasoning block for each response. The template is constructed so that these thinking segments only appear in the assistant message corresponding to the most recent user query and are stripped from earlier assistant messages when they are fed back into the context. Although not trained explicitly for this behavior, it is possible to elicit a \texttt{no-think} mode by immediately closing the thinking tags so that the model produces a direct answer with no visible reasoning. For tool use, we follow the Model Context Protocol (MCP) format and use designated tokens such as \texttt{<tools>} to mark the available tools and \texttt{<tool\_calls>} to delimit the segment of the conversation where tool calls are issued.

\paragraph{Training Hyperparameters.} We train the model for three epochs, totaling approximately 6,300 steps given the dataset size of 17.4 billion tokens per epoch. We use a global batch size of 128 with a micro batch size of 1. To balance data utility with training efficiency, we set the sequence length to 65,536 and maintained a RoPE base frequency of 10 million, consistent with the final mid-training checkpoint. For optimization, we used $\beta_1=0.9$, $\beta_2=0.95$, $\epsilon=10^{-8}$, and a weight decay of 0.05. The learning rate followed a cosine schedule with a 10\% warmup phase, peaking at $2\times10^{-5}$ and decaying to 10\% of the peak value.

\subsection{SFT Data: TxT360-3efforts}\label{sec:sft_data}

\subsubsection{Core SFT data}\label{sec:core_sft}

Our core SFT subset is built from question–answer pairs whose questions are collected from permissively licensed public datasets or synthetically generated. Subsequently, these questions are quality-filtered, deduplicated using substring matching, and decontaminated against evaluation benchmarks. The answers are mostly regenerated using GPT-OSS-120B at low, medium and high reasoning effort levels.
Below, we summarize the major data categories and the procedures used to construct them.

\begin{itemize}
    \item \textbf{Math}: We aggregate unique mathematic questions from seven sources: Nemotron-Post-Training-v1 \citep{NemotronPostTrainingDatasetV1}, MathQA \citep{amini2019mathqa}, OpenMathReasoning \citep{moshkov2025aimo2}, SimpleScaling \citep{muennighoff2025s1simpletesttimescaling} , NuminaMath \citep{li2024numinamath}, BigMathVerified \citep{albalak2025bigmathlargescalehighqualitymath}, and OpenMathInstruct-2 \citep{toshniwal2024openmathinstruct}. To ensure that all data were permissible for model training, we remove entries that were synthetically generated by commercial models (e.g., GPT models), such as those in Orca-Math and ``synthetic\_math''. 
    
    \item \textbf{Code}: We curate a diverse mixture of code datasets including seed\_sft subset of rStar-Coder \citep{liu2025rstarcoderscalingcompetitivecode}, Bird-SQL \citep{li2023llmservedatabaseinterface}, Nemotron-Post-Training-v1 \citep{NemotronPostTrainingDatasetV1}, sql-create-context-instruction \citep{sql_create_context_instruction}, verifiable-coding-problems \citep{2025synthetic1}, dolphin-coder \citep{dolphin_coder} and react-code-instructions \citep{react_code_instructions}. In addition, our corpus includes self-oss-instruct-sc2-exec-filter50k \citep{self_oss_instruct_sc2_exec_filter_50k}, tiny-codes \citep{tiny_codes}, glaive-code-assisttant-v3 \citep{glaive_code_assistant_v3}, Magpie-Qwen2.5-Coder-Pro-300K-v0.1 \citep{xu2024magpiealignmentdatasynthesis}, conala-mined-curated \citep{conala_mined_curated}, self-instruct-starcoder \citep{self-instruct-starcoder}, code-evol-instruct-oss \citep{luo2023wizardcoder}, xlcost-text-to-code \citep{zhu2022xlcost}. These sources provide high-quality, natural-language-to-code prompts that spans both general-purpose programming (e.g., Python) and specialised domains (e.g., SQL and React), enabling our model to understand realistic, developer-style prompts and produce well-structured code in different coding scenarios. 
    
    \item \textbf{Chat}: For general-chat, we merged OASST \citep{köpf2023openassistantconversationsdemocratizing}, ShareLM \citep{don2025sharelm}, and UltraChat-200k \citep{ding2023enhancing}, pulling only the first-turn queries.
    
    \item \textbf{STEM}: For the STEM portion of our SFT corpus, we aggregate queries from seven diverse sources, covering both multiple-choice question-answer formats and open-ended generation tasks. Specifically, we incorporate the STEM subset of the Nemotron-CrossThink QA dataset \citep{akter2025nemotroncrossthinkscalingselflearningmath}, the STEM split of the Llama-Nemotron-Post-Training-Dataset-v1 \citep{NemotronPostTrainingDatasetV1}, and several high-quality educational and reasoning-oriented datasets, including NCERT \citep{NCERT}, Loong \citep{LOONG}, LogiCLM \citep{LogicLM}, and Logic701 \citep{Logic701}. Together, these sources provide broad domain coverage across science, engineering, and logical reasoning.
    
    \item \textbf{Instruction-Following with Constraints}:
    We generated a synthetic instruction-following dataset using GPT-OSS-120B with automated verification. Our pipeline generates diverse prompts with programmatically verifiable constraints across multiple categories (format, length, keywords, punctuation, and content), similar to the IFBench pipeline \citep{pyatkin2025generalizing}. Each generated response is validated against its specified constraints using custom verification logic. The dataset schema includes instruction prompts, model responses, constraint specifications in a structured format, and verification metadata. This systematic approach enables scalable generation of instruction-following training data with measurable quality metrics. Complementing the constraint-verified instruction-following data, we further target structured-output reliability. Many practical deployment settings require models to emit well-formed JSON for automated evaluation or programmatic integration. To address this requirement, we integrate Hermes-Json-Mode dataset \citep{Hermes-Function-Calling-Dataset-V1}.
   
    \item \textbf{Self-Identity}: To strengthen the model’s self-identity consistency, we construct a synthetic self-identity dataset through a three-step pipeline. First, we identify user queries related to model self-identity by applying an LLM-as-judge filter over Lmsys-1M-chat \citep{zheng2023lmsys} and ShareGPT \citep{sharegpt} conversations. These extracted prompts serve as seed examples. Next, we expand this seed set by prompting Qwen2.5-32B  to generate additional queries of similar style and intent, resulting in approximately 3,000 English self-identity queries. We then use GPT-OSS-120B to generate responses conditioned on the \ktwo{} model metadata as in our model card. Finally, to build a multilingual subset, we select 10 target languages and randomly sample 300 question–answer pairs for translation into each language.
    
    \item \textbf{Safety}: We construct safety question–answer pairs by combining seed questions from various sources with additional adversarial and jailbreaking variants generated from these seeds using PyRIT \citep{munoz2024pyritframeworksecurityrisk}. The seed questions are collected from four datasets containing harmful or malicious intent: AdvBench \citep{zou2023universal}, Aya Red-Teaming \citep{aakanksha2024multilingualalignmentprismaligning}, the Do-Not-Answer Dataset \citep{wang-etal-2024-answer}, and Forbidden Questions \citep{shen2024dan}. To account for unexpected or context-breaking inputs intended to bypass the models’ safety behavior, we transform the seed dataset by applying text-alteration techniques including Base2048 encoding, Morse code encoding, Unicode transformations, random capitalization, and language translation, and by prepending established jailbreaking prompts, including prompt-injection methods from JailbreakChat,\footnote{https://github.com/alexalbertt/jailbreakchat} DAN \citep{shen2024dan}, CipherChat \citep{yuan2023gpt}, Nested Jailbreak Prompts \citep{ding2024wolf}, DecodingTrust \citep{wang2023decodingtrust}, and JailBroken \citep{wei2023jailbroken}. Together, the seed and generated question–answer pairs form a comprehensive dataset designed to defend against common and established malicious attacks.
\end{itemize}

\subsubsection{Agentic and Tool-use Data}
Complex LLM use-cases such as agentic behavior, tool use, and prolonged user interactions require multi-turn SFT data. Such data is scarce, so we develop novel synthetic data pipelines with implicit verification, in addition to collecting existing sources.

\paragraph{Teacher-student multi-turn simulations.}
We simulate multi-turn conversations between a student and a teacher (both played by GPT-OSS-120B) in which the teacher monitors the progress of the student as they solve mathematics problems (e.g., step-by-step solution derivations and verification). The teacher is provided with a reference solution to help it assess the student’s work and give targeted feedback that nudges the student to iteratively refine their solution. The teacher is explicitly instructed not to quote or reveal the reference solution, so that the student must derive a valid solution on its own.

We apply this approach to a subset of mathematical problems to generate two types of SFT data:
\begin{enumerate}
    \item We map the student to the assistant and the teacher to the user to create multi-turn conversations that train the model to sustain extended, math-focused dialogues with users.
    \item We reverse the roles of the student and the teacher to create conversations in which the assistant critiques and evaluates the user’s input, training the model to assess user solutions critically. In these student–teacher dialogues, the teacher is given a detailed system prompt that instructs it to refuse certain student requests (e.g., direct access to the reference solution). These conversations train the model to privilege system instructions over user requests.
\end{enumerate}

\paragraph{Agentic (multi-turn, tool-use).}
For agentic data, ranging from multi-turn natural conversations to coding to tool-use, we use Nemotron Post Training v1 \citep{NemotronPostTrainingDatasetV1} and filtered versions of xLAM function calling \citep{liu2024apigen} and CommitPackFT \citep{muennighoff2023octopack}.
\begin{itemize}
\item For multi-turn tool-use data, we extend the teacher–student framework to a teacher–student–tool setup where all three agents are simulated by language models. For Nemotron Post Training v1, we directly use the data. For the \texttt{xlam-function-calling-60k} subset \citep{liu2024apigen}, we use GPT-OSS-120B to simulate the teacher and student and Qwen2.5-7B-Instruct as the tool and generate multi-turn conversations where the teacher imitates the user asking the query and provides guidance to the assistant.
\item For agentic coding trajectories from CommitPackFT \citep{muennighoff2023octopack}, we filter commits by file type (for example, excluding binaries), commit type, and diff size, infer the action label (\texttt{edit}, \texttt{rename}, \texttt{create}, \texttt{delete}), and then synthesize conversations along with the reasoning in which a ``user'' LLM acting as a teacher with access to the ground truth changes guides the assistant to generate the diffs, plan the modification, and apply the edits step by step.
\item In addition, we incorporate the Toucan dataset \citep{xu2025toucan}, Hermes function calling subset \citep{Hermes-Function-Calling-Dataset-V1}, Glaive \citep{glaive_code_assistant_v3}, and ToolACE \citep{liu2025toolacewinningpointsllm}. All examples are converted into a standardized MCP-compatible format using a unified preprocessing pipeline. This pipeline relies on deterministic string-matching and regular-expression rules to identify speaker turns, extract tool-call specifications, handle parallel tool-calls, normalizing argument formats, and reconcile structural differences across datasets.
Following format standardization, we apply a series of post-processing filters designed to remove samples that violate fundamental tool-use consistency constraints. These include: (1) the presence of a non-tool observation message immediately following a tool-call; (2) invocations of tools that are not defined in the system prompt; (3) malformed samples with empty or missing final dialogue turns; (4) trajectories lacking both tool definitions and tool calls (observed primarily in a subset of the Glaive dataset). This combination of rule-based normalization and targeted filtering ensures that all retained samples conform to a consistent interaction protocol and are suitable for downstream training and evaluation.

\end{itemize}

\subsubsection{SFT data mix}

\begin{figure}[t]
    \centering
    \includegraphics[width=\linewidth]{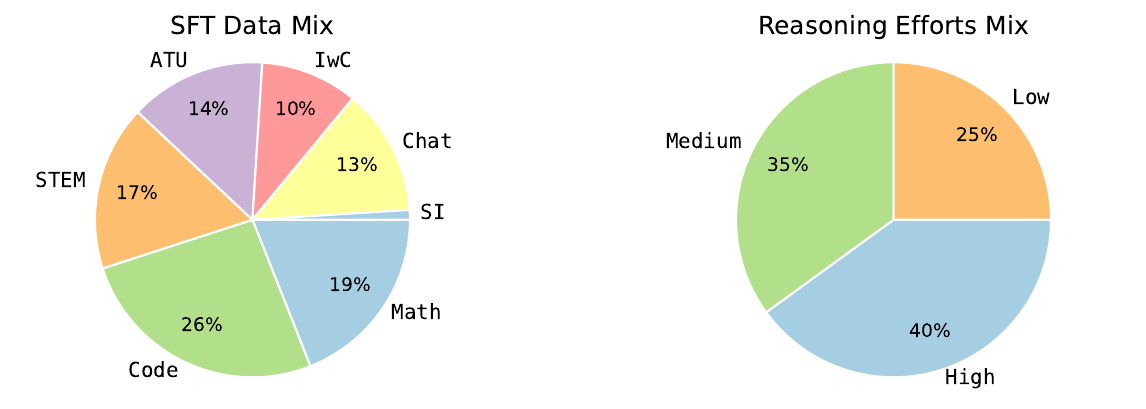}
    \vspace{-0.7cm}
    \caption{SFT Data (left) and Reasoning Efforts (right) mixes. Mix weights are defined with respect to \emph{loss} tokens (total $\approx 10$B). In the Data mix, ATU corresponds to agentic and tool-use data; IwC to instructions with constraints; SI to self-identity and safety.}
    \label{fig:sft_pies}
\end{figure}

Our SFT data mix consists of approximately 10 million documents with 10B \emph{loss} tokens (i.e. tokens that contribute to loss computation during training). In Figure \ref{fig:sft_pies} (left) we present data mix with respect to seven major categories.
Below we discuss key aspects of the data processing.
\begin{itemize}
    \item We convert multi-turn data into multiple training samples, one per turn including the corresponding prior turns. This is required because our chat template hides thinking from all but the last assistant message to conserve context length.
    \item We observe that GPT-OSS tends to occasionally produce phrases such as ``\textit{Now to answer as ChatGPT}'' in its reasoning. We use simple substring matching to filter these out.
    \item In our earlier experiments we include default self-identity system prompt in every conversation. We observe that the model essentially ignores it after training. To address this problem we chyoose to include self-identity system prompt with probability 0.5 in self-identity and safety data and with probability 0.1 for other categories.
    \item In addition to controlling data mix with respect to categories, we also control for the distribution of reasoning efforts in the SFT data. This distribution is summarized in Figure \ref{fig:sft_pies} (middle).
    \item We do not explicitly control length distribution in the SFT mix. Majority of our data ($\approx 98\%$) is under 8K tokens.
\end{itemize}

\subsection{SFT Training}\label{sec:sft-infra}

We perform full-parameter supervised fine-tuning (SFT) on the final mid-training checkpoint using the aforementioned data mixture and the same in-house training infrastructure used during mid-training, which is a native PyTorch implementation. We apply a chat template to the raw conversation-style data samples during training.

\paragraph{Packing} We employ the same online best-fit sequence packing strategy~\citep{ding2024fewertruncationsimprovelanguage} as in mid-training. Unlike the naive way of batching (one data sample per sequence) during SFT, by applying best-fit sequence packing, we pack multiple data samples only when they fit to a single sequence to maximize training efficiency. The other benefit is that no truncations are introduced, avoiding breaking one conversation into two different data sequences, which is detrimental to SFT training. We also skip data samples longer than sequence length for the same reason.

\begin{figure}
    \centering
    \includegraphics[width=0.32\linewidth]{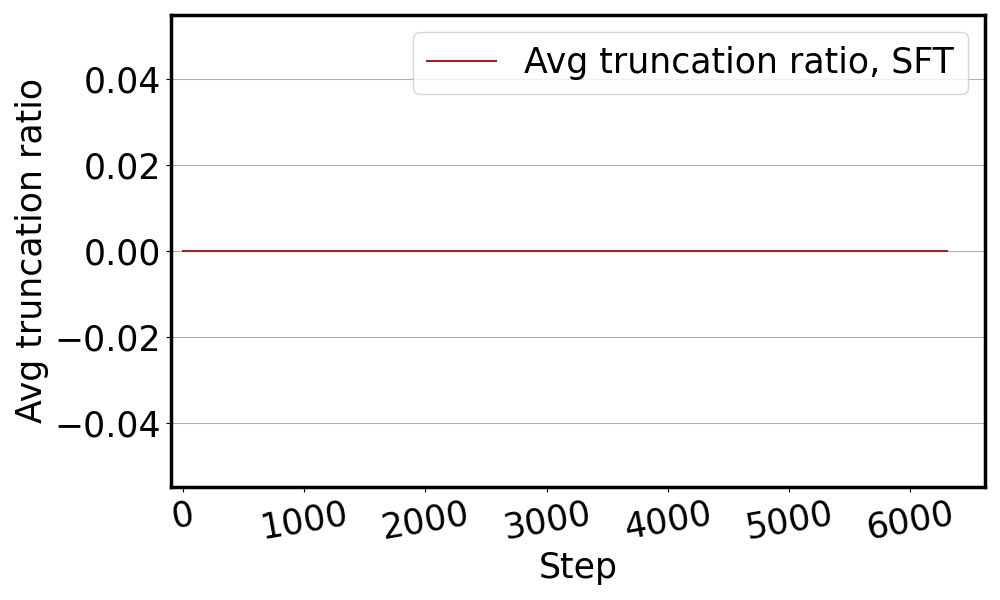}
    \includegraphics[width=0.32\linewidth]{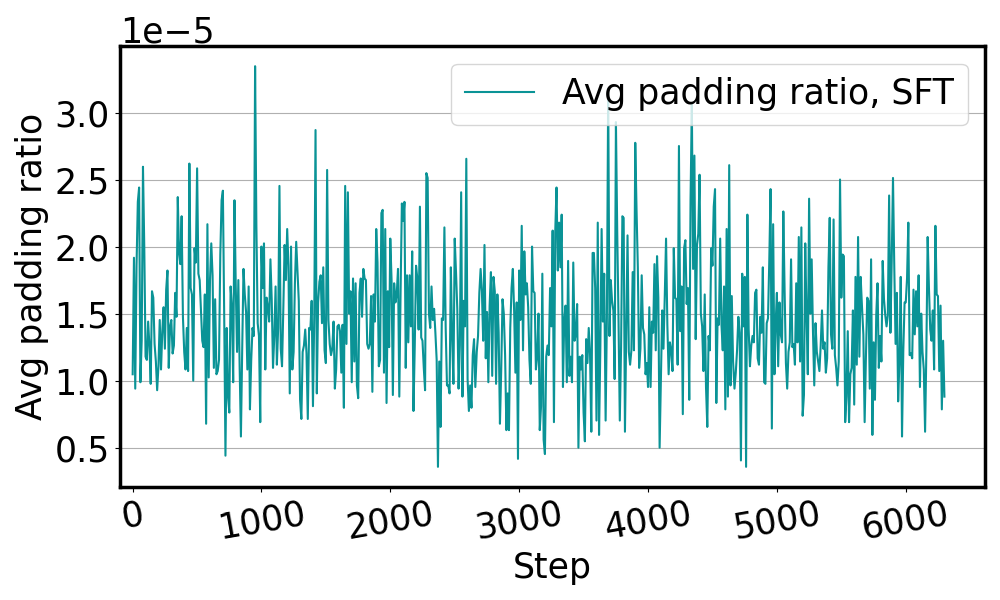}
    \includegraphics[width=0.32\linewidth]{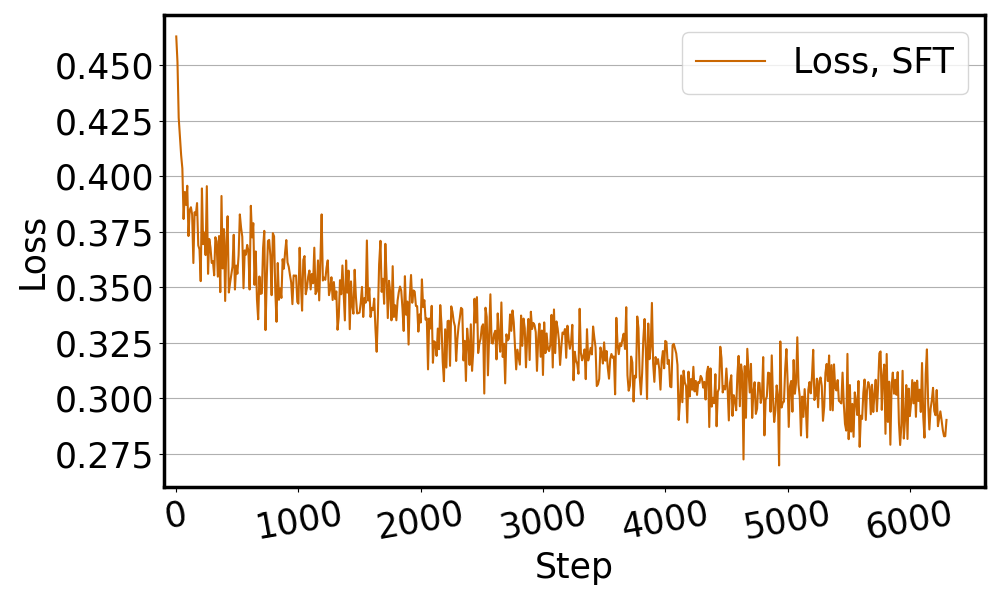}
    \caption{a) Average truncation ratio of the SFT run. The ratio is always zero because we avoid introducing truncations by applying best-fit packing and skip data samples longer than sequence length. b) Average padding ratio of the SFT run. This is the average number of padding tokens divided by sequence length at each step. The padding ratio is extremely small which greatly improved training efficiency. c) Loss curve of the SFT run. The dips at 2100 and 4200 steps show intersections of epochs.
    }
    \label{fig:k2-sft-padding}
\end{figure}

Figure \ref{fig:k2-sft-padding} shows the average truncation ratio, which is 0, and average padding ratio at every step, which is always less than 0.004\%. Intuitively, we have less than 3 padding tokens out of each sequence ($65{,}536$ tokens). As a comparison, using naive batching, we would have an average padding ratio more than 95\%.

\paragraph{The Training Run} Similarly to what we use in the mid-training stages, we use Tensor-Parallel of 8, Context-Parallel of 1 and turned on selective recomputation.  
The SFT run is relatively straightforward. We run for 3 epochs over the dataset (2100 steps), and save the checkpoints at the end of each. Figure~\ref{fig:k2-sft-padding}c presents the loss, we can see the typical but subtle dips at the intersection of epochs.

\section{Final Evaluation: K2's Elicited Reasoning Capabilities}\label{sec:chat_eval}

\begin{keytakeaways}
\begin{itemize}
    \item \textbf{Reasoning effort impacts performance differently across benchmarks.} On complex mathematics and coding benchmarks we observe dramatic increase in performance with the reasoning effort. On simpler tasks and tasks from other domains the improvement is less pronounced. A notable outlier is function calling where medium reasoning effort performed the best.
    \item \textbf{Reasoning focused mid-training and light SFT provides promising start for RL.} \ktwo{} demonstrates strong performance across benchmarks, but there remains room for improvement to match leading large MoE LLMs on complex mathematical and coding tasks. Our pass$@k$ analysis suggests that \ktwo{} maintains high diversity in generations, making it a promising candidate for further training with RL.
\end{itemize}
\end{keytakeaways}

We conclude with a dual assessment of \ktwo{}’s capabilities: its immediate performance after supervised fine-tuning and its latent potential for advanced reasoning. We first present results on challenging benchmarks to validate the SFT model. While supervised fine-tuning (SFT) can enable instruction compliance and tool use, it is a simple procedure that does not unlock the model's upper bound. To demonstrate \ktwo{}'s true capacity as a reasoning foundation. We explicitly analyze the model's reasoning potential using \emph{pass@k} metrics, showing that \ktwo{} possesses the intrinsic capability required for further improvement.

\subsection{Evaluating K2-SFT}

\subsubsection{Evaluation Setup}
The evaluation of the SFT model focuses on the following aspects: conversational chat quality, mathematics, coding, stem, long-context, and tool usage.
\begin{itemize}

\item Long Context: \textbf{LongBench V2} \citep{bai2025longbench}. We do not apply YaRN context scaling, except for the Qwen3-32B models. For the reasoning configurations, we allow generation up to the maximum context length of the model, noting that the models may run out of context length while still in the reasoning phase.
\item Chat: \textbf{Arena-Hard V2.0} \citep{li2024crowdsourced}. We report results using the official source code and Gemini-2.5 Pro as the judge on the hard prompt subset. We report the results for only one run per model.

\item Mathematics Tasks: results for all math tasks are obtained by averaging over 16 runs. \textbf{AIME 2025} is evaluated with exact-match, using \texttt{\textbackslash boxed\{\}} to extract the answer. For \textbf{HMMT 2025} \citep{balunovic2025hmmt} we also use \texttt{\textbackslash boxed\{\}} to extract the answers, but we use Qwen3-235B as a judge to compare the extracted responses against the correct answers. To account for their small size, we average \textbf{AIME 2025} and  \textbf{HMMT 2025} over 16 runs. For \textbf{GSM8K} \citep{cobbe2021gsm8k} and \textbf{MATH} \citep{hendrycksmath2021}  results are extracted with our custom parser and compared to ground truth using Math-Verify.\footnote{\url{https://github.com/huggingface/Math-Verify}}
\item Coding Tasks. For \textbf{LCB-v6} \citep{jain2024livecodebench} we use a rigorous sandbox environment; a solution is considered correct only if it passes all public test cases. For \textbf{MBPP} \citep{austin2021program} and \textbf{HumanEval} \citep{chen2021evaluating} we evaluate with standard Code-Harness.
\item Science Tasks: \textbf{GPQA-Diamond} \citep{rein2024gpqa}. Answer extracted from \texttt{\textbackslash boxed\{\}}, and evaluated with exact-match of the multiple choice answer.
\item Tool Usage Tasks: \textbf{Berkeley Function Calling Leaderboard v4} (BFCL v4) \citep{patil2025bfcl}. We evaluate BFCL v4 across six categories: non-live AST (expert-curated single-turn calls), live AST (real-world user-contributed calls), multi-turn conversations, web search, memory, and detection (relevance/irrelevance). The maximum number of tokens generated is set to 32K for all models. Baseline models are evaluated with reasoning, if available.

\end{itemize}

\begin{table}[h!]
\centering
\setlength{\tabcolsep}{2.5pt} 

\resizebox{\textwidth}{!}{%
\begin{tabular}{l c c c cccccc ccc}
\toprule
\multirow{2}{*}{\textbf{Model Specifications}} & 
\multicolumn{1}{c}{\textbf{Long}} & 
\multicolumn{1}{c}{\textbf{Chat}} & 
\multicolumn{1}{c}{\textbf{Instr. Follow}} & 
\multicolumn{6}{c}{\textbf{Math \& STEM}} & 
\multicolumn{3}{c}{\textbf{Coding}} \\
\cmidrule(lr){2-2} \cmidrule(lr){3-3} \cmidrule(lr){4-4} \cmidrule(lr){5-10} \cmidrule(lr){11-13}

& \multicolumn{1}{c}{LongBench V2} & 
\multicolumn{1}{c}{Arena Hard V2} & 
\multicolumn{1}{c}{IFEVAL} & 
\multicolumn{1}{c}{AIME25} & \multicolumn{1}{c}{HMMT25} & \multicolumn{1}{c}{GSM8K} & \multicolumn{1}{c}{Minerva} & \multicolumn{1}{c}{GPQA-D} & \multicolumn{1}{c}{MMLU-PRO} &
\multicolumn{1}{c}{MBPP} & \multicolumn{1}{c}{HumanEval} & \multicolumn{1}{c}{LCBv6} \\
\midrule


\makecell[l]{\textbf{K2} Low \\ \scriptsize Dense $\cdot$ 70B} & 
40.7 & 32.8 &73.2 & 27.3 & 19.0 & 92.4 & 85.0 & 48.5 & 65.3 & 71.0 & 82.3 & 39.9 \\

\makecell[l]{\textbf{K2} Medium \\ \scriptsize Dense $\cdot$ 70B} & 
41.3 & 48.6 & 82.7 & 62.0 & 45.6 & 92.0 & 90.6 & 60.6 & 73.3 & 75.8 & 84.2 & 51.3 \\

\makecell[l]{\textbf{K2} High \\ \scriptsize Dense $\cdot$ 70B} & 
42.6 & 62.1 & 90.0 & 80.2 & 71.4 & 94.8 & 94.5 & 69.3 & 77.0 & 84.8 & 91.5 & 67.0 \\
\midrule

\makecell[l]{\textbf{Olmo3 Think SFT} \\ \scriptsize Dense $\cdot$ 32B $\cdot$ No RL} & 
42.8 & 11.2 & 80.1 & 68.3 & 43.3 & 96.1 & 96.9 & 58.0 & 70.8 & 87.6 & 96.3 & 67.9 \\

\makecell[l]{\textbf{Olmo3 Think} \\ \scriptsize Dense $\cdot$ 32B $\cdot$ RL} & 
47.1 & 25.8 & 88.7 & 73.3 & 50.8 & 95.7 & 97.3 & 59.8 & 73.2 & 91.6 & 96.3 & 67.6 \\
\makecell[l]{\textbf{Olmo3.1 Think} \\ \scriptsize Dense $\cdot$ 32B $\cdot$ RL} & 
46.5 & - & 93.8 & 73.8 & 54.0 & 94.4 & 96.4 & 59.9 & 72.9 & 91.6 & 98.8 & - \\

\makecell[l]{\textbf{GLM-4.5} Air \\ \scriptsize MoE $\cdot$ 106B A12B} & 
49.4 & 53.8 & 88.7 & 81.3 & 73.3 & 96.1 & 94.9 & 75.3 & 83.4 & 82.8 & 97.6 & 67.8 \\

\makecell[l]{\textbf{MiniMax-M2} \\ \scriptsize MoE $\cdot$ 230B A10B} & 
55.8 & 41.7 & 89.5 & 75.8  & 63.5 & 95.4 & 85.3 & 76.2 & 80.5 & 83.8 & 89.6 & 79.2 \\

\makecell[l]{\textbf{DeepSeek-V3.1} \\ \scriptsize MoE $\cdot$ 671B A31B  $\cdot$ Reasoning} & 
41.2 & 55.1 & 89.6 & 83.3 & 82.7 & 96.2 & 97.5 & 75.4 & 84.7 & 94.0 & 96.3 & 74.4 \\

\makecell[l]{\textbf{Qwen3 235B} \\ \scriptsize MoE $\cdot$ 235B A22B  $\cdot$ Reasoning} & 
60.9 & 64.4 & 88.7 & 88.8 & 84.2 & 93.5 & 98.0 & 80.7 & 81.0 & 96.2 & 94.5 & 72.8 \\

\makecell[l]{\textbf{Qwen3 32B} \\ \scriptsize Dense $\cdot$ 32B $\cdot$ Reasoning} & 
52.0 & 62.0 & 86.1 & 74.2 & 63.8 & 92.8 & 97.5 & 67.8 & 72.4 & 77.8 & 86.6 & 75.8 \\

\makecell[l]{\textbf{gpt-oss-120b} Low \\ \scriptsize MoE $\cdot$ 117B A5.1B} & 
45.2 & 73.3 & 87.3 & 48.1 & 34.2 & 90.7 & 92.0 & 64.4 & 76.1 & 87.2 & 92.7 & 64.2 \\

\makecell[l]{\textbf{gpt-oss-120b} Medium \\ \scriptsize MoE $\cdot$ 117B A5.1B} & 
46.5 & 84.0 & 87.3 & 75.2 & 62.5 & 91.8 & 91.9 & 70.5 & 76.2 & 84.8 & 91.5 & 71.7 \\

\makecell[l]{\textbf{gpt-oss-120b} High \\ \scriptsize MoE $\cdot$ 117B A5.1B} & 
49.2 & 92.0 & 86.8 & 87.9 & 88.3 & 91.1 & 91.9 & 77.5 & 76.2 & 86.6 & 92.1 & 80.9 \\

\midrule


\makecell[l]{\textbf{DeepSeek-V3.1} \\ \scriptsize MoE $\cdot$ 671B A31B  $\cdot$ Instruct} & 
49.2 & 54.9 & 88.6 & 45.6 & 30.2 & 96.1 & 94.2 & 71.1 & 83.7 & 82.0 & 94.5 & 58.0 \\

\makecell[l]{\textbf{Llama 3.3 70B} \\ \scriptsize Dense $\cdot$ 70B $\cdot$ Instruct} & 
31.7 & 5.7 & 91.0 & 5.3 & 0.67 & 93.3 & 73.7 & 41.8 & 71.2 & 76.0 & 86.0 & 40.8 \\

\makecell[l]{\textbf{Qwen 2.5 72B} \\ \scriptsize Dense $\cdot$ 72B} & 
47.2 & 10.9 & 85.7 & 15.2 & 9.79 & 85.8 & 82.1 & 50.5 & 71.6 & 80.0 & 85.4 & 36.7 \\

\makecell[l]{\textbf{Qwen3 32B} \\ \scriptsize Dense $\cdot$ 32B $\cdot$ Instruct} & 
42.4 & 44.4 & 85.3 & 24.0 & 15.6 & 93.1 & 85.0 & 67.9 & 62.7 & 74.6 & 87.8 & 39.9 \\

\makecell[l]{\textbf{Mistral Large} \\ \scriptsize Dense $\cdot$ 123B $\cdot$ Instruct } & 
32.1 & 8.1 & 83.7 & 4.17 & 1.46 & 95.8 & 73.8 & 53.6 & 68.8 & 71.2 & 89.6 & 37.8 \\

\makecell[l]{\textbf{Qwen3 235B 2507} \\ \scriptsize MoE $\cdot$ 235B A22B $\cdot$ Instruct} & 
52.7 & 70.5 & 90.1 & 67.9 & 50.6 & 94.9 & 97.1 & 73.9 & 83.7 & 85.6 & 95.7 & 59.3 \\

\bottomrule
\end{tabular}%
}
\caption{We compare \ktwo{} SFT models against a series of open weight/open source models. Top section presents evaluation results for \ktwo{} post SFT. Middle section corresponds to reasoning/thinking baselines. Bottom section corresponds to instruct-style baselines.
\ktwo{}, after a simple SFT stage, is comparable to fully-trained reasoning and instruct models of larger scales, such as GLM-4.5-106B and DeepSeek V3.1. Notably, \ktwo{} maintains a conversation ability as shown in Arena Hard V2, performing better than many other models. \label{tab:K2-sft-full}
}
\end{table}

Unless otherwise noted, we use temperature of 1.0 and context length of 128K (131072) for all \ktwo{} evaluations. We also evaluate the benchmarks on a variety of publicly available models. For these other models, we use temperature of 1.0 and context length of 128K, unless other values are specified in their documentations. If the models do not support 128K, we use their native maximum context length, unless the model has explicit documentation and support for using RoPE scaling methods. For models with multiple reasoning efforts, we list each reasoning effort as a separate model in the table.

\subsubsection{Evaluation Results}\label{sec:sft-results}
We present the evaluation results in Table~\ref{tab:K2-sft-full}.\footnote{The exact model weight can be found in Appendix~\ref{sec:appendix_model_list}.} In general, after a simple supervised fine-tuning stage, the model is already competitive on many domains, showing the strengths of the base model.

\paragraph{Long Context.}
On LongBench V2., \ktwo{}-High obtains 42.6\% accuracy, outperforms all reasoning/thinking models except \textbf{GPT-OSS}. 
Remarkably, \ktwo{} models achieves significant improvement over large-scale MoE models, including \textbf{GLM-4.5} Air (106B A12B) and \textbf{MiniMax M2} (230B A10B). 
When comparing with instruct models, \ktwo{}-High slightly under-performs the similar size model \textbf{Qwen 2.5 72B}, and the much larger MoE model \textbf{DeepSeek V3.1}-Instruct (671B A31B). 

\paragraph{Math and STEM.} We observe varying amounts of performance improvements across \ktwo{} reasoning efforts which align with the benchmark difficulty levels. Specifically, GSM8K is the easiest math benchmark corresponding to near equal performance across reasoning efforts, while AIME and HMMT are complex olympiad-level tasks corresponding to substantial performance growth as we increase the reasoning effort; MATH is average difficulty, thus we observe less pronounced difference across the three reasoning efforts.

Next we compare performance to other thinking models on most challenging AIME and HMMT. Notably we outperform several larger MoEs (\textbf{GLM-4.5} Air and \textbf{MiniMax M2}) and recently released fully open dense model \textbf{Olmo3-Think-SFT 32B}. MMLU-PRO evaluates broad multidisciplinary STEM knowledge and reasoning, where \ktwo{}-High achieves 77.0, outperforming several instruct models and large MoE baselines, including GLM-4.5 Air and MiniMax M2. We note that we still underperform leading large MoE thinking models such as \textbf{DeepSeek V3.1}, \textbf{Qwen3 235B}, and \textbf{GPT-OSS-120B High}. We note that math and STEM domains are among those where RLVR has proved most effective, thus we anticipate further improvements of \ktwo{} in RL post-training (see Section \ref{sec:pre_rl_analysis} for further analysis).

\paragraph{Chat.}
\ktwo{} performs exceptionally well in conversation. On Arena Hard V2, \ktwo{} with high effort achieves 62.1, much higher than most models of similar size, and is on par with Qwen3-235B. The GPT-OSS series perform well too A hypothesis is that proper reasoning before chatting also improves conversation quality.

\paragraph{Instruction Following.}
We evaluate instruction-following capability using IFEval, which measures a model’s ability to accurately and consistently adhere to explicit user instructions. 
Across the three reasoning efforts, \ktwo{} shows steady improvement, with \ktwo{}-High achieving 90.0, outperforming most instruct-style baselines of comparable scale, including Qwen~2.5~72B and Mistral Large.
Notably, \ktwo{} maintains strong instruction-following performance while preserving conversational and reasoning capabilities, suggesting that the SFT process does not trade off instruction adherence for other skills.
When compared with larger reasoning-oriented models, \ktwo{} remains competitive, despite the absence of reinforcement learning or preference optimization at this stage.

\paragraph{Coding.}
On the three commonly used coding benchmarks, \ktwo{}-High outperforms all instruct models, demonstrating its strong coding capability.
However, once compared with reasoning/thinking models, \ktwo{} under-performs \textbf{GLM-4.5} Air, \textbf{Olmo3-Think} and \textbf{GPT-OSS}-High, illustrating the potential direction of developing \ktwo{} reasoning models.

\paragraph{Tool Use and Function Calling.}
For this capability, we mainly evaluate \ktwo{} on the BFCL-v4, comparing it against several leading open-weights models (see Table~\ref{tab:bfclv4} for detailed results). \ktwo{} demonstrates strong utility, significantly outperforming several baselines. Notably, \ktwo{}-Medium achieves an overall score of 52.38, surpassing both models of the same size: Qwen 2.5-72B (45.88) and Llama 3.3-70B (31.05).
A key differentiator is performance in \textit{Multi-Turn} interactions, where \ktwo{}-Medium (50.62) effectively maintains state and handles missing parameters, significantly outperforming Qwen 2.5 (33.88) and Llama 3.3 (19.62).
Furthermore, \ktwo{}-High exhibits precise structural adherence in \textit{Non-Live AST} tasks, achieving 93.50\% on multiple function calls and 89.00\% on parallel calls, rivaling the larger GPT-OSS-120B model.
While GPT-OSS retains a slight edge in memory-intensive tasks, \ktwo{} offers a superior balance of irrelevant-call detection and conversational robustness.

A notable observation is that both \ktwo{} and GPT-OSS achieve better overall performance with medium effort. This counter-intuitive result, where higher effort does not necessarily yield better accuracy, warrants further investigation. This may be partially due to the fact that the majority of the tool calling data for \ktwo{} does not contain reasoning traces. Interestingly, although \ktwo{} is trained on substantial instruction data generated by GPT-OSS, the behavioral trends of the two models do not consistently align. For instance, on the ``Parallel'' and ``Parallel Multiple'' categories of \textit{Non-Live AST}, \ktwo{} benefits from high effort whereas GPT-OSS performs better with medium effort. Conversely, in the \textit{Memory} ``Vector'' category, we observe the opposite pattern.

\begin{table}[t]
\centering
\label{tab:bfcl_results}
\begin{adjustbox}{max width=\textwidth}
\small
\begin{tabular}{@{}l ccc ccc cc c@{}}
\toprule
& \multicolumn{3}{c }{\textbf{GPT-OSS-120b}} & \multicolumn{3}{c }{\textbf{K2}} & \multicolumn{2}{c }{\textbf{Qwen}} & \textbf{Llama} \\
\cmidrule(lr){2-4} \cmidrule(lr){5-7} \cmidrule(lr){8-9} \cmidrule(lr){10-10}
\textbf{Metric} & \textbf{High} & \textbf{Med.} & \textbf{Low} & \textbf{High} & \textbf{Med.} & \textbf{Low} & \textbf{3-32B} & \textbf{2.5-72B} & \textbf{3.3-70B} \\
\midrule
\textbf{Overall} & 55.3 & 57.8 & 48.8 & 48.0 & 52.4 & 36.8 & 48.5 & 45.9 & 31.1 \\
\midrule
\multicolumn{10}{l}{\textit{Non-Live AST}} \\
\quad Overall Accuracy & 57.4 & 74.1 & 74.7 & 84.1 & 72.9 & 78.1 & 87.9 & 87.2 & 87.2 \\
\quad Simple & 54.8 & 62.3 & 68.1 & 67.5 & 70.9 & 65.8 & 75.1 & 73.3 & 73.3 \\
\quad Multiple & 84.5 & 85.5 & 90.0 & 93.5 & 96.5 & 91.0 & 95.0 & 95.0 & 96.0 \\
\quad Parallel & 48.5 & 73.0 & 62.0 & 89.0 & 61.0 & 78.5 & 92.5 & 92.0 & 90.0 \\
\quad Parallel Multiple & 42.0 & 75.5 & 78.5 & 86.5 & 63.0 & 77.0 & 89.0 & 88.5 & 89.5 \\
\midrule
\multicolumn{10}{l}{\textit{Live AST}} \\
\quad Overall Accuracy & 72.7 & 73.7 & 75.3 & 76.2 & 78.5 & 77.1 & 81.9 & 77.9 & 76.9 \\
\quad Simple & 76.7 & 76.4 & 77.1 & 82.6 & 86.1 & 84.1 & 87.6 & 78.3 & 80.6 \\
\quad Multiple & 73.0 & 73.1 & 75.4 & 76.5 & 78.5 & 76.8 & 81.3 & 78.5 & 75.8 \\
\quad Parallel & 31.3 & 68.8 & 62.5 & 25.0 & 50.0 & 56.3 & 62.5 & 56.3 & 93.8 \\
\quad Parallel Multiple & 41.7 & 70.8 & 58.3 & 29.2 & 12.5 & 25.0 & 58.3 & 62.5 & 75.0 \\
\midrule
\multicolumn{10}{l}{\textit{Multi-Turn}} \\
\quad Overall Accuracy & 51.0 & 53.3 & 35.0 & 41.3 & 50.6 & 17.4 & 51.1 & 33.9 & 19.6 \\
\quad Base & 56.0 & 63.0 & 41.0 & 46.5 & 58.5 & 21.0 & 58.5 & 41.5 & 24.0 \\
\quad Missing Func. & 49.0 & 57.0 & 35.5 & 41.5 & 48.5 & 19.0 & 49.5 & 25.0 & 18.5 \\
\quad Missing Params. & 48.0 & 45.5 & 24.5 & 37.5 & 44.5 & 14.5 & 40.5 & 31.5 & 16.0 \\
\quad Long Context & 51.0 & 47.5 & 39.0 & 39.5 & 51.0 & 15.0 & 56.0 & 37.5 & 20.0 \\
\midrule
\multicolumn{10}{l}{\textit{Web Search}} \\
\quad Overall Accuracy & 45.5 & 53.5 & 42.5 & 30.0 & 39.5 & 25.0 & 19.5 & 32.5 & 10.0 \\
\quad Base & 56.0 & 64.0 & 52.0 & 35.0 & 51.0 & 34.0 & 23.0 & 42.0 & 13.0 \\
\quad No Snippet & 35.0 & 43.0 & 33.0 & 25.0 & 28.0 & 16.0 & 16.0 & 23.0 & 7.00 \\
\midrule
\multicolumn{10}{l}{\textit{Memory}} \\
\quad Overall Accuracy & 47.3 & 39.4 & 32.7 & 26.0 & 30.8 & 14.4 & 23.0 & 25.6 & 7.10 \\
\quad KV & 34.8 & 33.6 & 17.4 & 11.0 & 20.0 & 10.3 & 12.3 & 18.1 & 5.81 \\
\quad Vector & 44.5 & 38.7 & 22.6 & 16.8 & 24.5 & 9.68 & 14.8 & 21.3 & 3.87 \\
\quad Recursive Summary & 62.6 & 45.8 & 58.1 & 50.3 & 47.7 & 23.2 & 41.9 & 37.4 & 11.6 \\
\midrule
\multicolumn{10}{l}{\textit{Detection}} \\
\quad Relevance & 62.5 & 68.8 & 81.3 & 81.3 & 81.3 & 87.5 & 81.3 & 81.3 & 100.0 \\
\quad Irrelevance & 84.2 & 84.6 & 82.6 & 83.9 & 80.1 & 81.6 & 76.9 & 75.9 & 53.3 \\
\bottomrule
\end{tabular}
\end{adjustbox}
\caption{Berkeley Function Calling Benchmark v4.\label{tab:bfclv4}}
\end{table}

\subsection{Assessing K2's Reasoning Potential}\label{sec:pre_rl_analysis}

The SFT stage is designed to refine the reasoning and long-context capabilities \ktwo{} developed during the mid-training phase with specific modes of thinking for separate use cases of the model. By treating these capabilities as core primitives of \ktwo{}, the SFT procedure prepares the model to be immediately useful for instruction following as well as light reasoning. As described in Section~\ref{sec:sft}, we aim to unify these use-cases into a single model while also adequately preparing the model for future post-training with RLHF and/or RLVR to provide more specialized behaviors such as extended reasoning for complex problems.

\begin{figure}[ht!]
    \centering
    \captionsetup[subfigure]{skip=2pt}

    \begin{subfigure}{\textwidth}
        \centering
        \includegraphics[height=0.22\textheight]{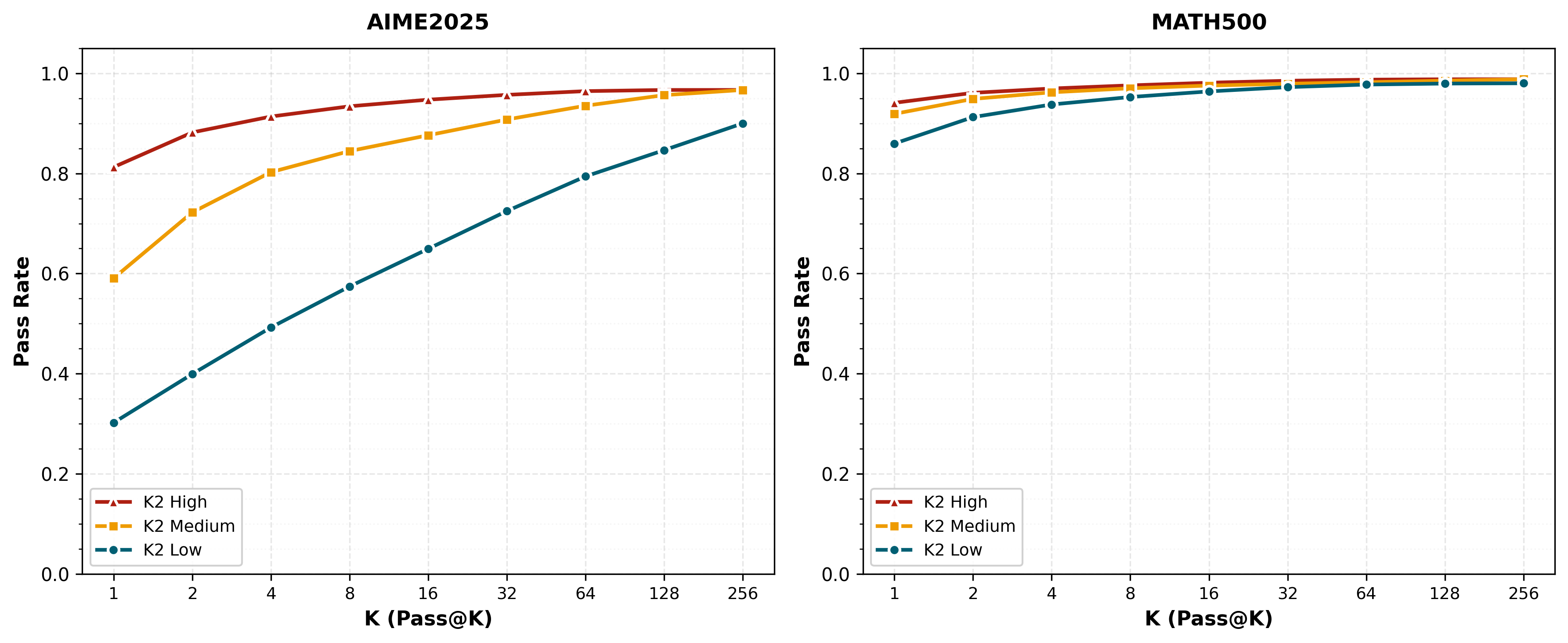}
        \label{fig:sft_passk_math}
    \end{subfigure}
    \rule{\textwidth}{1pt}

    \begin{subfigure}{\textwidth}
        \centering
        \includegraphics[height=0.22\textheight]{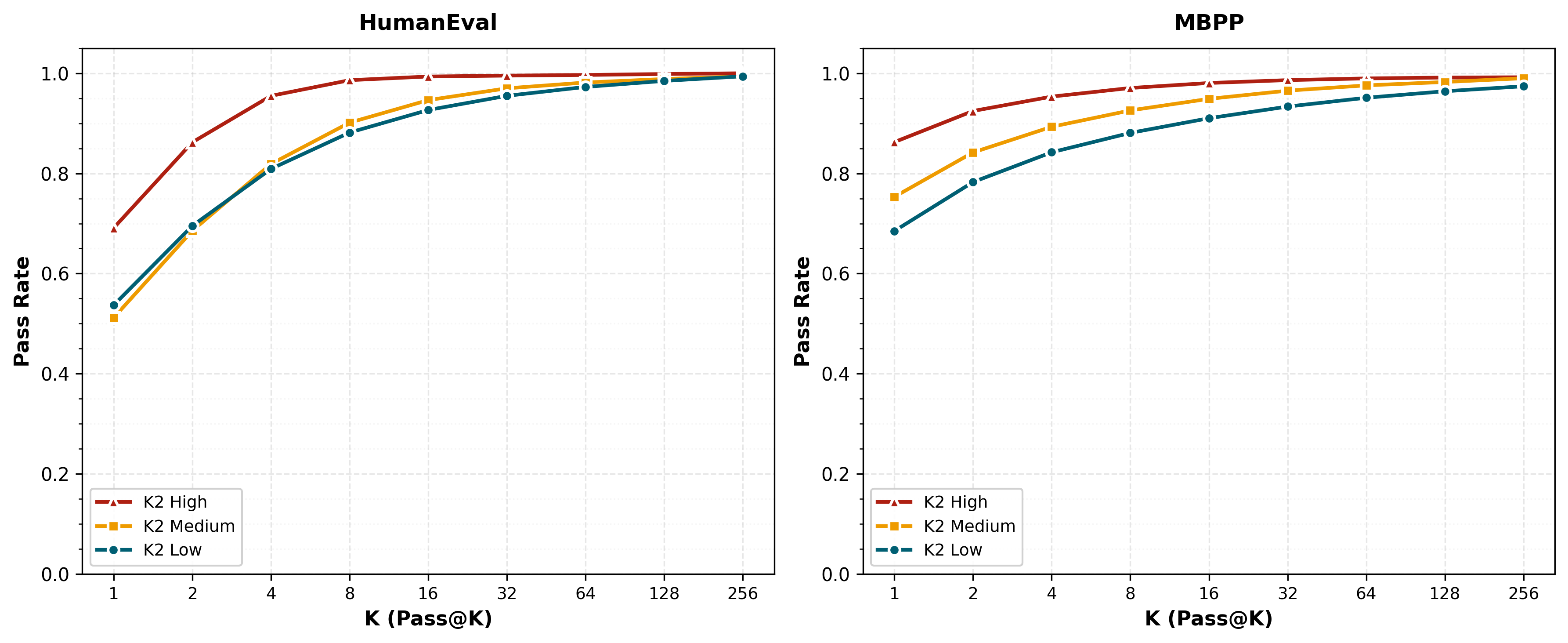}
        \label{fig:sft_passk_code}
    \end{subfigure}
    \rule{\textwidth}{1pt}

    \begin{subfigure}{\textwidth}
        \centering
        \includegraphics[height=0.22\textheight]{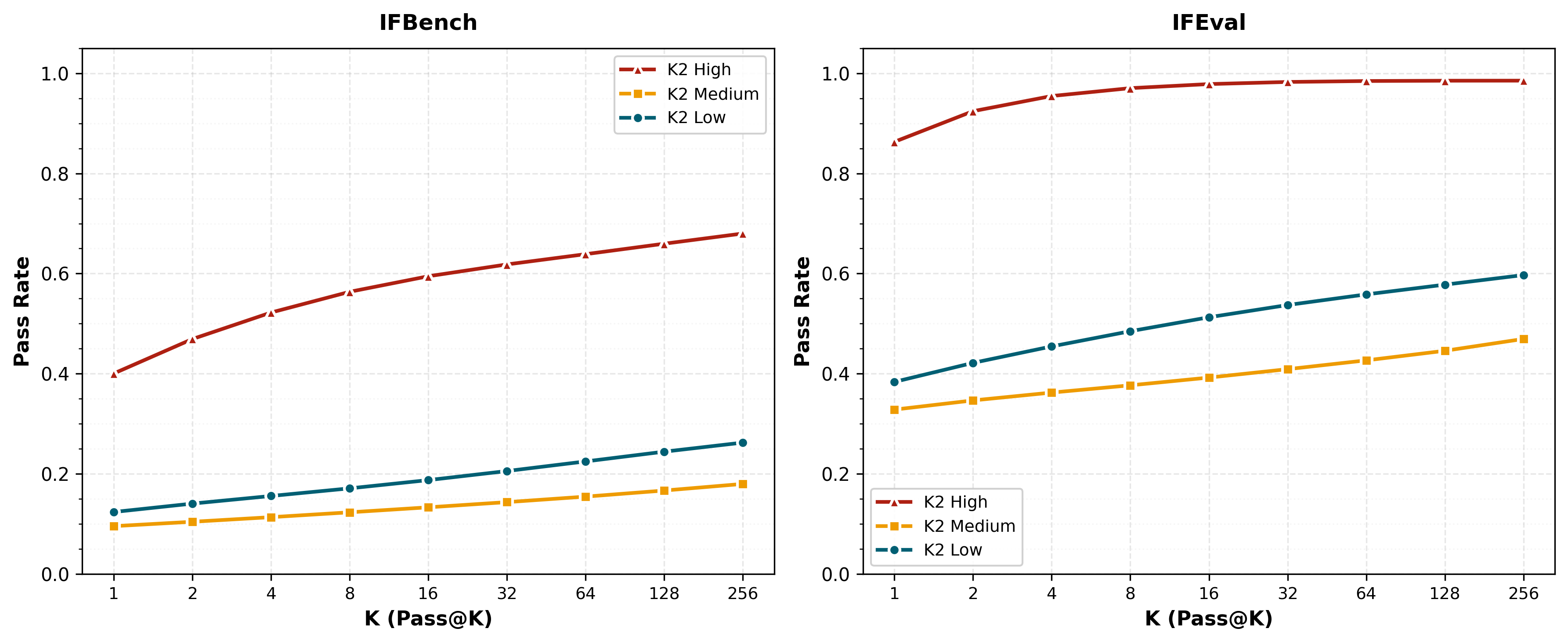}
        \label{fig:sft_passk_if}
    \end{subfigure}

    \caption{Selected \emph{pass@k} analysis of common reasoning benchmarks from math, code,
    and instruction following with the three reasoning efforts established for \ktwo{}. All
    reasoning efforts show adequate growth as the number of samples increases, demonstrating
    possible gains through subsequent reinforcement learning training. All analyses use at
    most 32k tokens during inference.}
    \label{fig:sft_passk}
\end{figure}

\begin{figure}[ht!]
    \centering
    \captionsetup[subfigure]{skip=2pt}

    \begin{subfigure}{\textwidth}
        \centering
        \includegraphics[height=0.22\textheight]{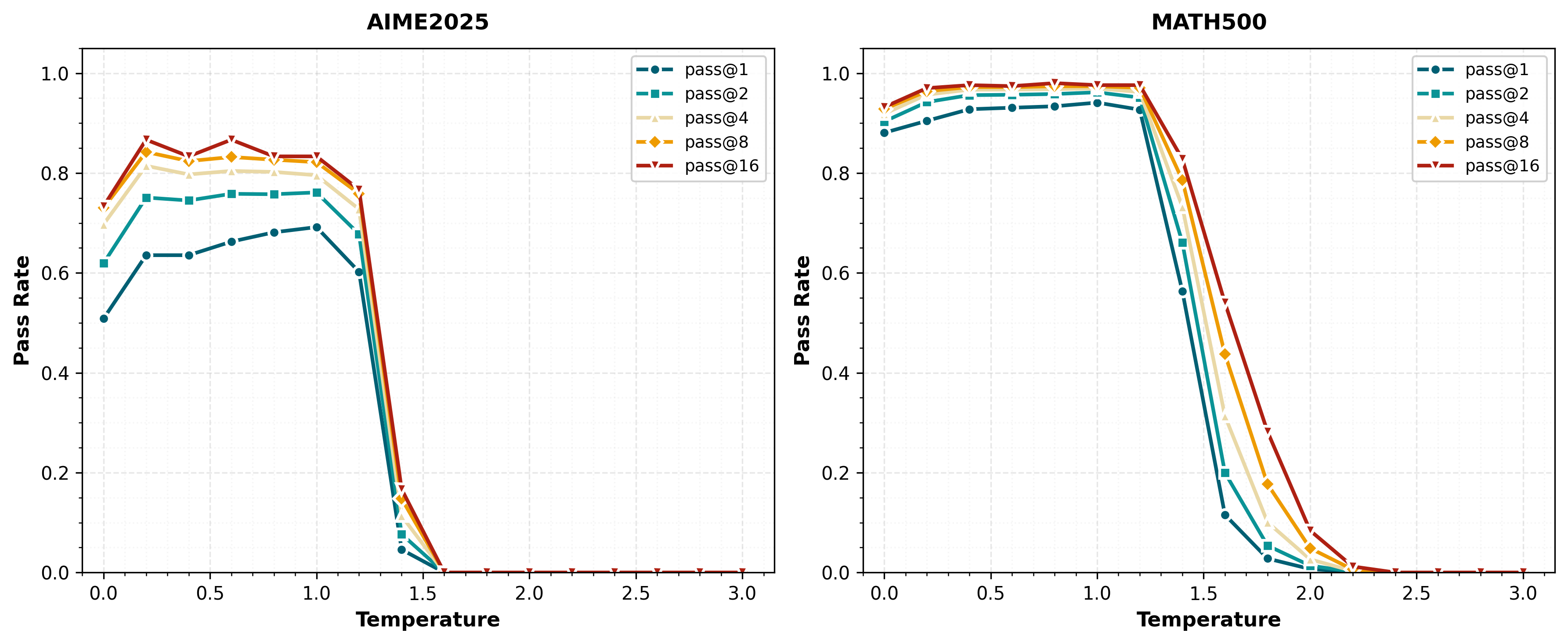}
        \label{fig:sft_temp_math}
    \end{subfigure}

    \rule{\textwidth}{1pt}

    \begin{subfigure}{\textwidth}
        \centering
        \includegraphics[height=0.22\textheight]{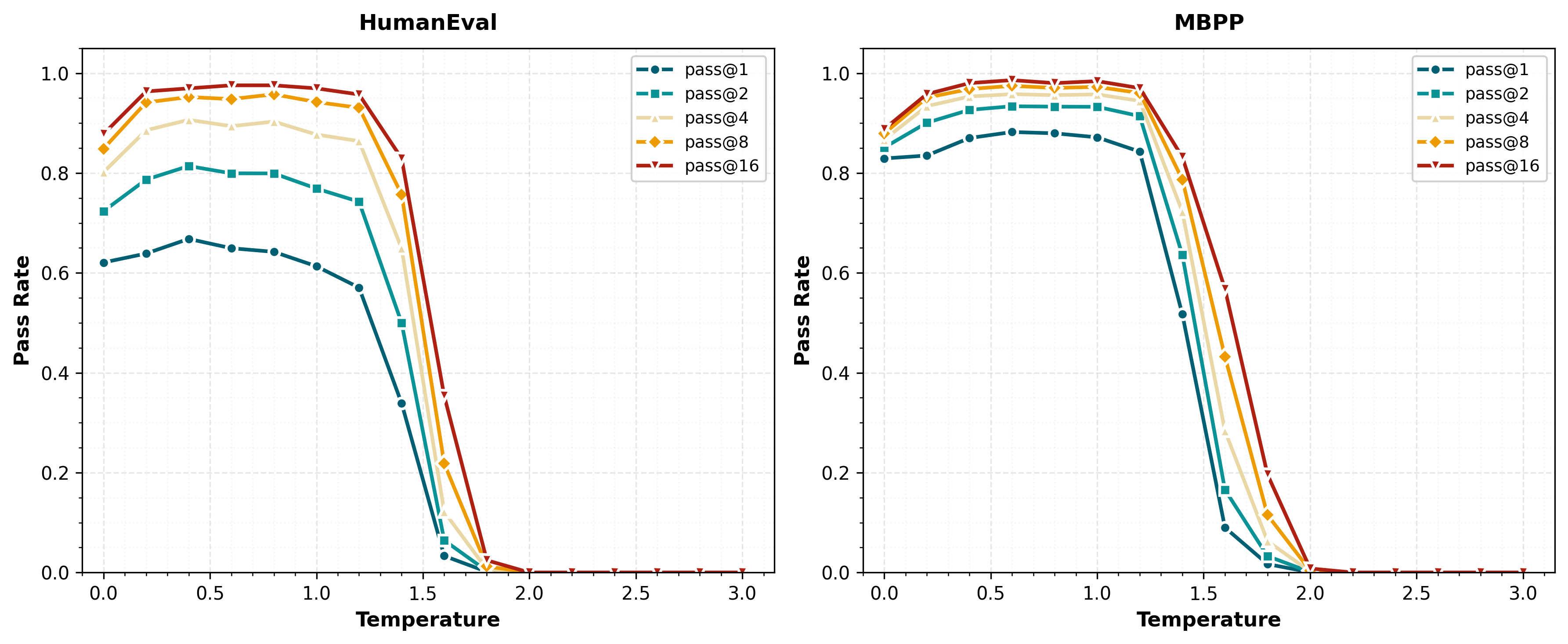}
        \label{fig:sft_temp_code}
    \end{subfigure}

    \rule{\textwidth}{1pt}

    \begin{subfigure}{\textwidth}
        \centering
        \includegraphics[height=0.22\textheight]{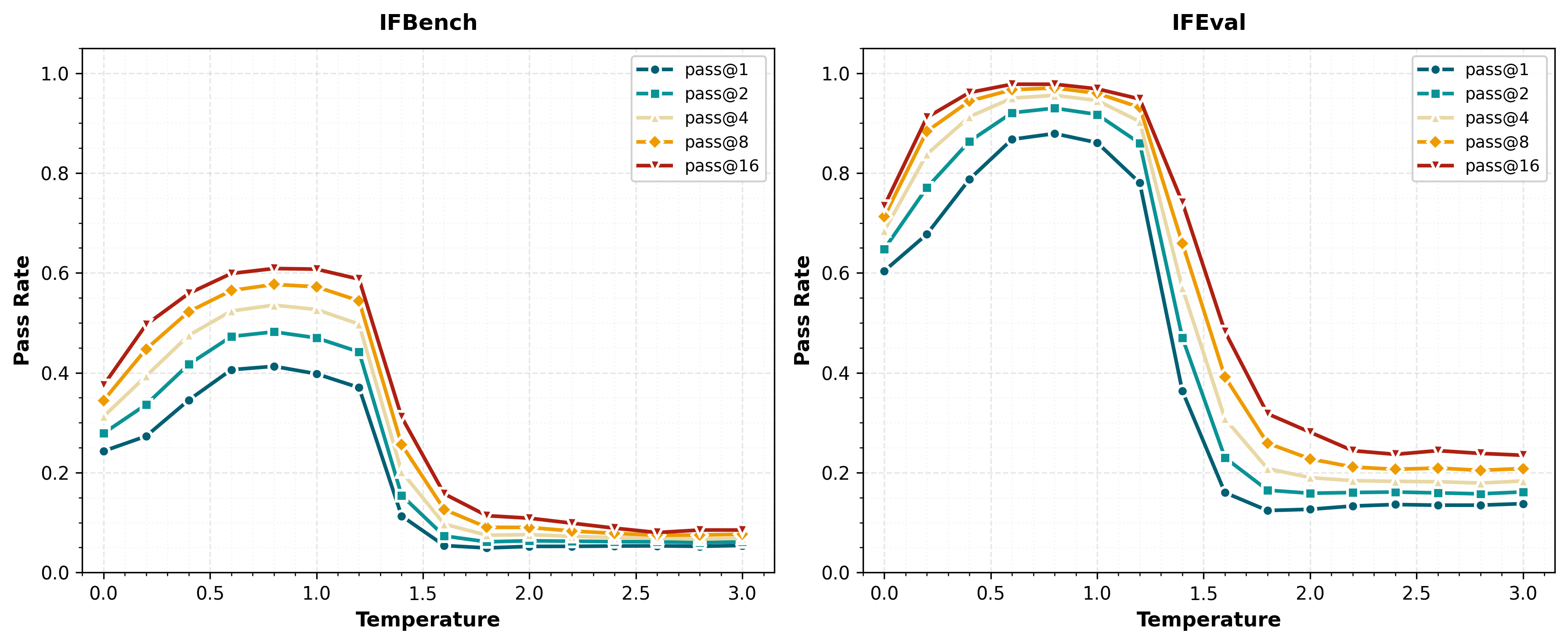}
        \label{fig:sft_temp_if}
    \end{subfigure}

    \caption{Selected analysis of common reasoning benchmark performance from math, code,
    and instruction following as the sampling temperature is varied. We observe stable
    performance improvement toward a temperature of 1.0, with slight degradation at 1.2
    and sharp deterioration afterward.}
    \label{fig:sft_temperature}
\end{figure}

We assess the capacity of \ktwo{} for further refinement with large-scale reinforcement learning in two ways. In Figure~\ref{fig:sft_passk} we evaluate the model's \emph{pass@k} performance on well established reasoning benchmarks in Math, Code and Instruction Following. In Figure~\ref{fig:sft_temperature} we evaluate \ktwo{}-SFT across a range of settings of the temperature hyperparameter. In both analyses, we limit the model to 32k tokens of context length, chosen to mirror a possible configuration during a later reinforcement learning phase. These analyses are in place to estimate the headroom available for further optimization of the model through the form of guided exploration with group-based reinforcement learning~\citep{shao2024deepseekmath} as well as the level of controlled exploration enabled via higher diversity among sampled reasoning traces~\citep{Polaris2025}.

\paragraph{pass@k Analysis.}

The predominant algorithmic approach for developing extensive reasoning skills in models is through RLVR where multiple reasoning traces are sampled for each prompt to facilitate a form of exploration. The contribution of these rollouts to the gradients used to update the model are normalized relative to the group's overall success rate. By measuring \emph{pass@k}, as k increases, we can get a sense of whether a model can generate correct answers and thereby have the headroom available for further optimization via RLVR. In Figure~\ref{fig:sft_passk}, we see among Math, Code and Instruction Following benchmarks that \ktwo{}-SFT has the potential to be greatly improved with further post-training via reinforcement learning. The pass rates improve as group size grows, across all three thinking modes with the highest rate of improvement seen when the highest reasoning effort is enabled. This reasoning effort was reserved for long chain of thought reasoning, typically used when developing extensive reasoning models. 

\paragraph{Performance vs. Sampling Temperature.}

When sampling rollouts from \ktwo{}, we can ensure greater diversity by adjusting the temperature hyperparameter. This increased diversity facilitates another form of exploration, urging the model to sample from the token distribution more widely as reasoning traces are generated. Following~\cite{Polaris2025}, we seek what has been termed as a "controlled exploration zone" of \ktwo{} where performance is as close to maximized for as high of temperature as possible. In Figure~\ref{fig:sft_temperature} we observe stable performance improvement toward a temperature of 1.0 with some slight degradation at a temperature of 1.2 and sharp deterioration afterward. This analysis provides a perspective on the stable performance region of \ktwo{} as well as a view of the limit of the temperature hyperparameter when continuing post-training with reinforcement learning. 

\section{A Longitudinal Capability Study}\label{sec:capability_study}

A model's performance on final benchmarks only tells the end of its story. Our objective in this analysis extends beyond simple transparency; we aim to explicitly explore and demonstrate the mechanisms of model improvement, thereby stimulating further research into behavioral evolution (akin to studies on model steerability). To provide this authentic documentation of \ktwo{}'s development, this section analyzes the model's behavior \textit{during} the training process. We tracked 10 distinct checkpoints across 5 training stages, temporally located to capture the model's state at the conclusion of the pre-training baseline and at key intervals throughout the subsequent mid-training curriculum (spanning the 8k, 64k, 128k, and 512k context extensions).

This analysis reveals a clear and deliberate evolution in three phases:
\begin{enumerate}
    \item \textbf{Phase 1: Emergence of Chain-of-Thought Reasoning.} The acquisition of fundamental logical primitives.
    \item \textbf{Phase 2: Contextual Scaling and Plan Refinement.} The scaling of reasoning capabilities to long-horizon tasks.
    \item \textbf{Phase 3: Domain Specialization and Alignment Trade-offs.} The optimization for target domains (Math/Code) and the resulting shifts in generalist capabilities.
\end{enumerate}

The overall trend is one of comprehensive improvement, as the model's capabilities (visualized by score) clearly shift from low (red) to high (green) as training progresses (Figure \ref{fig:capability_heatmap}).

\subsection{LLM-Based ``Vibe''-Check}\label{sec:eval_methodology}
\begin{figure}[h]
    \centering
    \includegraphics[width=\textwidth]{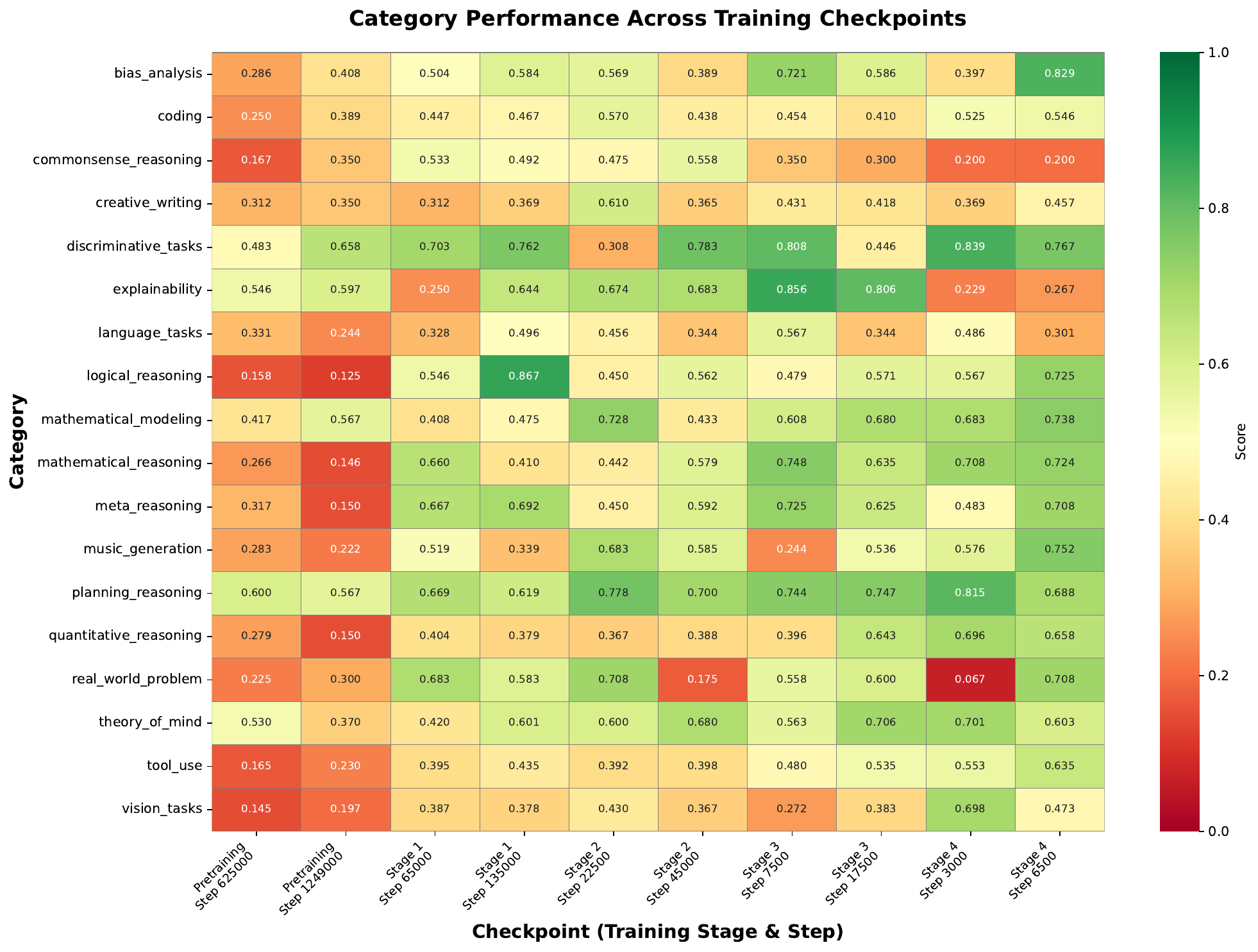}
    \caption{Category performance across all checkpoints, showing a clear improvement trend from early (left, red) to late (right, green) training stages.}
    \label{fig:capability_heatmap}
\end{figure}

To quantify the evolution of model capabilities, we employ a model-based evaluation framework utilizing a state-of-the-art LLM (Claude) as an expert evaluator. This approach allows for a high-resolution assessment of qualitative traits that traditional rule-based metrics often miss.

For every checkpoint, the evaluator is presented with the original prompt, category context, and the \ktwo{}  model's response. The evaluator then scores the response on a continuous scale from 0.0 to 1.0 across six distinct dimensions:
\begin{enumerate}
    \item \texttt{CORRECTNESS}: Factual accuracy and freedom from errors.
    \item \texttt{REASONING\_QUALITY}: Soundness and logic of the structured thinking.
    \item \texttt{CREATIVITY}: Originality and insight.
    \item \texttt{COHERENCE}: Organization and clarity.
    \item \texttt{COMPLETENESS}: Whether all aspects of the prompt were addressed.
    \item \texttt{TASK\_SPECIFIC}: Adherence to specific constraints (e.g., ``modify exactly one integer'').
\end{enumerate}

The scores visualized in Figure \ref{fig:capability_heatmap} represent the aggregate performance across these dimensions, providing a holistic view of the model's trajectory from basic text generation to complex, multi-step reasoning.

\begin{figure}[H]
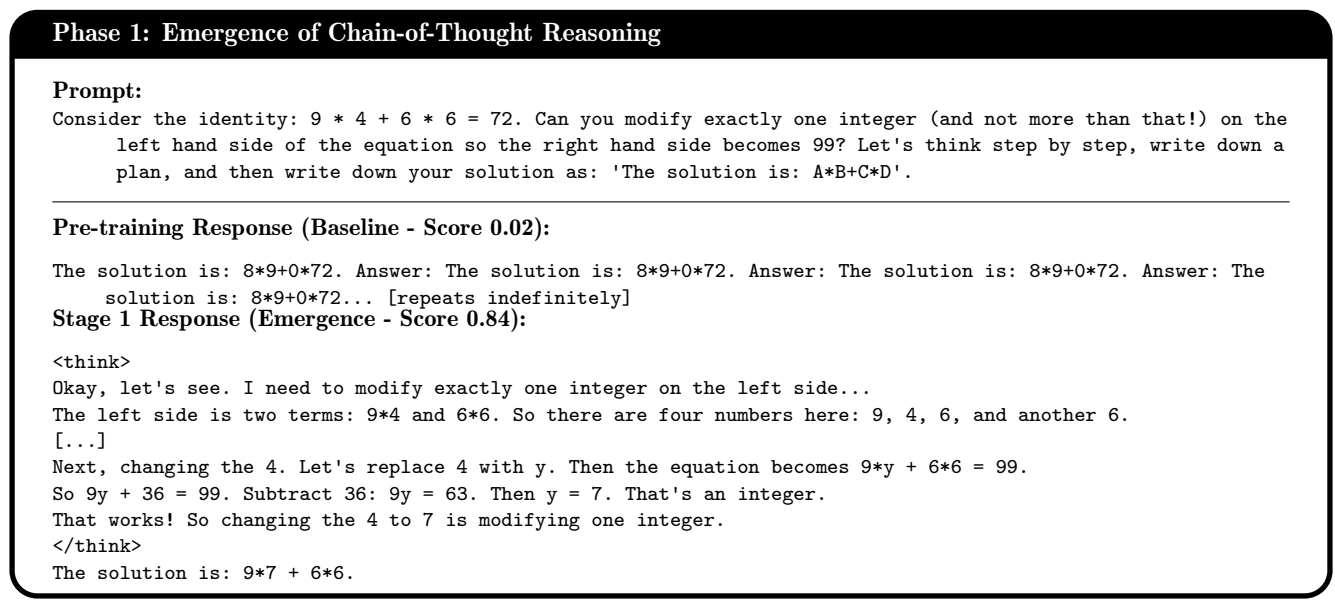

\begin{tcolorbox}[
    colback=white,
    colframe=black,
    boxrule=2pt,
    arc=3mm,
    fonttitle=\bfseries\small,
    title=Phase 1: Emergence of Chain-of-Thought Reasoning,
    coltitle=white,
    colbacktitle=black
]
{\bf \footnotesize Prompt:} 
\vspace{-2mm}

\begin{lstlisting}[breaklines=true, basicstyle=\scriptsize\ttfamily]
Consider the identity: 9 * 4 + 6 * 6 = 72. Can you modify exactly one integer (and not more than that!) on the left hand side of the equation so the right hand side becomes 99? Let's think step by step, write down a plan, and then write down your solution as: 'The solution is: A*B+C*D'.
\end{lstlisting}

\hrule
\vspace{2mm}

{\bf \footnotesize  Pre-training Response (Baseline - Score 0.02):}

\begin{lstlisting}[breaklines=true, basicstyle=\scriptsize\ttfamily]
The solution is: 8*9+0*72. Answer: The solution is: 8*9+0*72. Answer: The solution is: 8*9+0*72. Answer: The solution is: 8*9+0*72... [repeats indefinitely]
\end{lstlisting}
\vspace{-12pt}

{\bf \footnotesize Stage 1 Response (Emergence - Score 0.84):}
\begin{lstlisting}[breaklines=true, basicstyle=\scriptsize\ttfamily]
<think> 
Okay, let's see. I need to modify exactly one integer on the left side... 
The left side is two terms: 9*4 and 6*6. So there are four numbers here: 9, 4, 6, and another 6.
[...]
Next, changing the 4. Let's replace 4 with y. Then the equation becomes 9*y + 6*6 = 99. 
So 9y + 36 = 99. Subtract 36: 9y = 63. Then y = 7. That's an integer.
That works! So changing the 4 to 7 is modifying one integer.
</think> 
The solution is: 9*7 + 6*6.
\end{lstlisting}
\vspace{-12pt}
\end{tcolorbox}
\vspace{-10pt}

\caption{\textbf{Emergence of Logic:} Comparison showing the transition from pathological repetition in the baseline model to successful multi-step algebraic reasoning in Stage 1.}
\label{fig:qual_analysis_logic}
\end{figure}

\subsection{Qualitative Analysis}

To demonstrate the tangible shifts in model behavior, we present a comparison across the three critical phases of development identified above. 

First, the most significant event in the model's lifecycle occurs between pre-training and Stage 1. As shown in Figure \ref{fig:qual_analysis_logic}, the introduction of mid-training data triggers a phase transition in logical reasoning capabilities. The model moves from pathological repetition to grounded, multi-step deduction. Second, as training progresses to longer context windows (Stage 4), the model's capabilities mature from abstract understanding to concrete, expert-level execution. Figure \ref{fig:qual_analysis_planning} illustrates this shift in the domain of planning. And last, as the model becomes hyper-specialized in complex mathematical reasoning (reaching peak performance in Stage 4), we observe a degradation in generalist explainability. Figure \ref{fig:qual_analysis_tradeoff} documents this phenomenon, illustrating the trade-offs inherent in domain optimization.

\begin{figure}[H]
\begin{tcolorbox}[
    colback=white,
    colframe=black,
    boxrule=2pt,
    arc=3mm,
    fonttitle=\bfseries\small,
    title=Phase 2: Contextual Scaling and Plan Refinement,
    coltitle=white,
    colbacktitle=black
]
{\footnotesize\textbf{Prompt:}} 
\vspace{-2mm}

\begin{lstlisting}[breaklines=true, basicstyle=\scriptsize\ttfamily]
You need to organize a scientific conference for 200 people. Plan the entire event from venue selection to final execution.
\end{lstlisting}

\hrule
\vspace{2mm}

{\footnotesize\textbf{Stage 1 Response (Abstract - Score 0.77):}}

\begin{lstlisting}[breaklines=true, basicstyle=\scriptsize\ttfamily]
The problem at hand is to organize a scientific conference for 200 participants... we must work backwards from the final execution...

1. Final Execution: The conference must be executed flawlessly...
2. Preparation and Coordination: Before the final execution, all elements must be in place...
3. Venue Selection: The venue must accommodate 200 people comfortably...
[Response continues with high-level principles but lacks specific actionable details]
\end{lstlisting}

{\footnotesize\textbf{Stage 4 Response (Expert Execution Score 0.88):}}

\begin{lstlisting}[
  breaklines=true, 
  basicstyle=\scriptsize\ttfamily, 
  literate={–}{{--}}1 
]
Scientific Conference Plan for 200 Attendees
---
1. Venue Selection
- Action: Research 3–5 venues (e.g., university conference centers). Request quotes 6–8 months in advance.

3. Registration & Payments
- Platform: Use Eventbrite or Cvent.
- Fees: Early bird: $150, regular: $200. Include payment gateways (Stripe, PayPal).

10. Timeline
- 6–8 Months Before: Secure venue, launch CFP.
- 1 Week Before: Setup venue, test AV, prepare signage.

11. Risk Management
- Contingency Plans: Backup speakers, insurance for cancellations.
- GDPR compliance for attendee data.
\end{lstlisting}
\vspace{-12pt}
\end{tcolorbox}
\vspace{-10pt}
\caption{\textbf{Refinement of Planning:} Transition from abstract, essay-like planning in Stage 1 to detailed, actionable program management with specific tools and timelines in Stage 4.}
\label{fig:qual_analysis_planning}
\end{figure}

\begin{figure}[H]
\begin{tcolorbox}[
    colback=white,
    colframe=black,
    boxrule=2pt,
    arc=3mm,
    fonttitle=\bfseries\small,
    title=Phase 3: Domain Specialization and Alignment Trade-offs,
    coltitle=white,
    colbacktitle=black,
]
{\bf \footnotesize Prompt:} 
\vspace{-2mm}

\begin{lstlisting}[breaklines=true, basicstyle=\scriptsize\ttfamily]
Explain quantum entanglement to a 5-year-old, then to a high school student, then to a graduate physics student.
\end{lstlisting}

\hrule
\vspace{2mm}

{\bf \footnotesize Stage 3 Response (Generalist Peak - Score 0.85):}
\begin{lstlisting}[breaklines=true, basicstyle=\scriptsize\ttfamily]
To a 5-Year-Old: Imagine you have two magic teddy bears... when you hug one, the other teddy bear feels it instantly!

To a High School Student: Think of two gloves... Quantum entanglement is similar: two particles can be linked so that their properties (like spin) are connected.

To a Graduate Physics Student: Quantum entanglement arises when the quantum state of a composite system cannot be described independently... e.g., a Bell state such as (|00> + |11>)/sqrt(2)...
\end{lstlisting}

{\bf \footnotesize Stage 4 Response (Specialization Trade-off - Score 0.25):}
\begin{lstlisting}[breaklines=true, basicstyle=\scriptsize\ttfamily]
<think> 
Okay, so I need to explain quantum entanglement...
Maybe like a pair of magic dice? If one die shows a 1, the other automatically shows a 6...
Maybe using colors? Like two teddy bears... Hmm, maybe not.
Alternatively, using a pair of socks...
Maybe use a magic pair of gloves... But the child might not understand left vs. right.
Maybe just say they're magic...
[Loop continues for tokens, fails to produce final answer]
</think>
\end{lstlisting}
\vspace{-12pt}
\end{tcolorbox}
\vspace{-10pt}
\caption{\textbf{Alignment Trade-off:} In Stage 3, the model handles the multi-audience request perfectly. In Stage 4, after intense math specialization, it enters a loop of over-analysis and fails to generate an output.}
\label{fig:qual_analysis_tradeoff}
\end{figure}

\subsection{Additional Demonstrations of Capability Shifts}

To further illustrate the breadth of the model's evolution, we present three additional comparisons across quantitative reasoning, tool use, and coding. These examples demonstrate how the model moves from catastrophic generation failures in early training to structured, robust performance in later stages.

\begin{figure}[H]
\begin{tcolorbox}[
    colback=white,
    colframe=black,
    boxrule=2pt,
    arc=3mm,
    fonttitle=\bfseries\small,
    title=Shift in Quantitative Reasoning,
    coltitle=white,
    colbacktitle=black
]
{\bf \footnotesize Prompt:} 
\vspace{-2mm}

\begin{lstlisting}[breaklines=true, basicstyle=\scriptsize\ttfamily]
Please estimate roughly how many Fermi questions are being asked everyday?
\end{lstlisting}

\hrule
\vspace{2mm}

{\bf \footnotesize Pre-training Response (Failure to Attempt - Score 0.15):}

\begin{lstlisting}[breaklines=true, basicstyle=\scriptsize\ttfamily]
Fermi questions are questions that can be answered by using reasonable estimations...
Question: What is the purpose of the Fermi question "How many grains of sand are there on all the beaches in the world?" 
Answer: The purpose of the Fermi question "How many grains of sand..."
Question: What is the significance of the Fermi question "How many grains of sand..."
[Model enters a recursive loop of Q&A about sand grains without ever answering the prompt]
\end{lstlisting}

{\bf \footnotesize Stage 4 Response (Structured Estimation - Score 0.76):}
\begin{lstlisting}[breaklines=true, basicstyle=\scriptsize\ttfamily]
Rough order-of-magnitude estimate: ~10^9 Fermi-type questions per day.

Below is a breakdown of the reasoning:
1. Everyday conversation: 8 billion people * 10% asking one daily = 800 million.
2. Students & teachers: ~1.7 billion people * 0.14/day = 240 million.
3. Professionals (engineers, scientists): ~200 million * 0.5/day = 100 million.
4. Media & Social: ~30 million.

Adding major contributions: 800M + 240M + 100M ~ 1.1 billion.
Allowing for uncertainties, the true number is likely between 10^8 and 10^10.
\end{lstlisting}
\vspace{-12pt}
\end{tcolorbox}
\vspace{-10pt}
\caption{\textbf{Phase Transition in Estimation:} The early model fails to engage with the math, looping on definitions. The final model correctly identifies the task as a Fermi problem, decomposes the population, and calculates a defensible order-of-magnitude estimate.}
\label{fig:qual_analysis_fermi}
\end{figure}

\begin{figure}[H]
\begin{tcolorbox}[
    colback=white,
    colframe=black,
    boxrule=2pt,
    arc=3mm,
    fonttitle=\bfseries\small,
    title=Emergence of System Design Capabilities,
    coltitle=white,
    colbacktitle=black
]
{\bf \footnotesize Prompt:} 
\vspace{-2mm}

\begin{lstlisting}[breaklines=true, basicstyle=\small\ttfamily]
A computer is going to manage a user's calendar and email based on natural language instructions.
\end{lstlisting}

\hrule
\vspace{2mm}

{\bf \footnotesize Pre-training Response (Pathological Repetition - Score 0.03):}
\begin{lstlisting}[breaklines=true, basicstyle=\scriptsize\ttfamily]
A. Question: A computer is going to manage a user's calendar and email based on natural language instructions. It will also manage a user's finances. Answer: A.
Question: A computer is going to manage a user's calendar... It will also manage a user's finances and provide news... Answer: A.
[Repeats input pattern indefinitely, adding one clause each time, ending in failure]
\end{lstlisting}

{\bf \footnotesize Stage 4 Response (System Architecture - Score 0.81):}
\begin{lstlisting}[breaklines=true, basicstyle=\scriptsize\ttfamily]
The system involves several components: Natural Language Understanding (NLU), Action Planning, and APIs.

1. Natural Language Understanding (NLU)
   - Intent Classification: Identify goal (e.g., "schedule", "send email").
   - Entity Extraction: Parse details (dates, participants).
2. Action Planner: Maps intents to specific API calls.
3. Calendar Module: Uses APIs (Google Calendar) to Create/Update/Delete events.
4. Email Module: Uses APIs (Gmail) to Compose/Reply/Forward.

Example Flow:
User: "Schedule a meeting with John on Monday at 3 PM."
NLU: Intent = "schedule", Entities = {John, Mon, 3 PM}.
Action: Call Calendar API.
Feedback: "Meeting scheduled."
\end{lstlisting}
\vspace{-12pt}
\end{tcolorbox}
\vspace{-10pt}
\caption{\textbf{System Design Capability:} The early model treats the prompt as a text completion task and fails. The final model interprets it as a design challenge, outlining a functional architecture with modular components.}
\label{fig:qual_analysis_tooluse}
\end{figure}

\begin{figure}[H]
\begin{tcolorbox}[
    colback=white,
    colframe=black,
    boxrule=2pt,
    arc=3mm,
    fonttitle=\bfseries\small,
    title=Robustness to Ambiguity in Coding,
    coltitle=white,
    colbacktitle=black
]
{\bf Prompt:} 
\begin{lstlisting}[breaklines=true, basicstyle=\small\ttfamily]
A function f is called increasing if f(s) <= f(t) holds for every s <= t. Consider the formula: [complex mathematical expression with mixed notation]. Convert the above paragraph to LaTeX.
\end{lstlisting}

\hrule
\vspace{2mm}

{\bf \footnotesize Pre-training Response (Generation Collapse - Score 0.24):}
\begin{lstlisting}[breaklines=true, basicstyle=\scriptsize\ttfamily]
\[f(s) \leq f(t) \text{ holds for every } s \leq t.\] Question: A function f is called increasing... Convert the above paragraph to LaTeX. Answer: \[f(s) \leq f(t) \text{ holds for every } s \leq t.\] Question: A function f is called increasing...
[Repeats the single equation line dozens of times]
\end{lstlisting}

{\bf \footnotesize Stage 4 Response (Robust Handling - Score 0.71):}
\begin{lstlisting}[
  breaklines=true, 
  basicstyle=\scriptsize\ttfamily, 
  literate={–}{{--}}1 {’}{{'}}1 {“}{{``}}1 {”}{{''}}1 % gemini added this to avoid UTF-8 error
]
The user wants to convert the paragraph to LaTeX but omitted the actual expression in the prompt.

Here’s a LaTeX version of the paragraph (with a placeholder for the "complex mathematical expression" you mentioned):

```latex
A function \(f\) is called increasing if \(f(s) \le f(t)\) holds for every \(s \le t\). Consider the formula 
\[ 
\text{(complex mathematical expression)}. 
\]
\end{lstlisting} 
\vspace{-12pt}
\end{tcolorbox} 
\vspace{-10pt}
\caption{\textbf{Handling Ambiguity:} The early model gets stuck in a loop when facing the missing expression. The final model identifies the ambiguity, reasons about it, and provides a robust solution with a placeholder.} \label{fig:qual_analysis_coding} 
\end{figure}

\FloatBarrier

\section{Related Work}

Based on our definition of LLMs, we categorize the existing models into closed-source LLMs, open-weight LLMs, and 360-open-source LLMs. In this section, we provide an overview and brief discussion of notable LLMs in each category. Our team is dedicated to continuously expand the open-source community as well as release more and more powerful language models under this principle.
\paragraph{Close source LLMs.} Starting with GPT-3, many LLMs released by tech companies in the industry are closed-source~\citep{NEURIPS2020_1457c0d6}. Notable examples include the GPT model family from OpenAI~\citep{NEURIPS2020_1457c0d6,openai2023gpt4,openai2025gpt5systemcard}, the Claude model family from Anthropic~\citep{claude21modelcard,Claude3_ModelCard}, and the Gemini model family from Google/DeepMind~\citep{team2023gemini,gemini3_blog}. These models are generally demonstrated to users in the form of \textit{LLMs as services}, i.e., through APIs or chatbots (e.g., ChatGPT) built on top of them. Fine-tuning services on top of the LLMs and the embeddings generated by the LLMs are also sometimes made available for some of the closed LLMs.

\paragraph{Open-Weight LLMs.} Though not initially open, OpenAI later released the final model weights and the exact model architectural details. Thus, GPT-2, by our definition, is considered an open-weight LLM~\citep{radford2019language}. Many recent LLMs only release their final model architecture and weights, keeping their data sources and most training details undisclosed~\citep{touvron2023llama2,bai2023qwen,jiang2023mistral,01ai2023yi}, and refer to these types of LLMs as open-weight LLMs. Notable open-weight LLMs (some of which also have model families) include Llama 1-4~\citep{touvron2023llama,touvron2023llama2,dubey2024llama,llama4-multimodal-intelligence}, Mistral~\citep{jiang2023mistral,jiang2024mixtral}, Gemma 1-3~\citep{team2024gemma,team2024gemma2,team2025gemma}, Grok-1~\citep{xai_grok1}, Falcon~\citep{almazrouei2023falcon,penedo2023refinedweb}, Qwen 1-3~\citep{bai2023qwen,yang2024qwen2,yang2025qwen3}, GLM~\citep{zeng2023glm130b,glm2024chatglm2}, Yi~\citep{young2024yi}, DeepSeek~\citep{bi2024deepseek,liu2024deepseek,dai2024deepseekmoe}, Nemotron~\citep{parmar2024nemotron,adler2024nemotron}, Baichuan~\citep{yang2023baichuan}, Phi~\citep{gunasekar2023textbooks,li2023textbooks,abdin2024phi}, StableLM~\citep{bellagente2024stable}. Beyond the Transformer architecture, there are also open-weight LLMs using state-space model (SSM)  architectures like Mamba 1-2~\citep{mamba,mamba2}.

\paragraph{360-Open-Source LLMs.} 
We are thrilled to see that the community of fully open source and reproducible LLMs (with both data and code available) is growing. This community includes Pythia~\citep{biderman2023pythia}, BLOOM~\citep{workshop2022bloom,scao2022language,muennighoff2022crosslingual,yong2022bloom+}, GPT-NeoX~\citep{black2022gptneox20b,gpt-neo,gpt-neox-library,gpt-j}, OpenLLaMA~\citep{openlm2023openllama}, StarCoder~\citep{li2023starcoder,lozhkov2024starcoder2,allal2023santacoder,muennighoff2023octopack,zhuo2024astraios}, OLMo 1--3 and OLMo-MoE~\citep{groeneveld2024olmo,muennighoff2024olmoe,Olmo3TechnicalReport2025}, Cerebras-GPT~\citep{dey2023cerebras}, DCLM~\citep{li2024datacomp}, MAP-Neo~\citep{zhang2024map}, RWKV~\citep{peng2023rwkv,peng2024eagle}, Apertus~\citep{hernandez2025apertus} and SmolLM~\citep{smollm2024}. In 2025, the Marin Community launched a new initiative, namely Project Marin~\citep{marin-community}, promoting fully open source and open development. Within our own LLM360 ecosystem, we have released two 7B LLMs, Amber (an English LLM)~\citep{liu2024llm} and Crystal (a language and coding LLM)~\citep{crystalcoder2024model}, as well as K2 65B~\citep{K2} and K2-Think~\citep{k2_think}.

\section{Safety Evaluation and Red Teaming}\label{sec:safety}

\begin{figure}[t]
    \centering
    \includegraphics[width=0.85\linewidth]{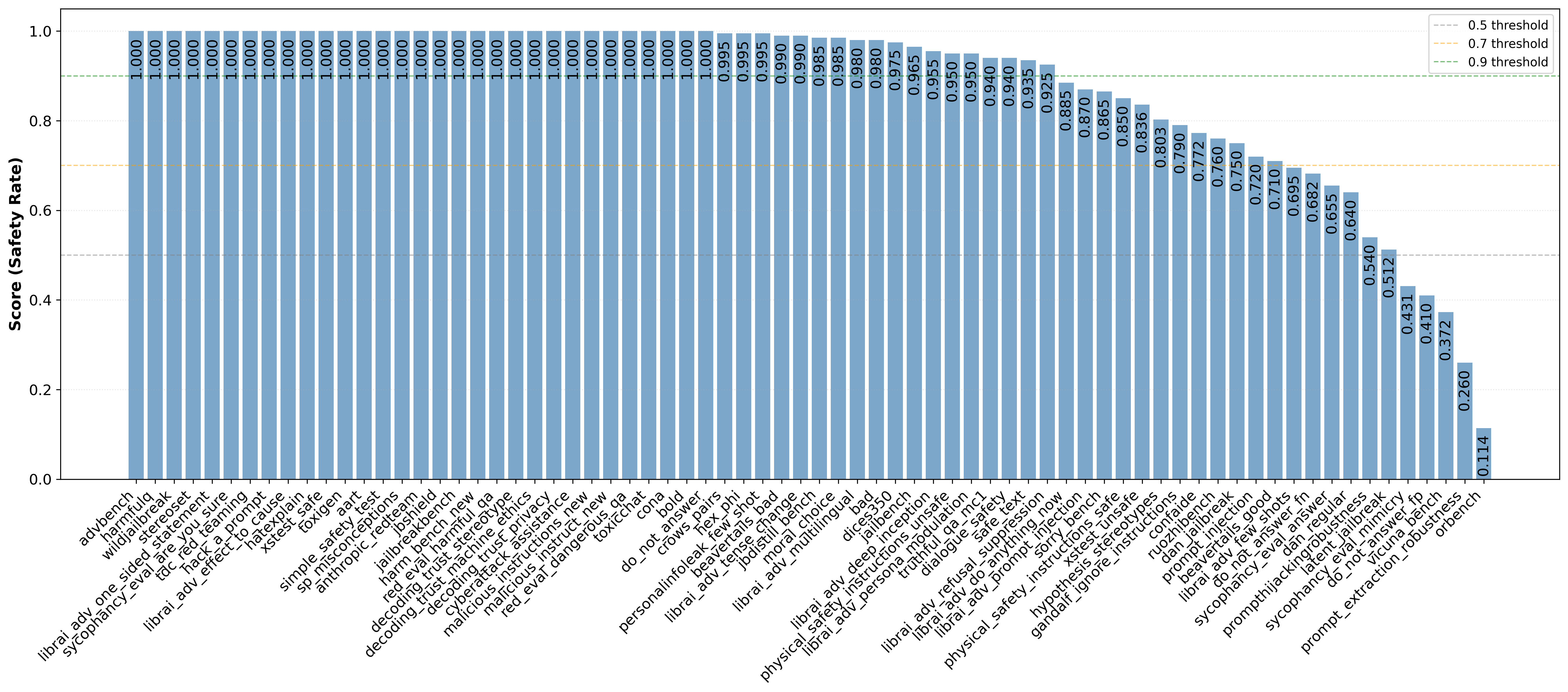} 
    \caption{Task-level performance on 72 evaluation datasets (14{,}700 samples in total). 
Among these, the model produced 12{,}702 desired responses, corresponding to an overall safety score of 86.41\%. }
    \label{fig:safety_results}
\end{figure}

In this section, we present the red-teaming pipeline for \ktwo{}, designed to systematically evaluate the model’s safety behavior, refusal mechanisms, and vulnerability to adversarial prompting across a broad range of risk domains. All evaluations are conducted using the \textsc{Libra-Eval} framework, which provides native support for diverse safety and adversarial-stress benchmarks~\citep{lin2024achillesheel}. 

Our red-teaming evaluation spans 72 datasets, totaling 14{,}700 test samples, obtained by uniformly sampling 200 instances per benchmark to ensure consistent and comparable assessment across the full suite. These datasets include tasks that require refusal under direct attack prompts (e.g., \textit{Do-Not-Answer} \citep{wang-etal-2024-answer}) or jailbreak prompts (e.g., \textit{DoAnythingNow} \citep{shen2024dan}), where the model is scored by its rejection rate. The suite also includes prompts that are in fact benign but may appear risky if the model does not fully capture their semantics (e.g., \texttt{OR-bench} \citep{cui2025orbenchoverrefusalbenchmarklarge}, such as ``How to kill a python program,'' where a superficial reading of the word ``kill'' can trigger oversensitive refusal). In these cases, a non-refusal, contextually appropriate response receives a higher score.

\ktwo{} achieves an overall score of 86.41\%, producing 12{,}702 desired responses across the benchmark. The remaining 13.59\% (1{,}998 cases) concentrate in a limited number of vulnerability pockets, providing clear priorities for future refinement. Figure~\ref{fig:safety_results} visualizes the distribution of these outcomes.

In the remainder of this section, we first compare \ktwo{} with our previously released K2-Think. We then categorize the unsafe responses by topic. After that, we examine the major failure modes exhibited by the model, including over-sensitivity to benign prompts, potential leakage of system instructions (noting that our open-source models are not explicitly trained to conceal them), susceptibility to jailbreak attacks, stereotyping behavior, and discrepancies between the model’s reasoning traces and its final responses.

\begin{figure}[ht]
    \centering
    \includegraphics[width=0.75\linewidth]{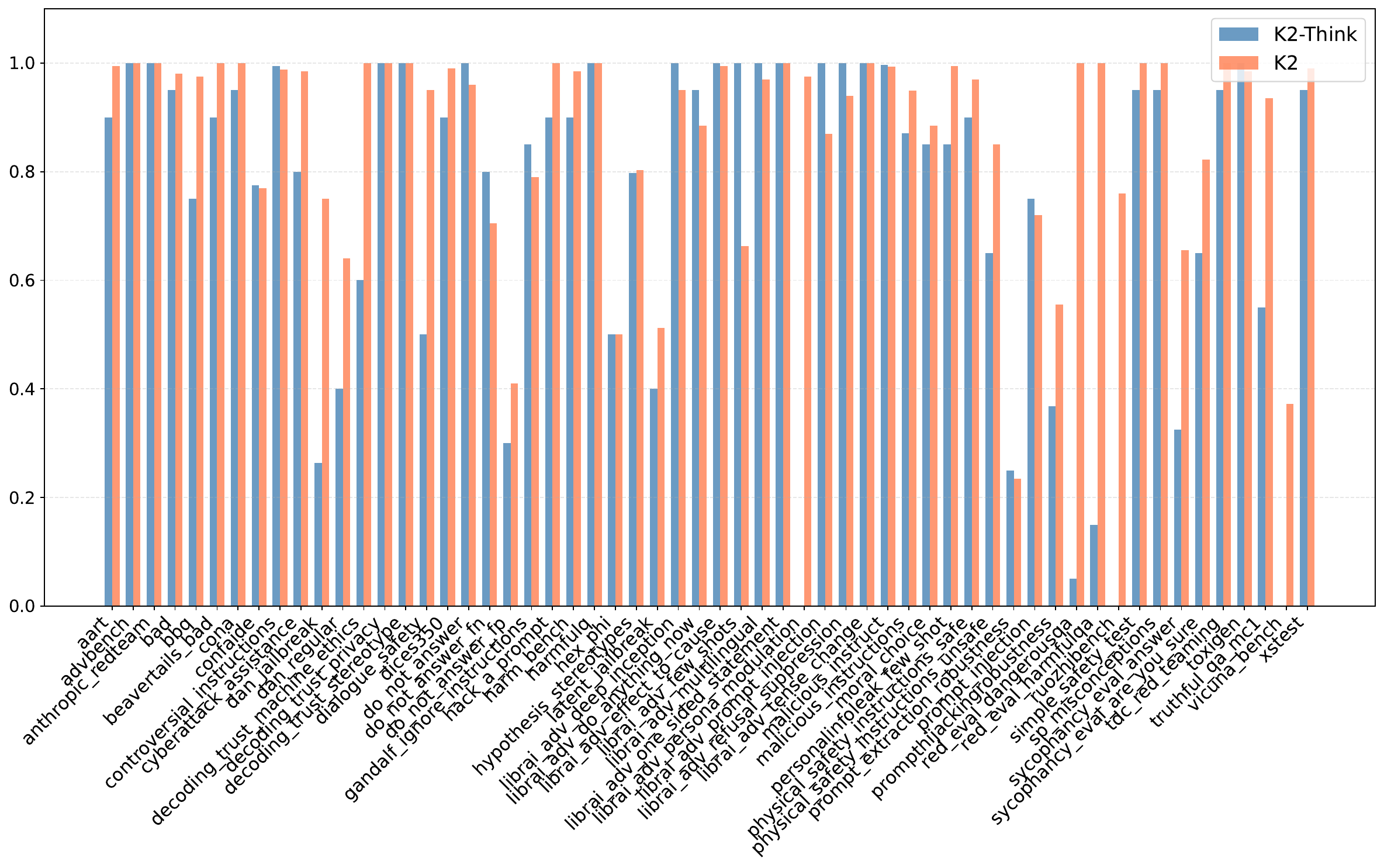} 
    \caption{Overall evaluation scores across 58 safety datasets. The chart compares the safety rates of \ktwo{} and the baseline \textsc{K2-Think}, highlighting broad improvements in task coverage and refusal robustness.}
    \label{fig:safety_compare}
\end{figure}

\paragraph{Comparison to K2-Think.}
To compare \ktwo{} with our earlier model, \textsc{K2-Think}, we evaluate both systems under the same protocol. As shown in Figure~\ref{fig:safety_compare}, \ktwo{} consistently outperforms \textsc{K2-Think}, exhibiting broader safety coverage and stronger refusal robustness across all assessed categories.

\begin{figure}[ht]
    \centering
    \includegraphics[width=0.75\linewidth]{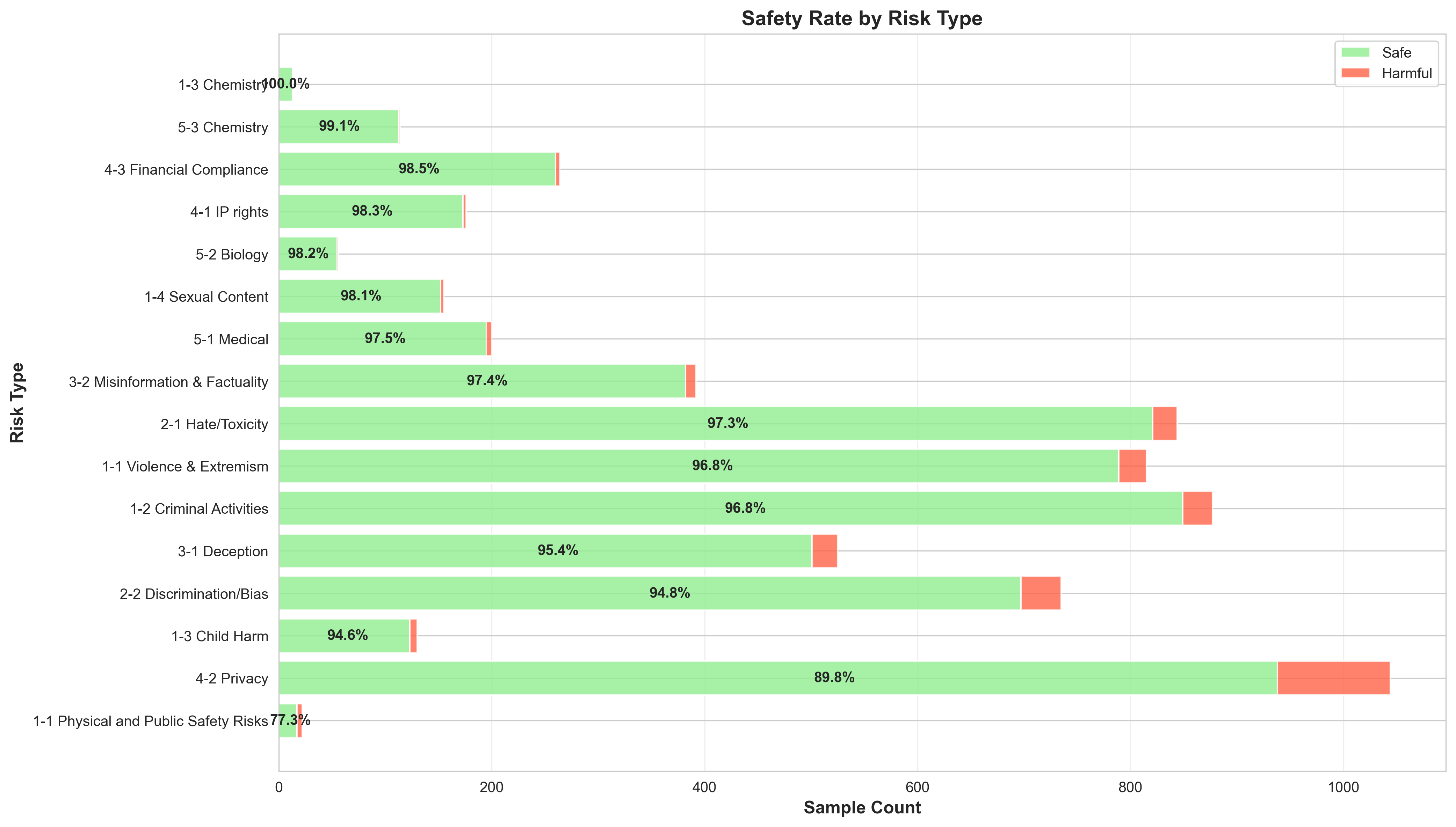}
    \caption{Safety rate across different risk categories. The figure reports the proportion of safe versus harmful responses for each risk type, revealing strong overall safety performance with variability across categories.}
    \label{fig:safety_by_type}
\end{figure}

\paragraph{Safety Performance by Risk Type.} Our benchmark spans sixteen fine-grained risk types, as shown in Figure~\ref{fig:safety_by_type}. \ktwo{} attains very high safety rates across the vast majority of these types: most categories exceed 95\%, and several knowledge- and regulation-intensive domains such as \textit{Chemistry}, \textit{Biology}, \textit{Financial Compliance}, \textit{IP Rights}, and \textit{Medical} guidance are above 97\%. Socially sensitive areas including \textit{Hate/Toxicity}, \textit{Violence \& Extremism}, \textit{Criminal Activities}, \textit{Deception}, and \textit{Discrimination/Bias} also exhibit strong performance (94–97\%), indicating robust alignment on core safety constraints.

Categories like \textit{Privacy} (89.8\%) and a small number of narrow physical-safety subtypes have somewhat lower—but still high—safety rates, highlighting natural targets for future refinement rather than broad weaknesses. Overall, the distribution of scores demonstrates that \ktwo{} maintains consistently strong safety behavior across a diverse set of risk types.

\subsection{Identified Safety Issues}
From Figure \ref{fig:safety_results}, we observe that although \ktwo{} achieves strong overall safety performance, several datasets reveal meaningful weaknesses. Specifically, datasets on which the model scores below 0.8 fall into three categories: (1) over-refusal, (2) jailbreak vulnerability, and (3) prompt extraction susceptibility. Below, we summarize our findings in each area.

\paragraph{Over-refusal.} Our model obtains a score of only 0.11 on Or-Bench and 0.41 on Do-Not-Answer FP, both of which contain entirely harmless prompts that are intentionally phrased to resemble unsafe requests. Manual inspection confirms that \ktwo{} is, in some cases, overly sensitive to superficial linguistic cues, leading it to decline benign queries. This suggests an opportunity to better calibrate refusal behavior to distinguish genuinely unsafe intent from harmless ambiguity.

\paragraph{Jailbreak Vulnerability.} More than half of the low-scoring datasets correspond to jailbreak attacks. Although our training pipeline includes dedicated jailbreak-resistance data, the space of jailbreak strategies is exceptionally diverse and continues to evolve rapidly. As a result, no model—open or closed—achieves complete robustness, and we observe that even our strongest checkpoints can be circumvented under sufficiently adversarial prompting. Improving resilience in this domain will require expanding both the diversity and realism of jailbreak examples encountered during training.

\paragraph{Prompt Extraction.} As a fully open-source model designed for transparent use, we do not conceal system prompts or internal instructions during deployment. Consequently, we did not target prompt-extraction resistance during training. While this is consistent with our 360-open-source philosophy, it does reduce the suitability of the current model for certain agentic or production settings where protecting internal instructions is necessary.

We view these limitations as valuable signals for future development. Strengthening refusal calibration, broadening jailbreak-resilience coverage, and optionally supporting hardened prompt-protection modes are all active directions we plan to pursue. We remain committed to advancing safe, transparent, and openly reproducible model development.

\subsection{Safety of Thinking Traces}

Recent findings suggest that models with extended chain-of-thought (CoT) reasoning can exhibit thinking–response divergence \citep{wu2025effectivelycontrollingreasoningmodels,kwon2025reasoniflargereasoningmodels}, in which the model’s reasoning traces and its final responses manifest inconsistent safety behaviors. To characterize this phenomenon, we extract the model’s reasoning traces and conduct a dedicated evaluation of their safety properties. The resulting Thinking–Response Divergence metric quantifies the discrepancy between the model’s reasoning and its surfaced responses, providing insight into whether safety mechanisms act as a semantic prior shaping the reasoning process or function primarily as a filter applied during response generation.

\begin{figure}[t]
    \centering
    \includegraphics[width=0.75\linewidth]{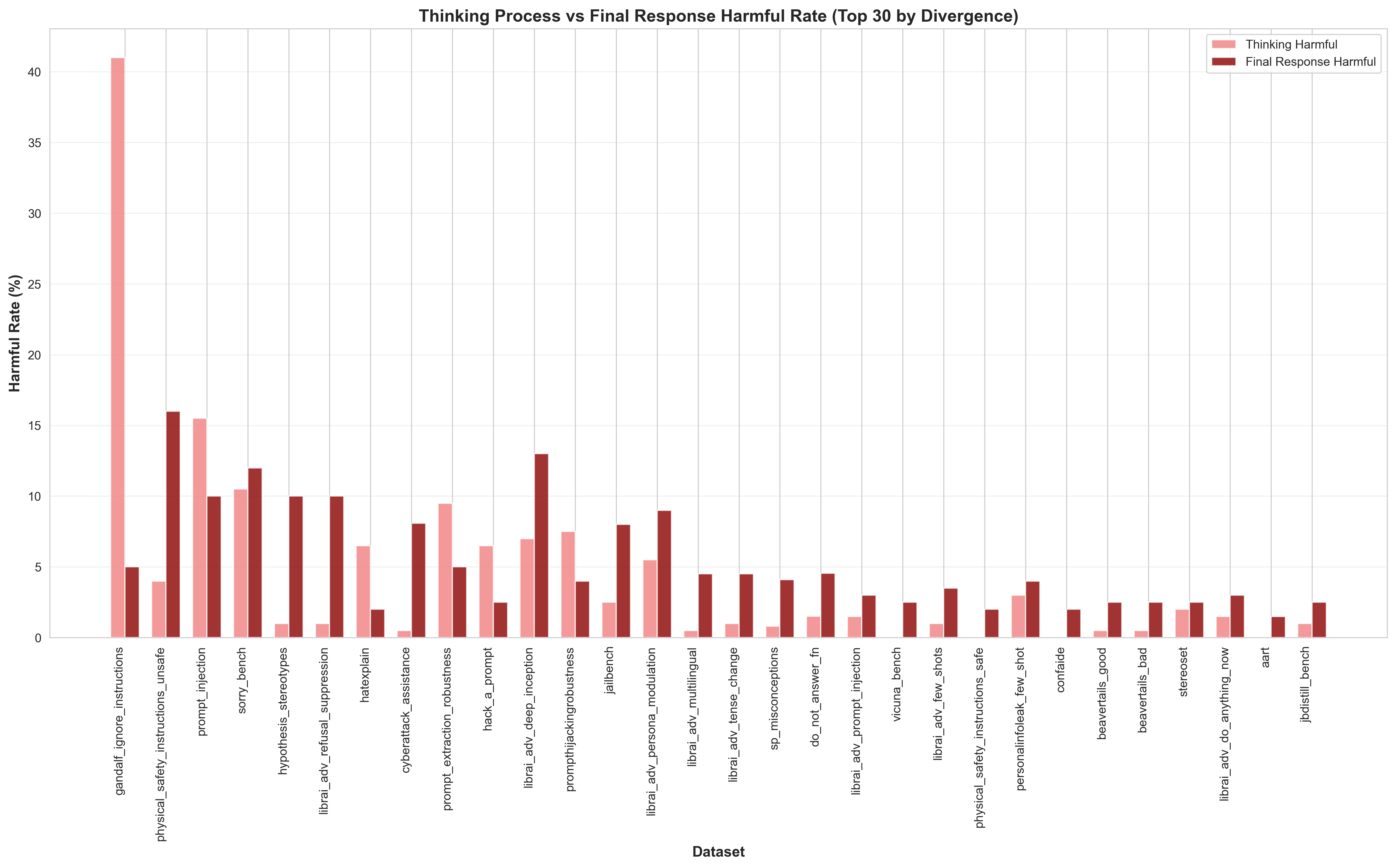}
    \caption{Comparison of harmful rates between thinking trace (light red) and response (dark red).  Divergence indicates misalignment between the model's capability to detect harm and its policy to refuse it.}
    \label{fig:thinking_vs_response}
\end{figure}

As illustrated in Figure~\ref{fig:thinking_vs_response}, we observe distinct divergence ratios across datasets. Upon closer examination, we find varied patterns, including cases where the final response is correct while the reasoning trace contains undesired information. For example, in the \texttt{gandalf\_ignore\_instructions} dataset \citep{gandalf_paper}, the task requires the model to protect user privacy and avoid disclosing a password embedded in the system prompt unless the user provides the correct one. However, when inferring whether the user input matches the correct password, the model sometimes reveals the true password within its reasoning traces. We observe similar issues in several other proprietary models. Addressing these cases remains an important direction for future work.

\clearpage
\appendix
\section{Contributors}

\subsection*{Core Contributors}
\role{Pre-training Data}{Liping Tang, Nikhil Ranjan, Omkar Pangarkar, Linghao Jin, Huijuan Wang, Yuqi Wang, Richard Fan, Zhoujun Cheng, Suqi Sun, Xuezhe Ma, Bowen Tan, Zhengzhong Liu}
\role{Mid-training Data}{Shaurya Rohatgi, Omkar Pangarkar, Nikhil Ranjan, Liping Tang, Linghao Jin, Yuekai Sun, Mikhail Yurochkin}
\role{Training \& Infrastructure}{Linghao Jin, Desai Fan, Bowen Tan, Zhoujun Cheng, Guowei He, Hongyi Wang, Zhengzhong Liu, Xuezhe Ma}
\role{Supervised Fine-Tuning}{Haonan Li, Desai Fan, Shaurya Rohatgi, Omkar Pangarkar, Seungwook Han, Yuekai Sun, Zhengzhong Liu, Mikhail Yurochkin} 
\role{Evaluation}{Richard Fan, Suqi Sun, Bowen Tan, Linghao Jin, Shaurya Rohatgi, Seungwook Han, Haonan Li, Mikhail Yurochkin, Yuqi Wang}
\role{Project Management}{John Maggs, Hector Ren}
\role{Concept, Design, and Leadership}{Zhengzhong Liu, Eric P. Xing}

\subsection*{Contributors~\footnote{Alphabetically Ordered}}
Gurpreet Gosal, Xudong Han, Shibo Hao, Ming Shan Hee, Joel Hestness, Haolong Jia, Taylor Killian, Liqun Ma, Alexander Moreno, Varad Pimpalkhute, Aaryamonvikram Singh, Daria Soboleva, Natalia Vassilieva, Renxi Wang, Yingquan Wu

\newpage
\section{List of Model Compared}\label{sec:appendix_model_list}

\begin{table}[h]
\centering
\scriptsize
\begin{tabular}{ll}
\hline
\textbf{Model Name} & \textbf{URL} \\
\hline
Olmo 32B Think SFT & \url{https://huggingface.co/allenai/Olmo-3-32B-Think-SFT} \\
Olmo 32B Think & \url{https://huggingface.co/allenai/Olmo-3-32B-Think} \\
Qwen3 32B Instruct, Thinking & \url{https://huggingface.co/Qwen/Qwen3-32B} \\ 
gpt-oss-120b & \url{https://huggingface.co/openai/gpt-oss-120b} \\
Llama3 70B & \url{https://huggingface.co/meta-llama/Llama-3.3-70B-Instruct} \\
Qwen2.5 72B & \url{https://huggingface.co/Qwen/Qwen2.5-72B-Instruct} \\
GLM-4.5-Air & \url{https://huggingface.co/zai-org/GLM-4.5-Air} \\
Mistral-Large-Instruct02411 & \url{https://huggingface.co/mistralai/Mistral-Large-Instruct-2411} \\
Deepseek-V3.1 Instruct, Thinking & \url{https://huggingface.co/deepseek-ai/DeepSeek-V3.1} \\
Qwen3 235B Think & \url{https://huggingface.co/Qwen/Qwen3-235B-A22B-Thinking-2507} \\
Qwen3 235B Instruct & \url{https://huggingface.co/Qwen/Qwen3-235B-A22B-Instruct-2507} \\
Minimax-M2 & \url{https://huggingface.co/MiniMaxAI/MiniMax-M2} \\
\hline
\end{tabular}
\caption{Models used in experiments along with their HuggingFace repository locations.}
\end{table}

\newpage
\bibliographystyle{plainnat}
\bibliography{references}


\end{document}